\documentclass[sigconf]{acmart}

\pdfoutput=1

\usepackage{booktabs} % For formal tables

\usepackage{epsfig}
\usepackage{graphicx}
\usepackage{amsmath}
\usepackage{amssymb}
\usepackage{epstopdf}
\usepackage{caption}
\usepackage{subcaption}
\usepackage{multirow}
\usepackage{adjustbox}
\usepackage{subcaption}
\usepackage{xcolor}

\usepackage{capt-of,etoolbox}
\acmYear{2017}
\acmConference{ICMR '17}{}{June 6--9, 2017, Bucharest, Romania.}
\begin{document}
\title{ DRAW: \\Deep networks for Recognizing styles of Artists Who illustrate children's books}
%\titlerunning{Short form of title} 
%\titlenote{Deep networks for Recognizing styles of Artists Who illustrate children's books}
%\subtitle{Extended Abstract}
%\subtitlenote{The }

%
\author{Samet Hicsonmez}
%\authornote{asdada}
%\orcid{asdad}
\affiliation{%
  \institution{Hacettepe University}
%  \streetaddress{P.O. Box 1212}
%  \city{Ankara} 
}
\email{samethicsonmez@hacettepe.edu.tr}

\author{Nermin Samet}
%\authornote{The secretary disavows any knowledge of this author's actions.}
\affiliation{%
  \institution{Middle East Technical University}
%  \streetaddress{P.O. Box 1212}
}
\email{nermin.samet@metu.edu.tr}

\author{Fadime Sener}
\affiliation{%
  \institution{University of Bonn}
%  \streetaddress{1 Th{\o}rv{\"a}ld Circle}
%  \country{Iceland}
}
\email{sener@informatik.uni-bonn.de}

\author{Pinar Duygulu}
\affiliation{
  \institution{Hacettepe University}
%  \streetaddress{P.O. Box 5000}
  }
\email{pinar@cs.hacettepe.edu.tr}
%
%\author{Sean Fogarty}
%\affiliation{%
%  \institution{NASA Ames Research Center}
%  \city{Moffett Field}
%  \state{California} 
%  \postcode{94035}}
%\email{fogartys@amesres.org}
%\affiliation{%
%  \institution{Palmer Research Laboratories}
%  \streetaddress{8600 Datapoint Drive}
%\affiliation{\institution{The Th{\o}rv{\"a}ld Group}}
%\email{jsmith@affiliation.org}
%
%\author{Julius P.~Kumquat}
%\affiliation{\institution{The Kumquat Consortium}}
%\email{jpkumquat@consortium.net}
%
%% The default list of authors is too long for headers}

\begin{abstract}
This paper is motivated from a young boy's capability to recognize an illustrator's style in a totally different context. In the book "We are All Born Free"~\cite{WeAreAllBornFree}, composed of selected rights from the Universal Declaration of Human Rights interpreted by different illustrators, the boy was surprised to see a picture similar to the ones in the "Winnie the Witch" series drawn by Korky Paul (Figure~\ref{fig:Illustrations}). The style was noticeable in other characters of the same illustrator in different books as well. The capability of a child to easily spot the style was shown to be valid for other illustrators such as Axel Scheffler and Debi Gliori. The boy's enthusiasm let us to start the journey to explore the capabilities of machines to recognize the style of illustrators.  

We collected pages from children's books to construct a new illustrations dataset consisting of about 6500 pages from 24 artists. We exploited deep networks for categorizing illustrators and with around 94\% classification performance our method over-performed the traditional methods by more than 10\%. Going beyond categorization we explored transferring style. The classification performance on the transferred images has shown the ability of our system to capture the style. Furthermore, we discovered representative illustrations and discriminative stylistic elements. 

\end{abstract}

%\begin{CCSXML}
%<ccs2012>
%  <concept_id>10010520.10010553.10010562</concept_id>
%  <concept_desc>Computer systems organization~Embedded systems</concept_desc>
%  <concept_significance>500</concept_significance>
%  <concept_id>10010520.10010575.10010755</concept_id>
%  <concept_desc>Computer systems organization~Redundancy</concept_desc>
%  <concept_significance>300</concept_significance>
%  <concept_id>10010520.10010553.10010554</concept_id>
%  <concept_desc>Computer systems organization~Robotics</concept_desc>
%  <concept_significance>100</concept_significance>
%  <concept_id>10003033.10003083.10003095</concept_id>
%  <concept_desc>Networks~Network reliability</concept_desc>
%  <concept_significance>100</concept_significance>
%</ccs2012>  
%\end{CCSXML}

%\ccsdesc[500]{Computer systems organization~Embedded systems}

% We no longer use \terms command


\begin{teaserfigure}
\centering
\begin{tabular}{cccccc}
%%%%% 1st row 
\includegraphics[width=0.08\textwidth]{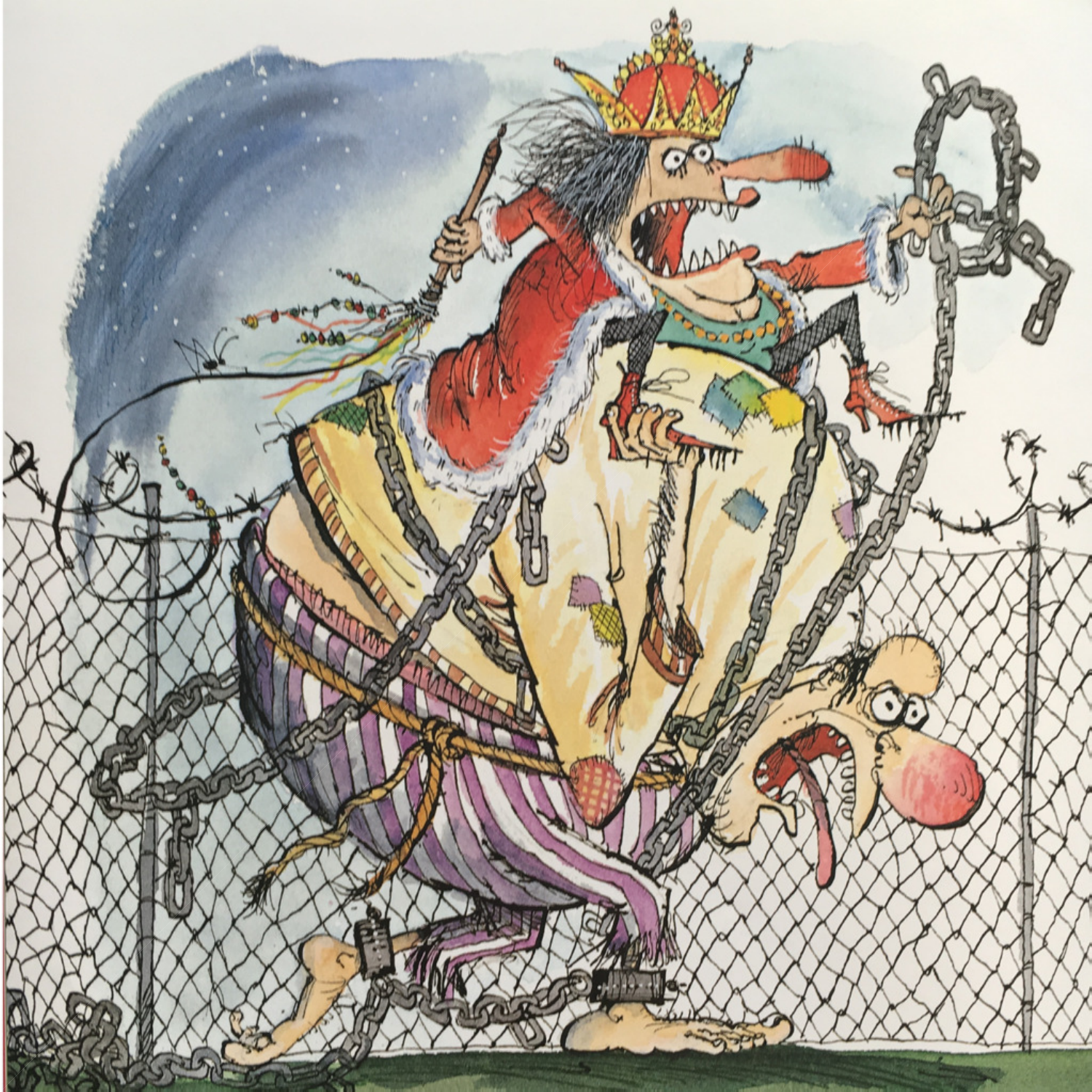} 
 \hspace*{0.001cm}
\includegraphics[width=0.08\textwidth]{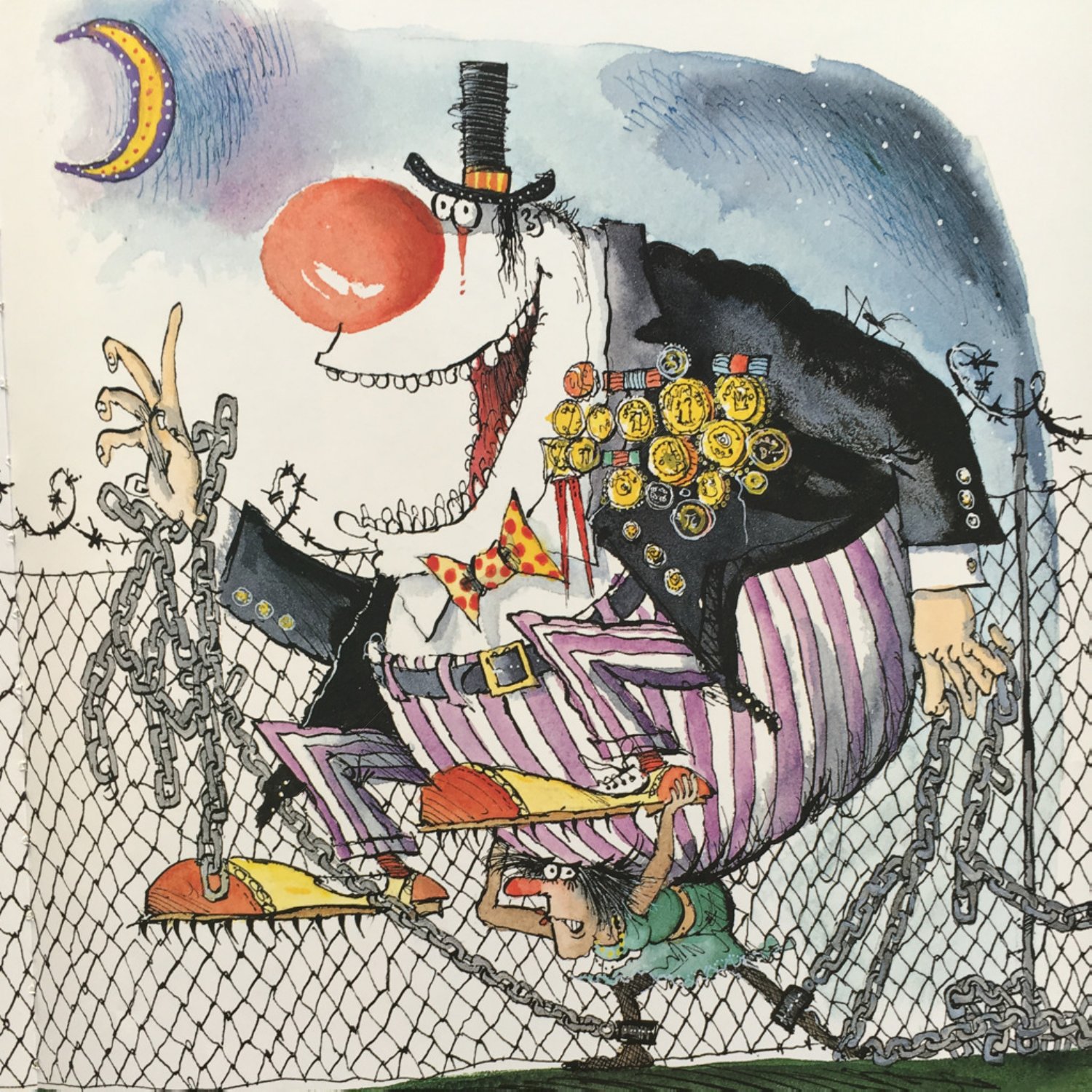} 
 \hspace*{0.25cm}
\includegraphics[width=0.08\textwidth]{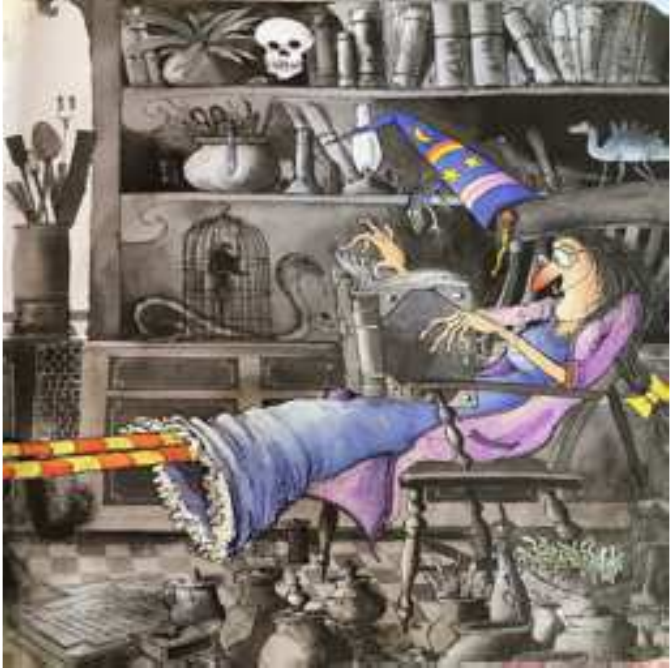} 
 \hspace*{0.001cm}
\includegraphics[width=0.08\textwidth]{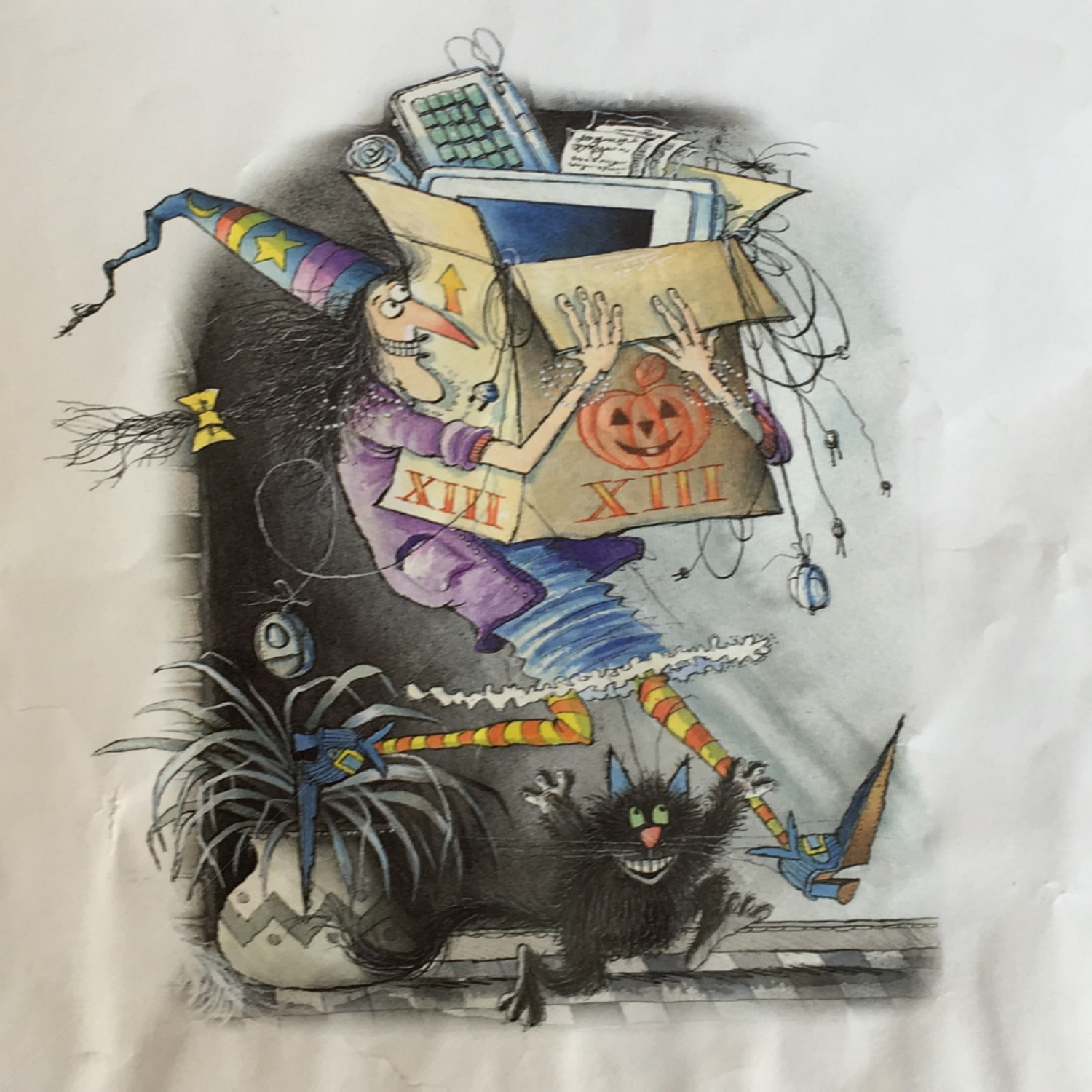} 
 \hspace*{0.25cm}
\includegraphics[width=0.08\textwidth]{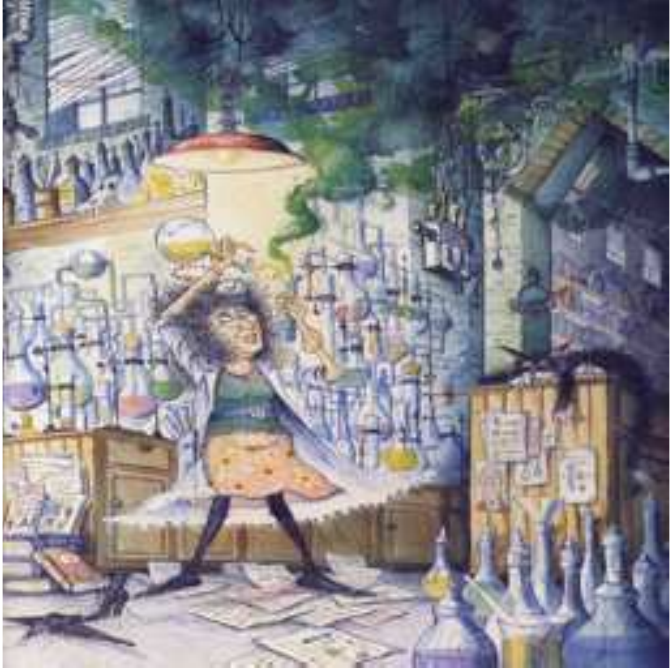} 
 \hspace*{0.001cm}
\includegraphics[width=0.08\textwidth]{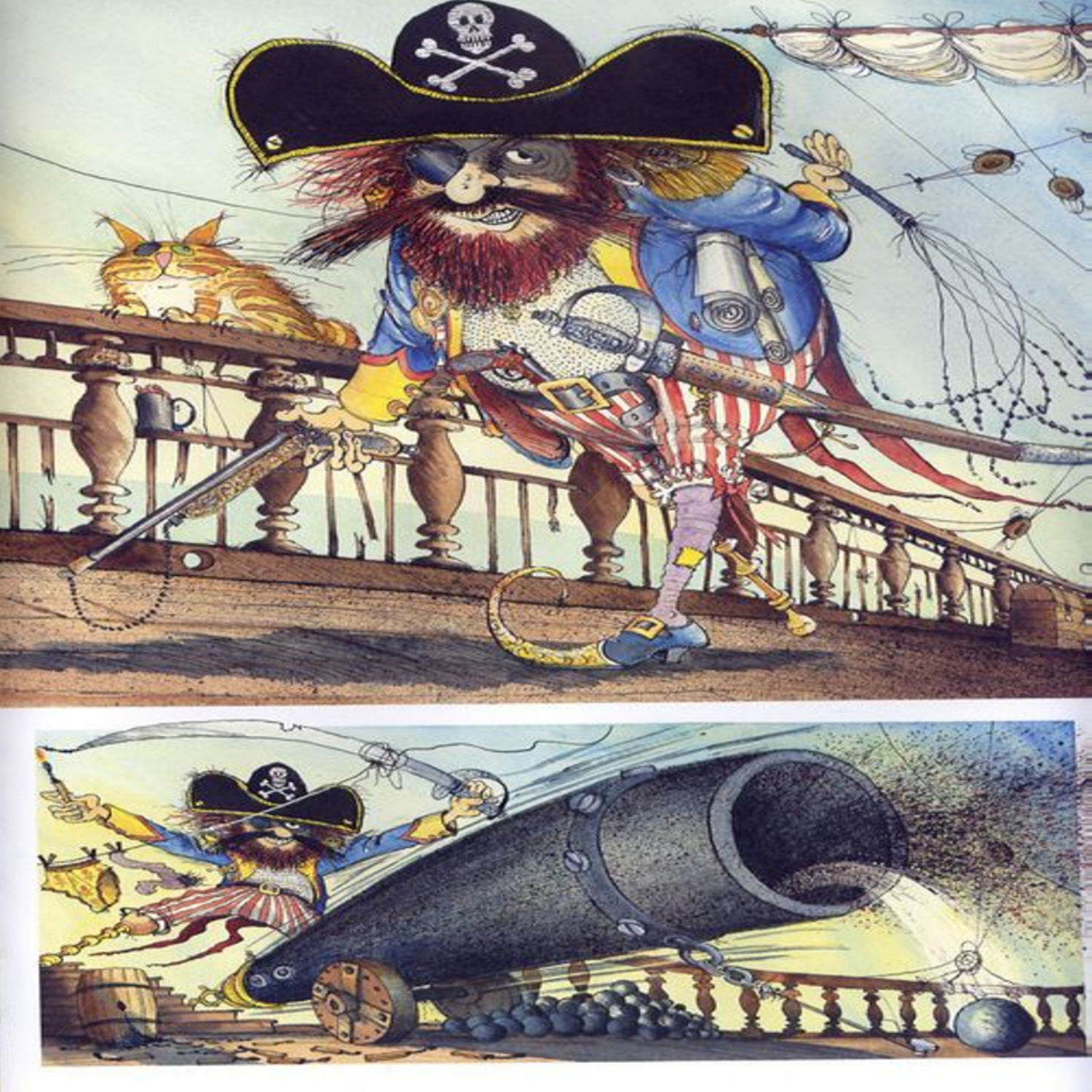} \\
%%%%% 2nd row 
\includegraphics[width=0.08\textwidth]{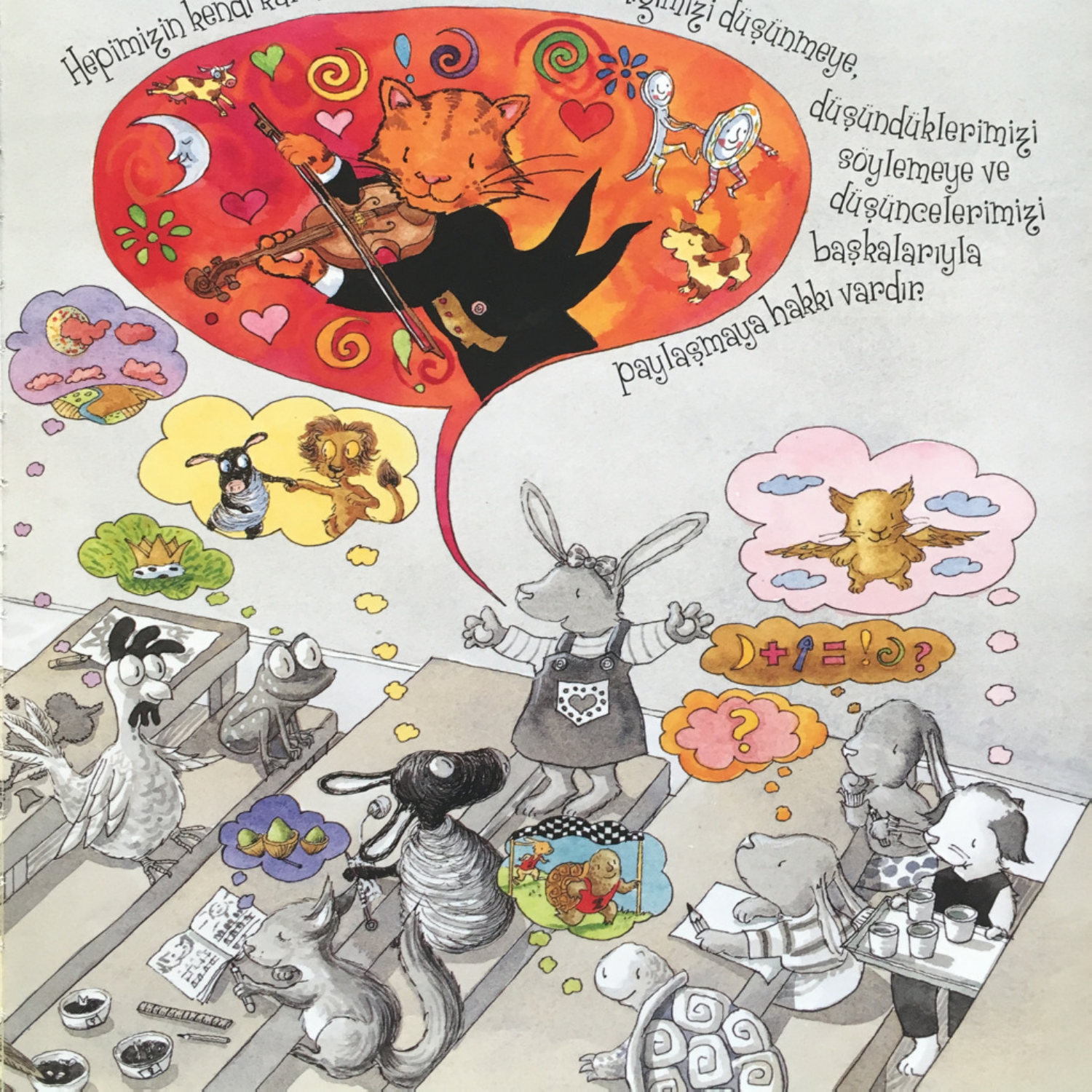} 
 \hspace*{0.001cm}
\includegraphics[width=0.08\textwidth]{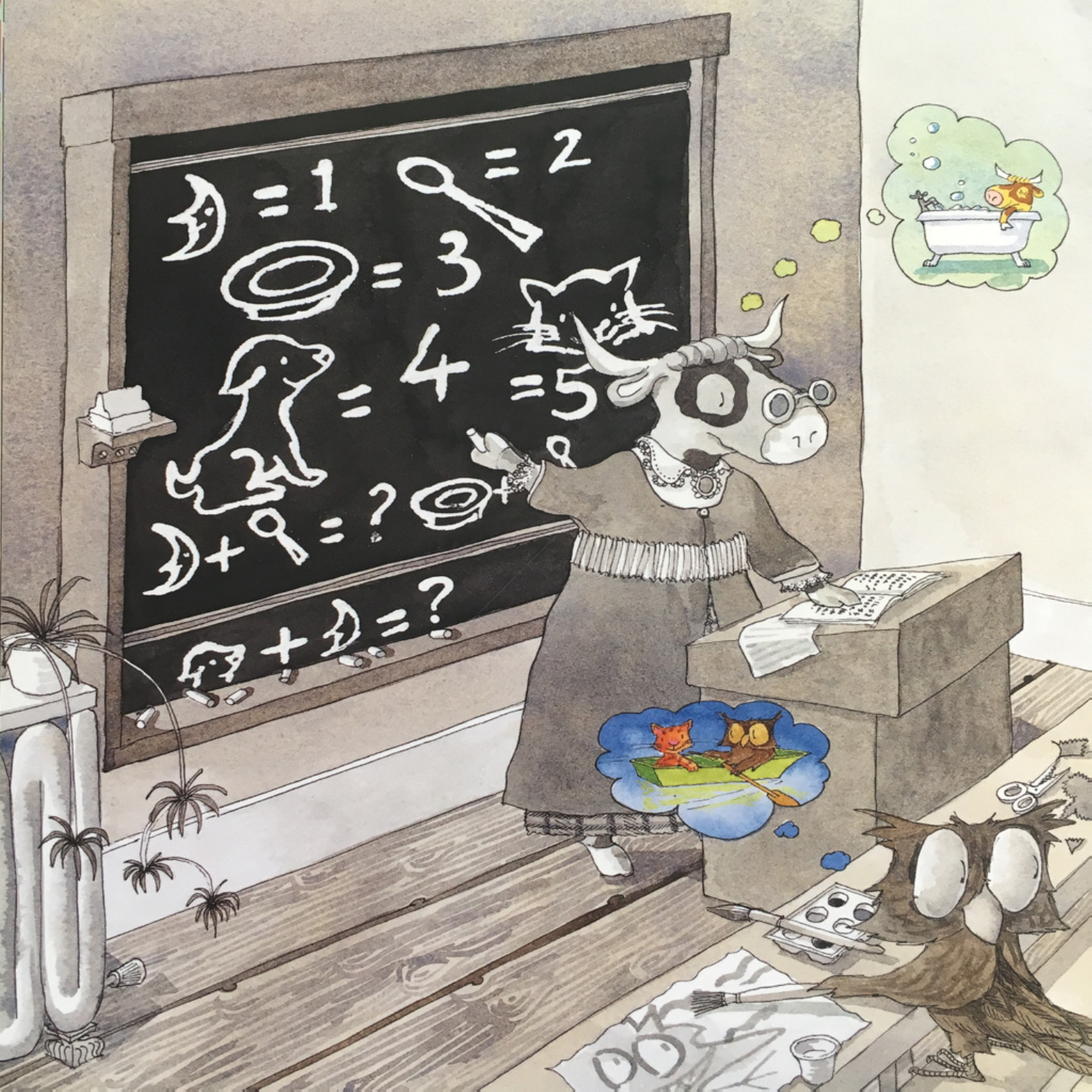} 
 \hspace*{0.25cm}
\includegraphics[width=0.08\textwidth]{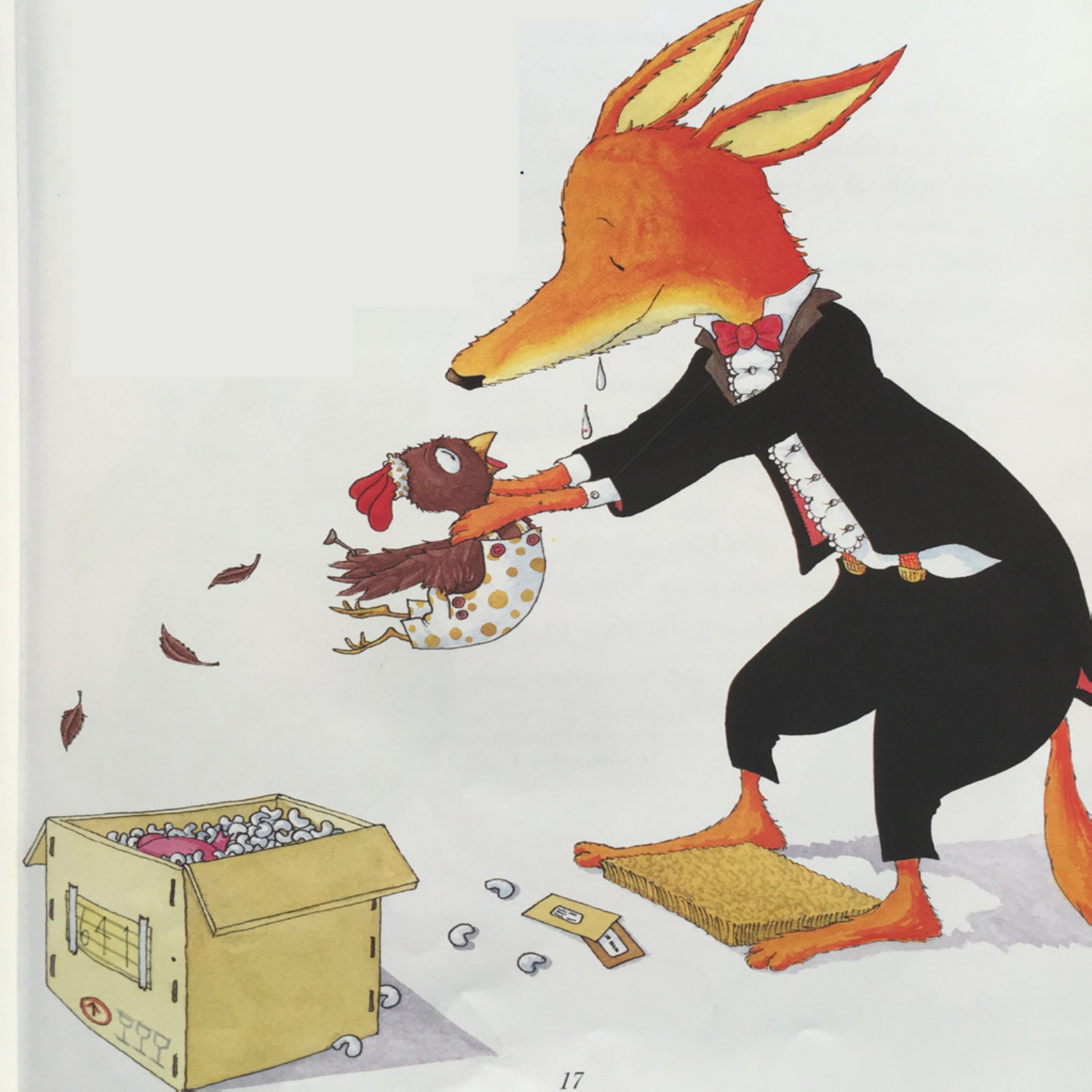} 
 \hspace*{0.001cm}
\includegraphics[width=0.08\textwidth]{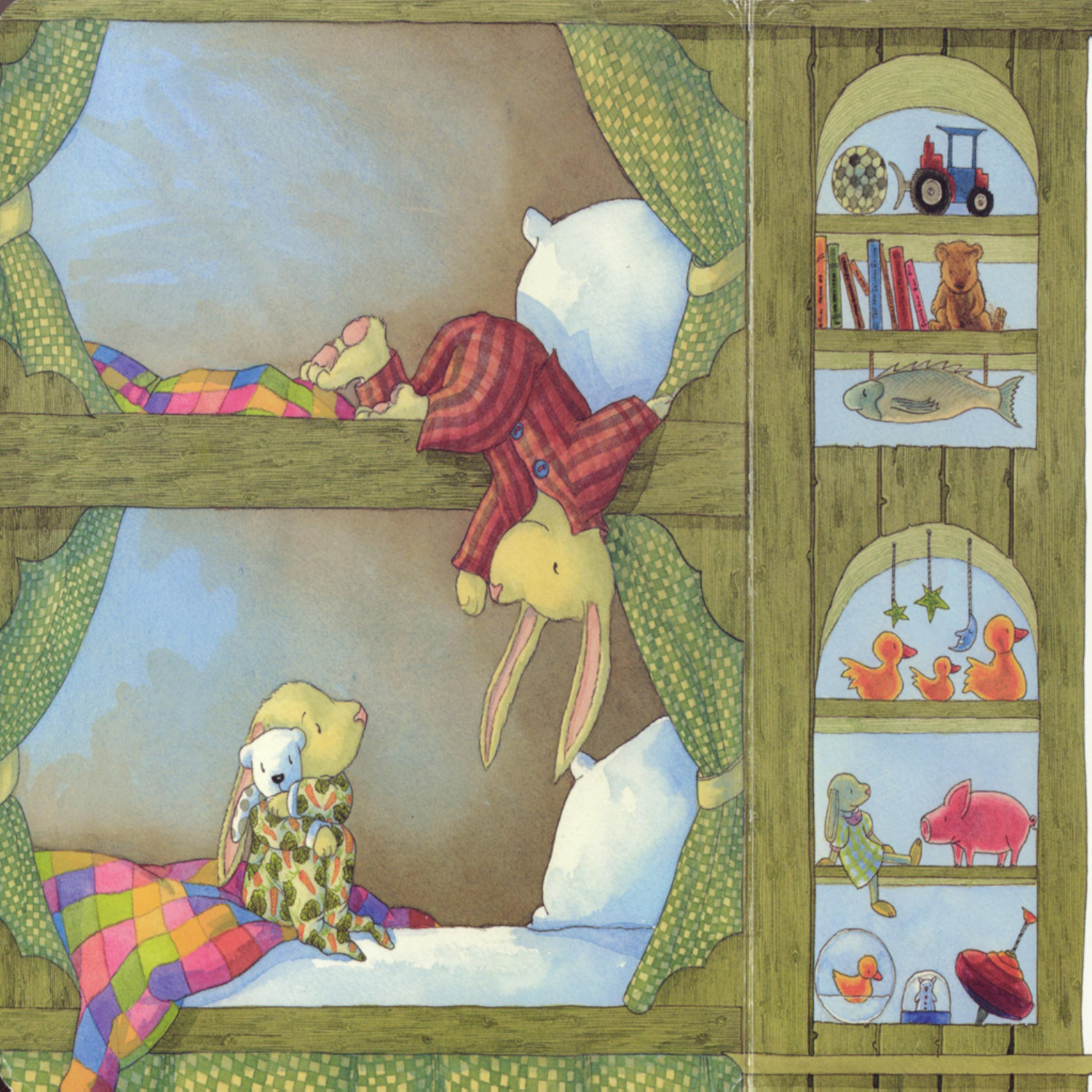} 
 \hspace*{0.25cm}
\includegraphics[width=0.08\textwidth]{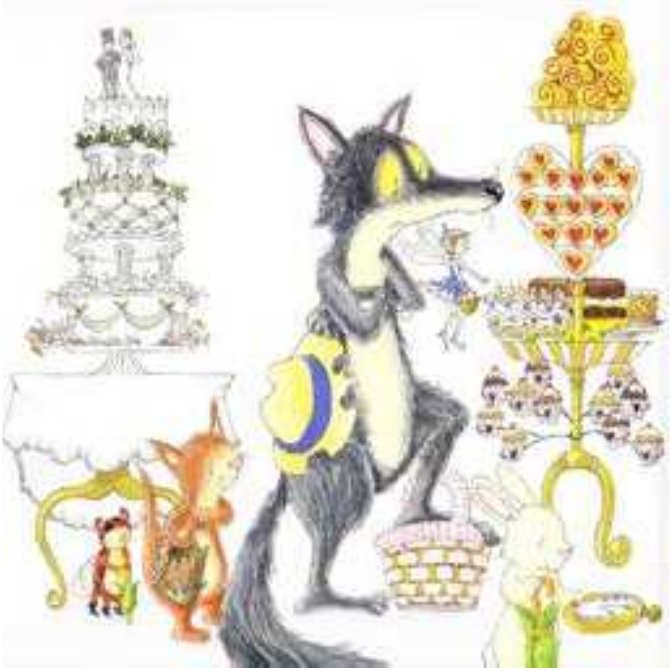} 
 \hspace*{0.001cm}
\includegraphics[width=0.08\textwidth]{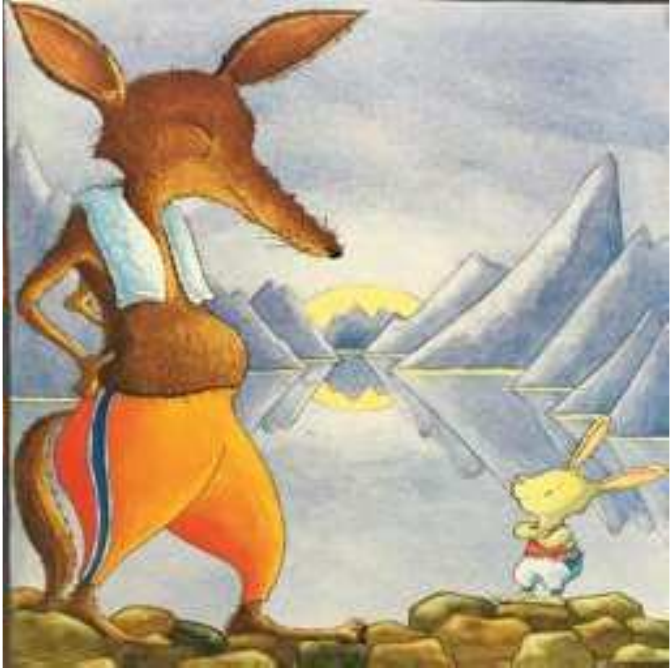} \\
%%%%% 3rd row 
\includegraphics[width=0.08\textwidth]{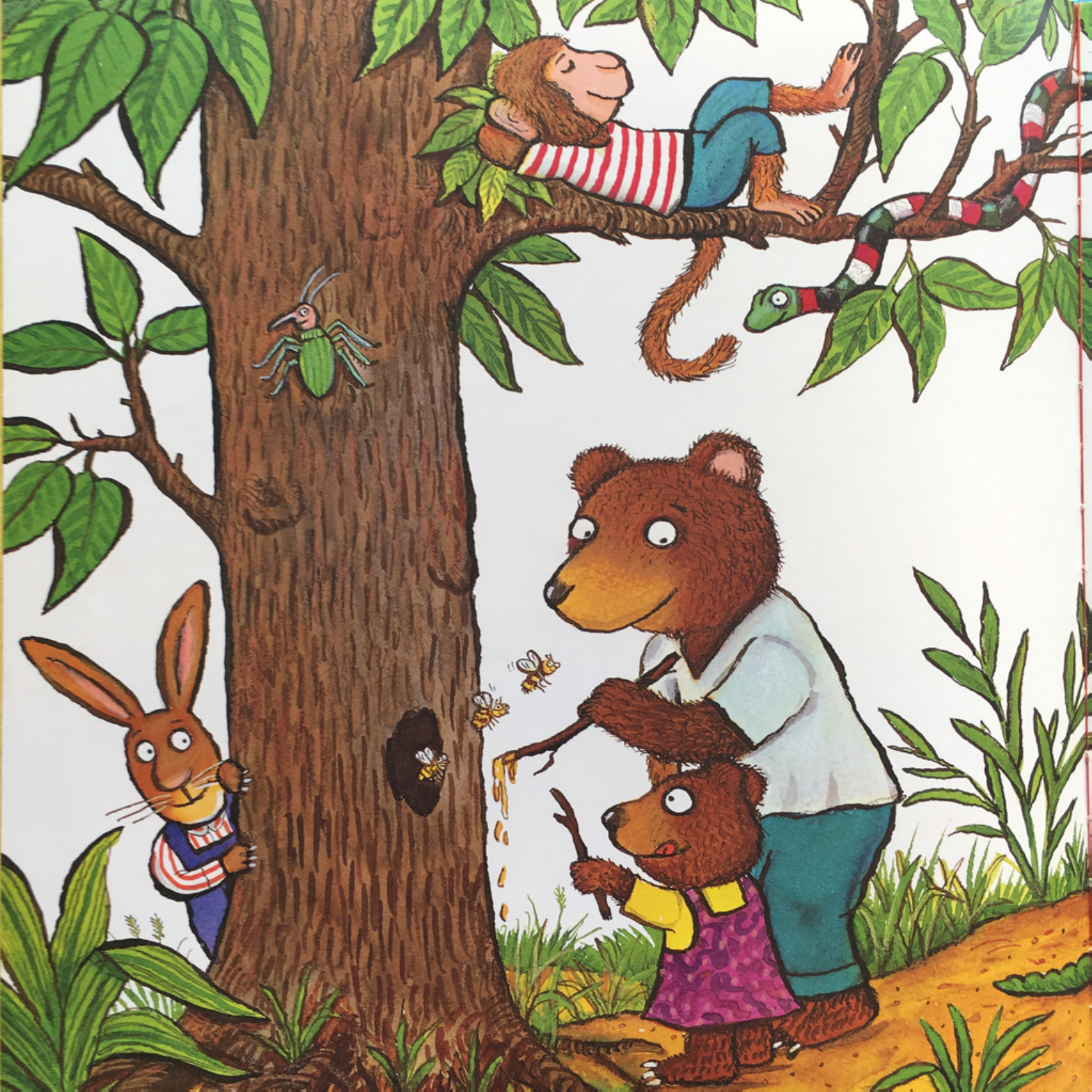} 
 \hspace*{0.001cm}
\includegraphics[width=0.08\textwidth]{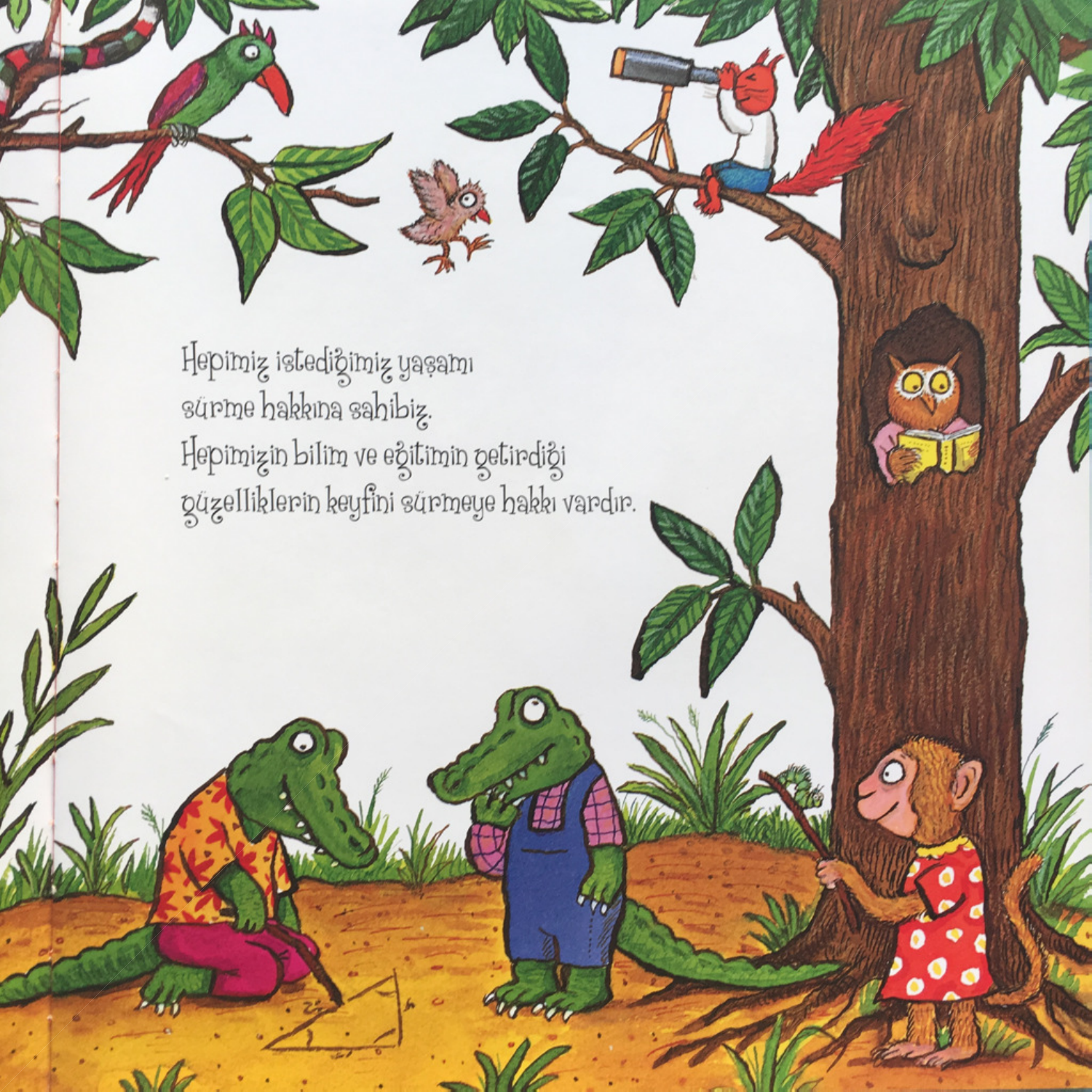} 
 \hspace*{0.25cm}
\includegraphics[width=0.08\textwidth]{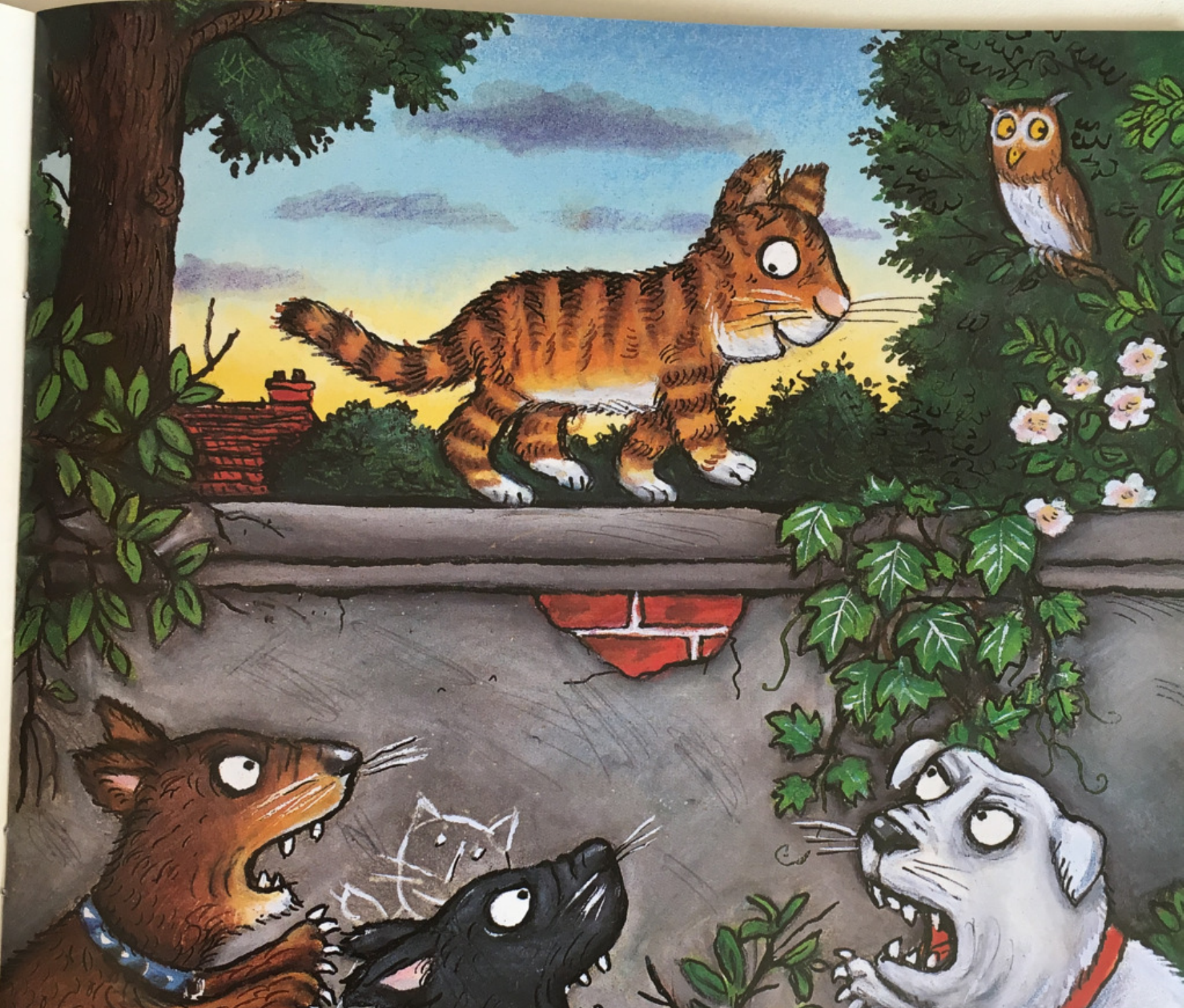} 
 \hspace*{0.001cm}
\includegraphics[width=0.08\textwidth]{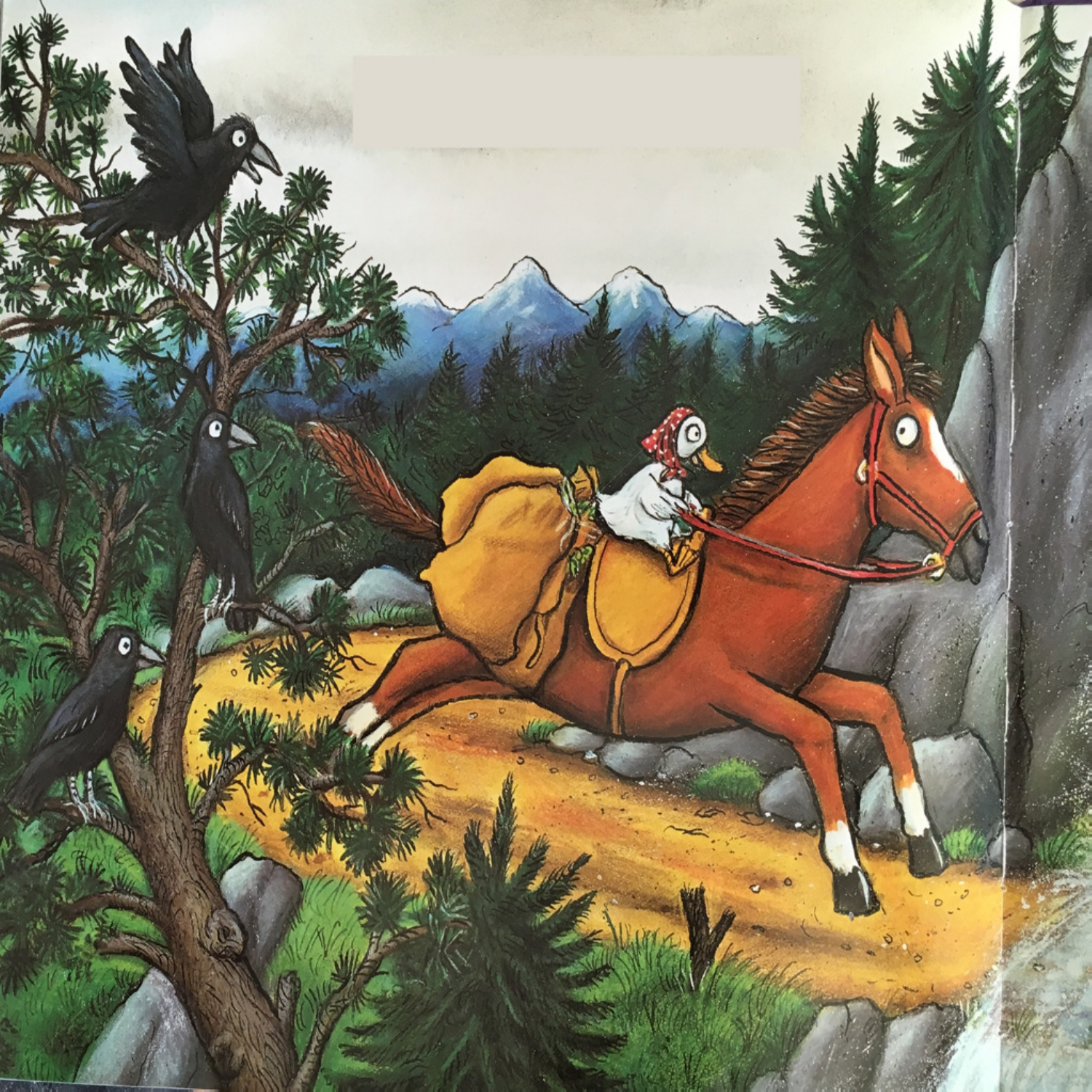} 
 \hspace*{0.25cm}
\includegraphics[width=0.08\textwidth]{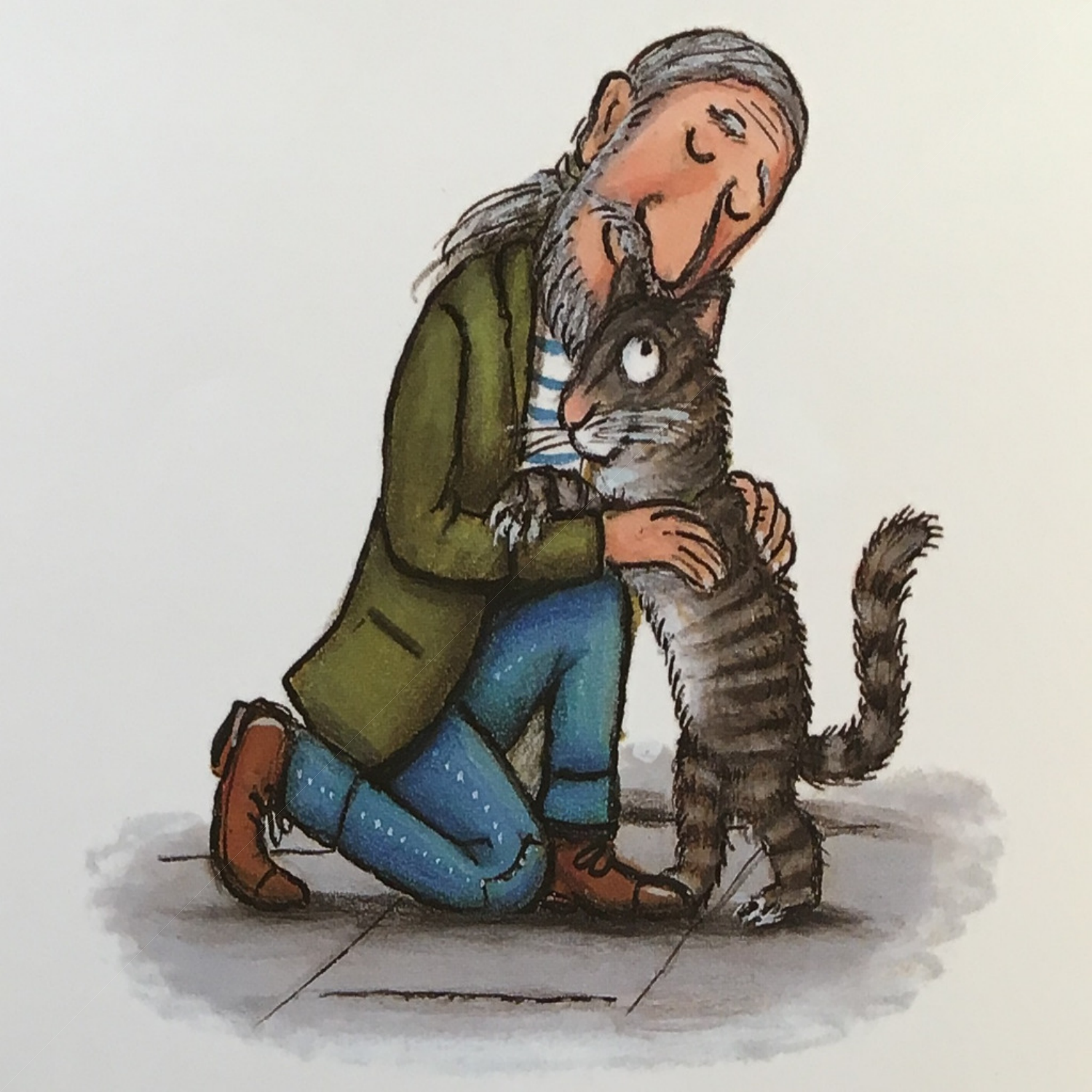} 
 \hspace*{0.001cm}
\includegraphics[width=0.08\textwidth]{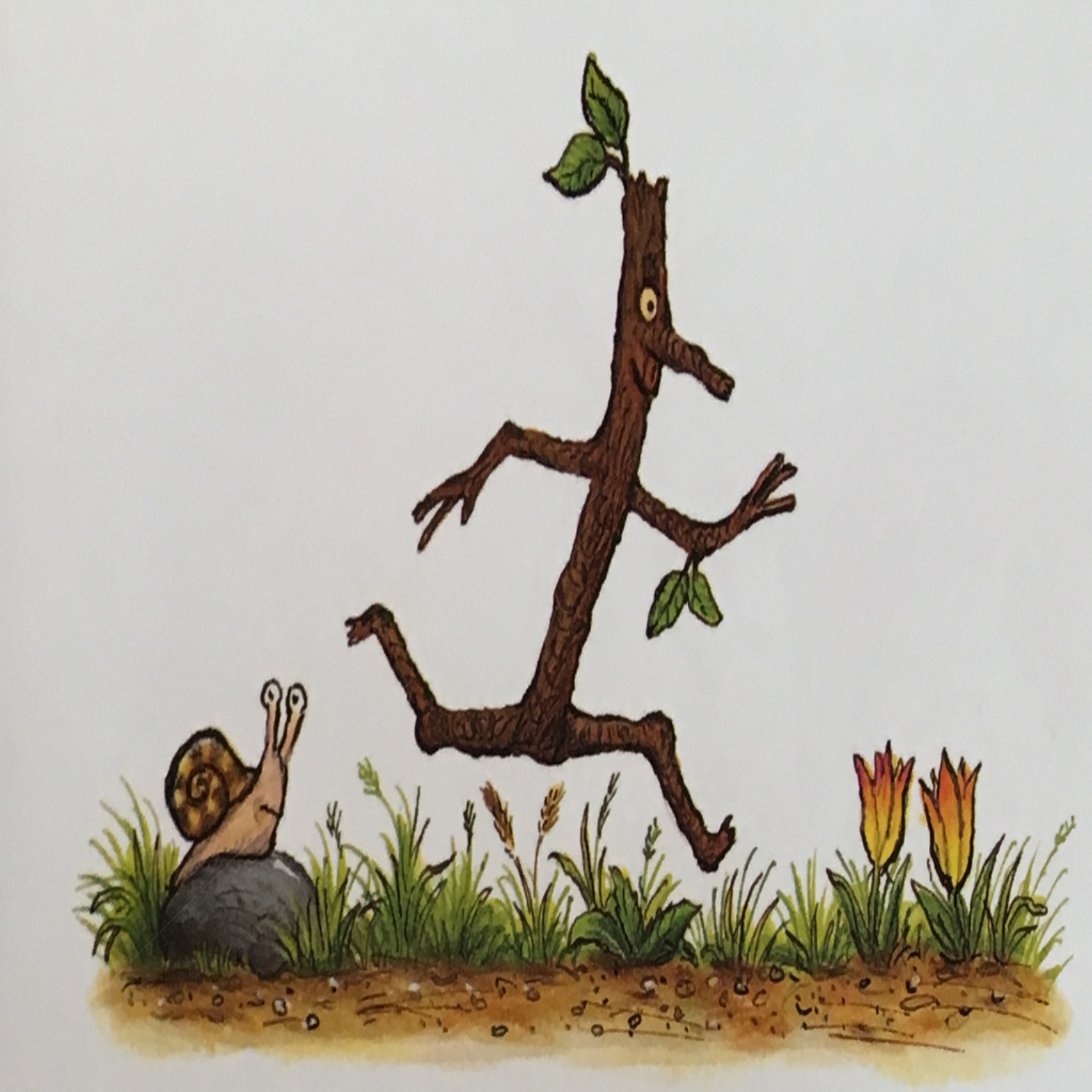} \\
 \end{tabular}
  \captionof{figure}{First and second columns show pages from~\cite{WeAreAllBornFree} and the other columns  show more examples from the same illustrators. Top: Korky Paul, middle: Debi Gliori, bottom: Axel Scheffler. As seen the styles are disctinctive for illustrators.}
\label{fig:Illustrations}
\end{teaserfigure}

\maketitle

\section{Introduction}
\label{sec:intro}
Illustrations help us to understand the message clearly and have been widely used in printed and visual media. Yet, the role of illustrations in children's books is more than being a simple picture accompanying the text. For the children who don't know how to read, those are the illustrations that make the children to understand the story. Those images help them to identify the characters, scenes and events in the books and let them to be prepared for the fascinating world of words when they start to read by themselves\footnote{http://www.maaillustrations.com/blog/article/the-role-of-illustration-in-childrens-book/}. 

This fact inspires many artists to draw illustrations for children's books. On the other hand, understanding, predicting, and analyzing people's taste of reading is a challenging problem, since the taste can depend on individuals' philosophical, psychological, political backgrounds. When it comes to children's books, especially from a child's perspective the choice mostly depends on the visual illustrations. Discovering the taste requires the understanding of the style characteristics of the illustrators.  Motivated by this observation, in this study we aim to understand the style of artists who draw children's books. 

Automatic understanding of artistic images could assist in organizing large 
collections and could be useful for art recommendation systems.
However, it is a difficult task mostly due to varying stylistic behavior of different artists.
Particularly with the increase of deep structures there has
been an interest towards this relatively less explored area. 

There have been recent efforts to understand
aesthetic perception of art works such as investigating the potential of a computer
to make aesthetic judgments~\cite{SprattE14}, quantifying creativity~\cite{elgammal2015quantifying}, aesthetic analysis of images by feature discovery~\cite{campbell2015feature}, and analyzing the artistic influence by comparing them to others~\cite{saleh2014toward}.  Even though classifying art is qualitative~\cite{dimaggio1987classification}, classification of art works has also emerged as another line of work. 
Bar et. al.~\cite{bar2014classification} worked on classification of
artistic styles by presenting a perceptiveness of deep neural network features in
identifying artistic styles in paintings. Li et. al.~\cite{LiChen} worked on automatically classifying paintings
as aesthetic or not. Lyu et. al.~\cite{lyu2004digital} focused on painter authentication. 
Identification of painters is also studied based on wavelet
analysis of brush strokes in paintings~\cite{Wang_2008_6032,LiWangJia}.
In~\cite{wiki_art_cnn, wiki_art_dataset} they aimed to classify fine-art paintings using CNNs on "Wikiart paintings"~\cite{karayev2013recognizing} data set. In ~\cite{wiki_art_cnn} they conducted experiments on their proposed CNN which is very similar to AlexNet~\cite{alexnet}. Best result is achieved when network is first trained on ImageNet dataset~\cite{imagenet}, then transfer learning applied to the network. 

Inspired by capabilities of humans who are able to recognize objects regardless
being in art or photography, Cai et. al. worked on automatically identifying objects in cross domains~\cite{CaiWCH15}. In~\cite{Crowley14, Crowley14a}, the authors focus on recognizing objects  in paintings learned from natural images. 

Collecting and labeling a dataset for artistic images is also a challenging task. Mensink et. al.~\cite{mensink2014rijksmuseum}
introduced a diverse dataset of over 1 million artworks, 700,000 of which are
prints to support and evaluate art classification. Carneiro et. al.~\cite{carneiro2012artistic}
presented a database of monochromatic artistic images. 
Crowley et al. ~\cite{Crowley14, Crowley14a} annotate a subset of publicly available 'Your Paintings'~\cite{yourpaintings} data set images with ten category labels from the PASCAL VOC data set~\cite{pascal-voc-2011}. In ~\cite{khan2014painting} presented a dataset which contains 4266 paintings from 91 different painters. 
Karayev et. al.~\cite{karayev2013recognizing} presented two novel data sets, one of them contains 80K Flickr photographs annotated with 20 style labels such as vintage, romantic, HDR etc., and the second one consist of 85K art paintings from 25 art styles like Baroque,  Roccoco, Cubism etc. 
%They find features extracted from pre-trained CNN model are the best for the task of aesthetic style analysis.

Some works concentrated on transferring artistic styles from  style images such as paintings to content images such as selfie pictures~\cite{style_transfer_2, johnson2016perceptual, rew_istek_3}. In~\cite{style_transfer}, the artistic style transfer pipeline tries to minimize feature reconstruction loss and style reconstruction loss at the same time by using features from pre-trained CNN model with forward and backward passes. Since backward computations increases computation time, to overcome this,~\cite{johnson2016perceptual} proposed a similar approach with using forward passes to minimize both feature and style reconstruction losses. 
Kyprianidis et. al.~\cite{kyprianidis2013state} presented a survey on state of the art techniques for transforming images and videos into artistically stylized renderings.

The studies that try to identify the style or genre for art images could be considered similar to ours~\cite{wiki_art_dataset, karayev2013recognizing, rew_istek_1, rew_istek_2}. However, they define style as a more generic term shared by several artists.  The work in~\cite{thomas2015s} that identifies the authorship of photographs, that is the photographer, is the most similar one to ours. Deep networks are also utilized in that study for qualitative evaluations.  

In the illustrator identification domain,  based on our knowledge the only work is~\cite{sener2012identification} where they tried to identify only four illustrators on a very small data set. They utilized several low-level descriptors such as HOG, GIST and SIFT and used a bag of words model to classify illustrations. In this work, we collected a larger data set and used their results as our baselines.

In some recent studies, illustrations are considered in the form of clip arts. In~\cite{Garces-2014}, a style similarity metric is designed by combining color, shading, texture and stroke features with relative comparisons collected via AMT, and this work was leveraged in~\cite{Garces-2016}
to obtain aesthetically coherent clusters for visualizations of clip art datasets. In~\cite{Furuya}, an unsupervised approach is proposed for stylistic comparison of illustrations again in the form of clip-arts.
The illustrations that we consider are specific to the artistic drawings in children books, and they are more challenging than the illustrations in clip-arts. \\

\noindent{\bf Our contributions:} We have several important contributions that will be described in detail in the following sections:
{\bf (1)} We attack to the problem of classifying styles of illustrators which is a more challenging task than classifying the content. {\bf (2)} We have constructed a new dataset of illustrations. Based on our knowledge this is the first comprehensive dataset specific to artistic illustrations from books. {\bf (3)} We focus on illustrations in children's books which have distinct characteristics in the sense that the imagination could lead to extreme characters and settings to happen that are difficult to be found in most of the photographs and paintings.  {\bf (4)} We explored different deep networks and compared them with low level features. {\bf (5)} We tested three different strategies for categorisation: novel instance recognition from seen books as well as unseen books, and book recognition. {\bf (6)} We exploited the style transfer method and showed the qualitative results for transferring styles from illustrators to cartoon images and natural photographs as well as to the illustrations of other illustrators. {\bf (7)} Further, we provided quantitative results for illustrator to illustrator transfer  utilizing the style categorization. {\bf (8) } We compared different methods and features in choosing representative illustrators and discriminative patches.   

%We also present extracted discriminative patches form learned filters in order to observe the characteristic of our illustration dataset. 
%Automatically recognizing illustrators from images is challenging since illustrators even can draw each of their different books in different ways. Besides famous children's book illustrators, there are many other illustrators. Furthermore, there are also lots of adapted series of books in literature. In our work,  we also try to predict illustrator of an unseen illustration book based on deep features.We demonstrate a variety of experiments including conventional methods and different deep architectures. Finally, wepresent and discuss detailed experiments of our system.}

%---------------------------------------------------------------------------------------------------------------------------Dataset

%-----------------------Table Dataset
\begin{table*}[t]
\centering
\caption{ Summary statistics of our Illustration dataset.} 
\begin{adjustbox}{max width=\textwidth}
\begin{tabular}{lcclcc}
\hline
 & \multicolumn{2}{c}{Dataset}  &     & \multicolumn{2}{c}{Dataset} \\
\cline{2-3} \cline{5-6}
Id - Illustrator & Book Cnt & Image Cnt & Id - Illustrator & Book Cnt & Image Cnt \\
\hline
 01- Axel Scheffler   & 13 & 532 & 13- Korky Paul  & 15 & 427 \\ 
 02- Ayse Inal   & 4 & 120 & 14- Leo Lionni  & 10 & 314 \\ 	
 03- Beatrix Potter & 9 & 269 & 15- Marc Brown  & 17 & 360 \\  
 04- Behic Ak  & 12 & 385 & 16- Maurice Sendak  & 7 & 263 \\  
 05- Bill Peet  & 11 & 513 &  17- Mo Willems  & 6 & 268 \\  
 06- David Mckee  & 12 & 199 & 18- Mustafa Delioglu  & 9 & 160 \\ 
 07- Debi Gliori  & 12 & 275 & 19- Patricia Polacco  & 9 & 284 \\  
 08- Dr. Seuss & 15 & 455 &  20- Ralf Butschkow & 6 & 152 \\  
 09- Eric Carle  & 14 & 304 & 21- Rosa Curto  & 8 & 288 \\  
 10- Eric Hill  & 9 & 148 & 22- Serap Deliorman  & 5 & 158 \\  
 11- Feridun Oral  & 5 & 140 & 23- Stephen Cartwright  & 5 & 179 \\ 
 12- Kevin Henkes  & 3 & 86 &  24- Tony Ross  & 7 & 189 \\ 
\hline
& Total Number of &  Illustrators: 24 & Books: 223 & Illustrations: 6468 &   \\
\hline 
\end{tabular}
\end{adjustbox}
\label{tab:ill_data_set_info}
\end{table*}

%-----------------------Figure Dataset
\begin{figure*}[t]
\centering

\includegraphics[width=0.075\textwidth]{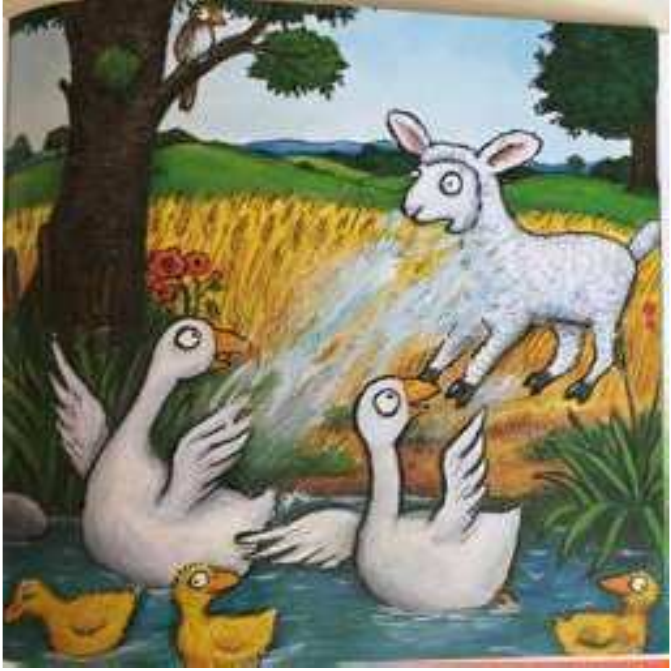}
\includegraphics[width=0.075\textwidth]{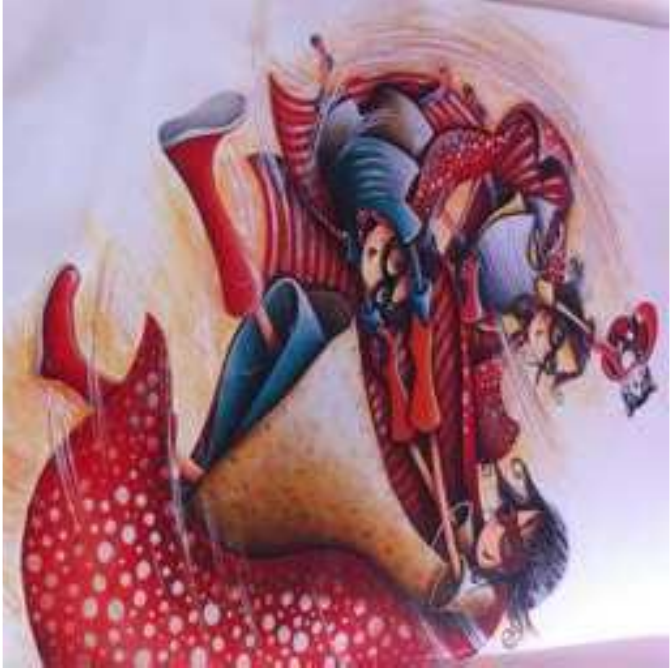}
\includegraphics[width=0.075\textwidth]{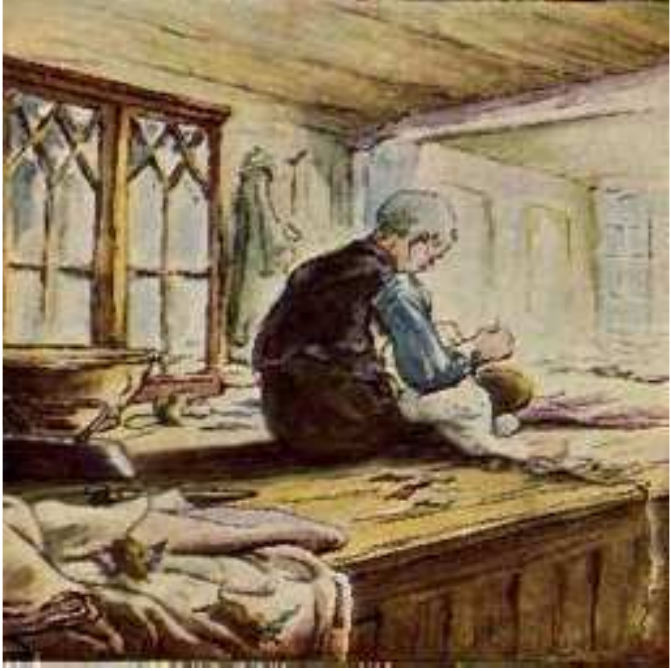}
\includegraphics[width=0.075\textwidth]{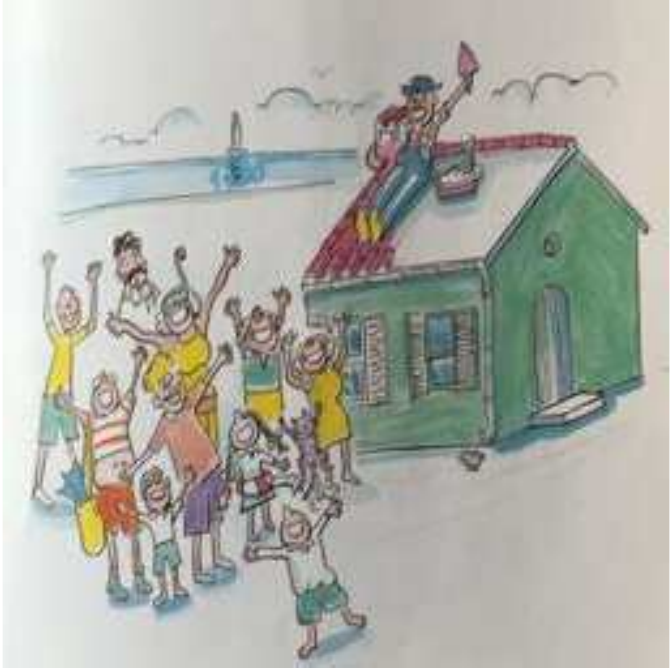}
\includegraphics[width=0.075\textwidth]{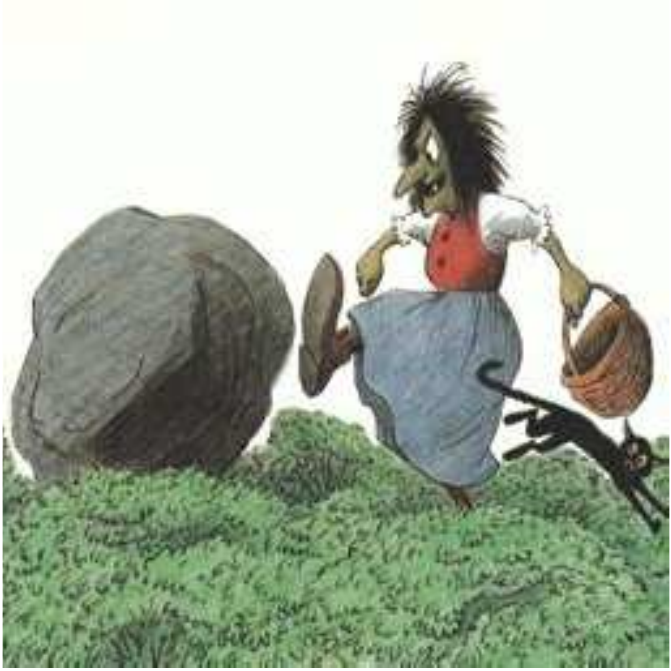}
\includegraphics[width=0.075\textwidth]{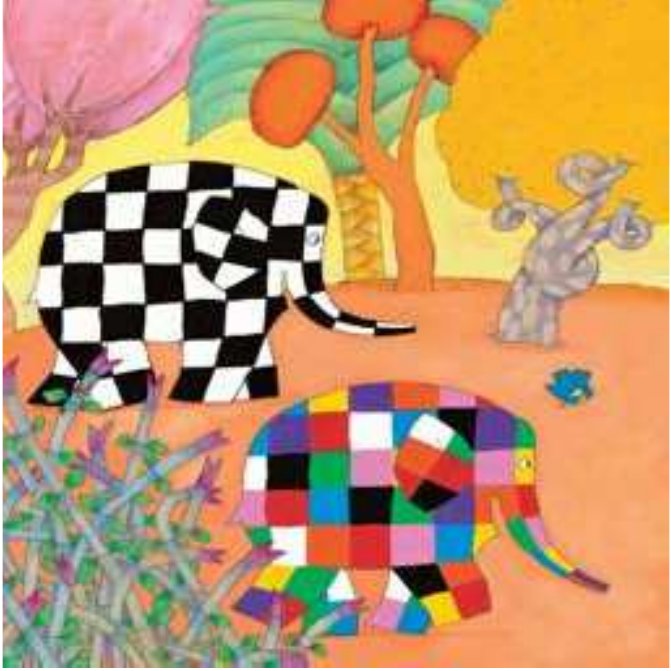}
\includegraphics[width=0.075\textwidth]{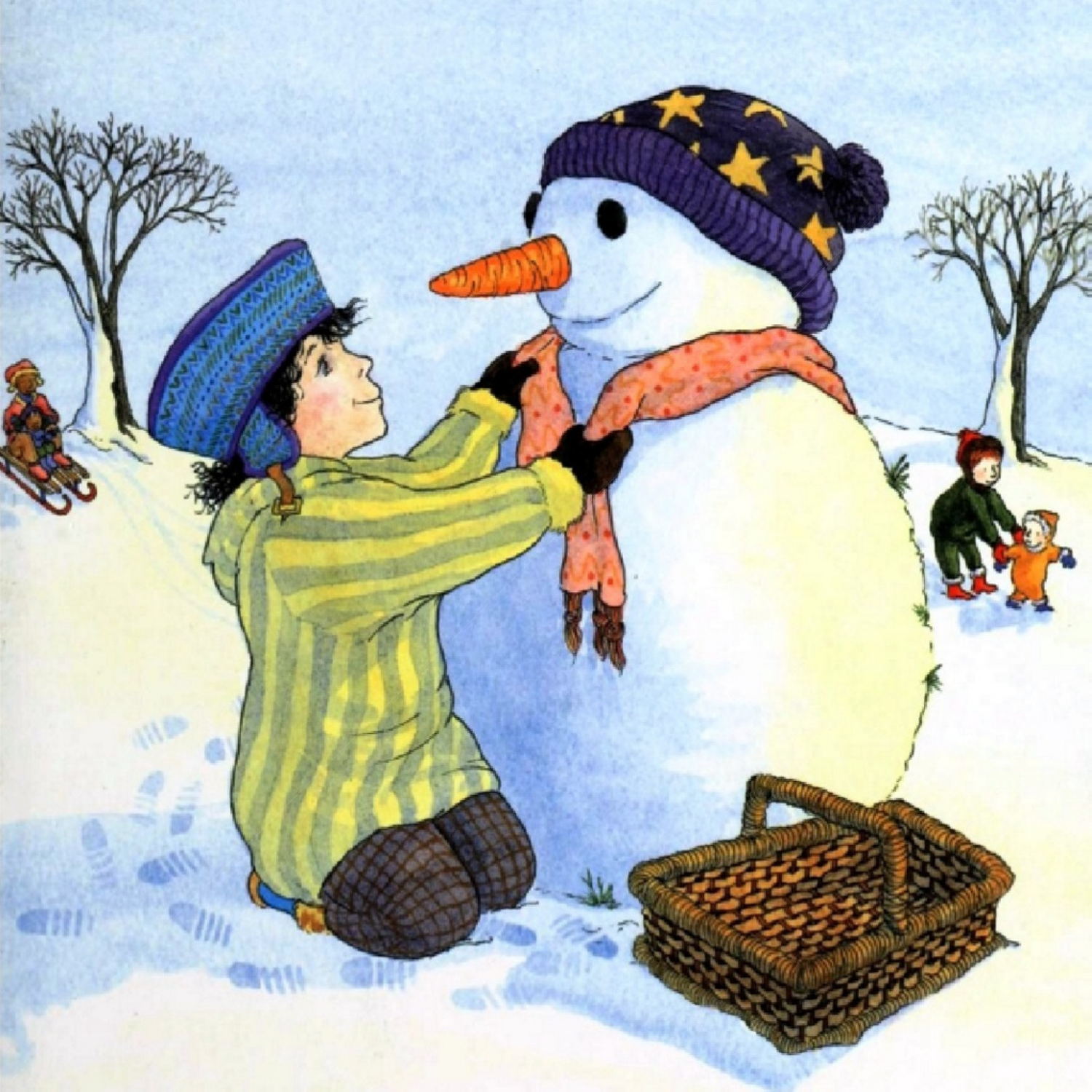}
\includegraphics[width=0.075\textwidth]{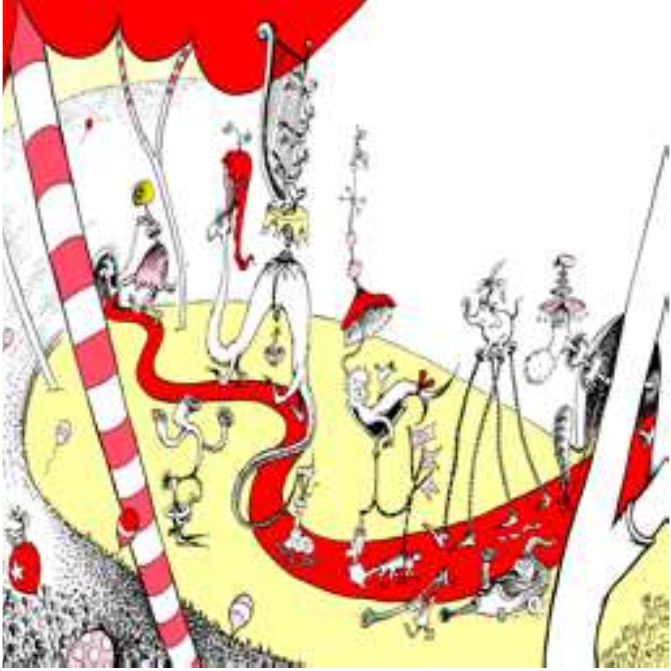}
\includegraphics[width=0.075\textwidth]{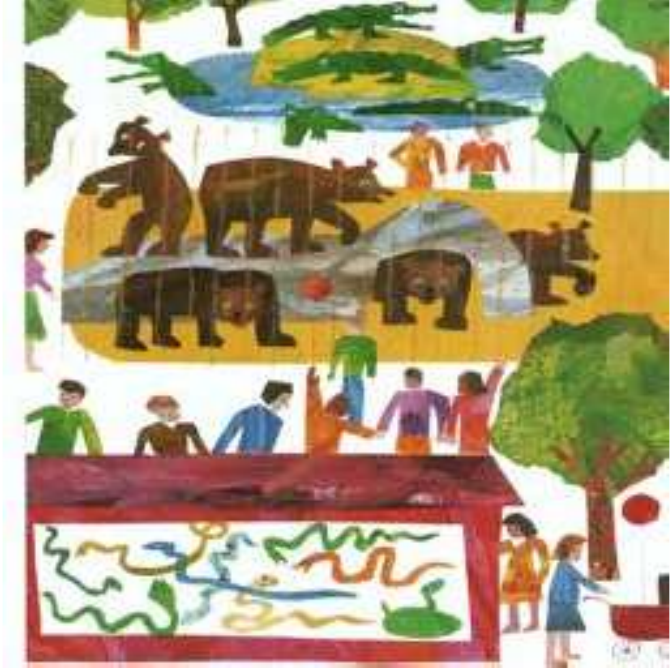}
\includegraphics[width=0.075\textwidth]{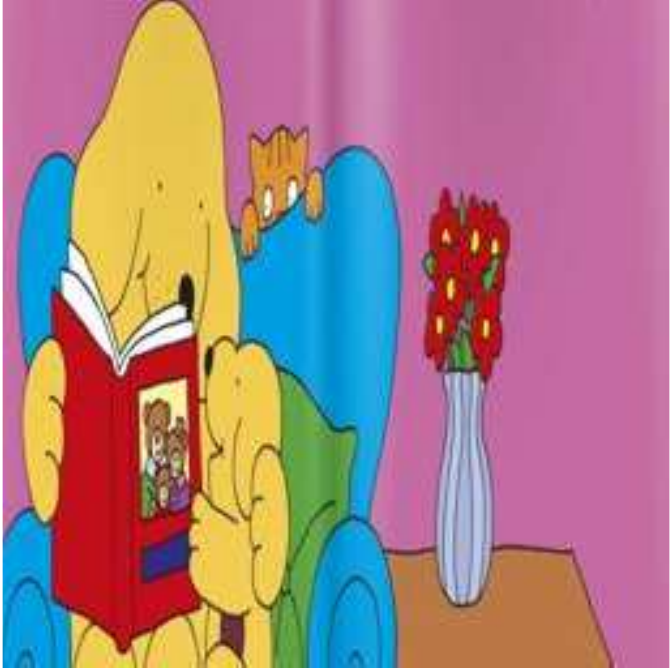}
\includegraphics[width=0.075\textwidth]{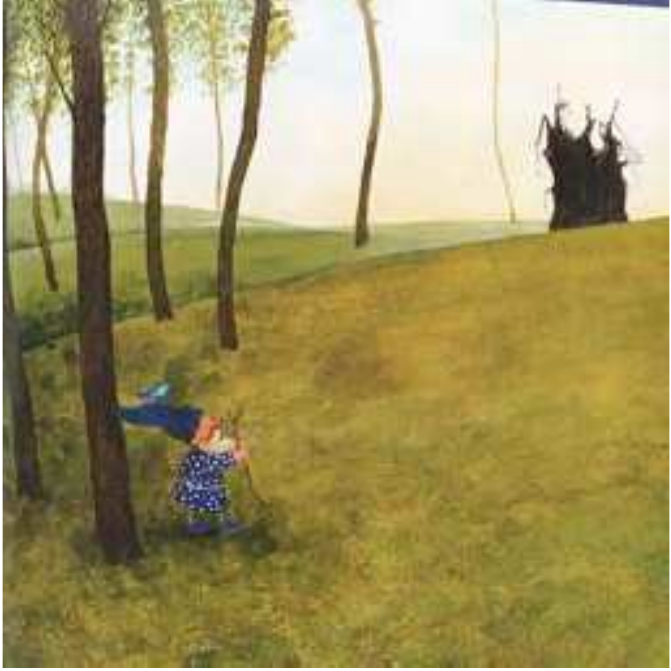}
\includegraphics[width=0.075\textwidth]{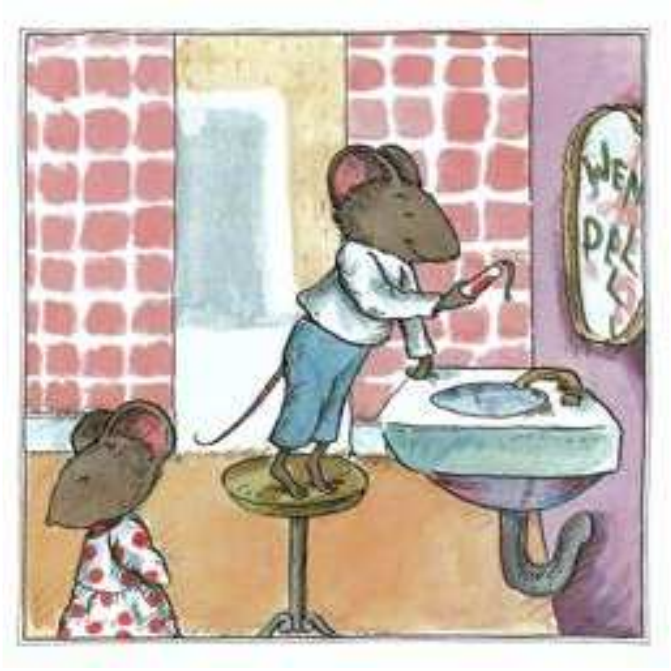} \\

\includegraphics[width=0.075\textwidth]{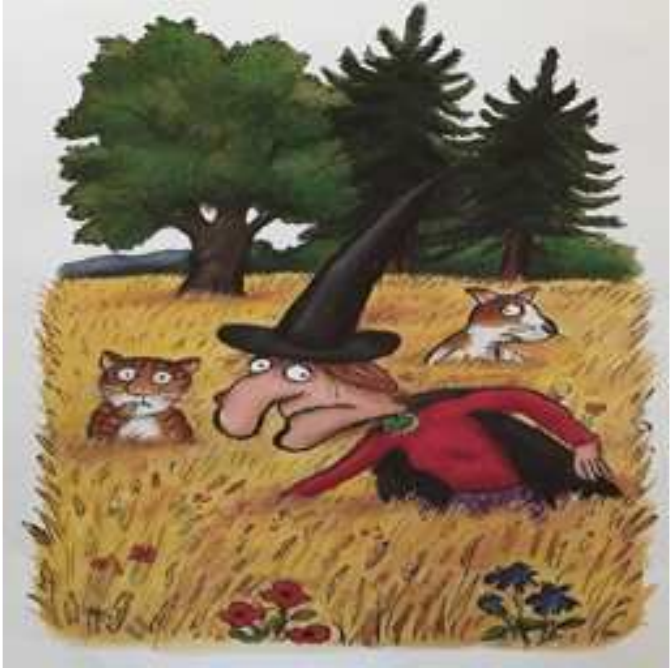}
\includegraphics[width=0.075\textwidth]{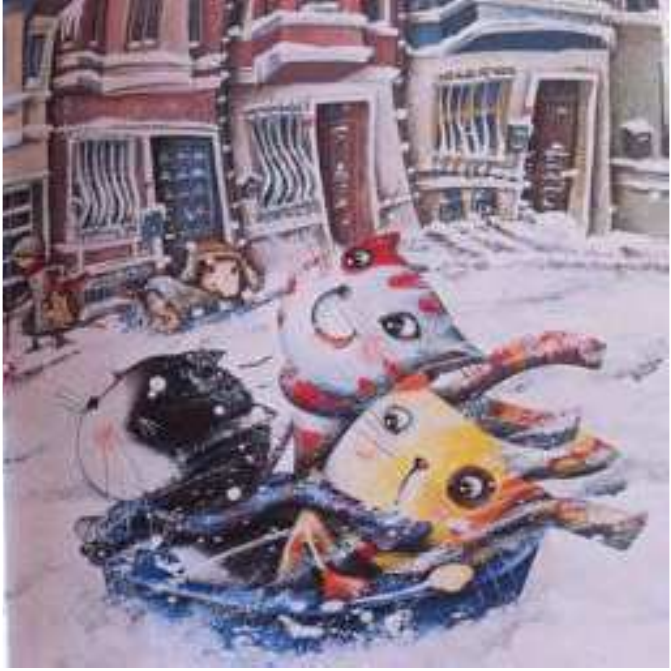}
\includegraphics[width=0.075\textwidth]{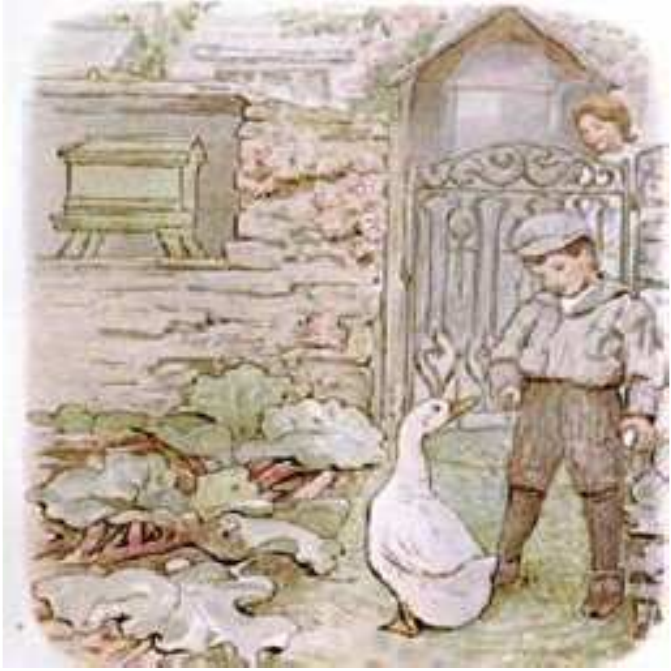}
\includegraphics[width=0.075\textwidth]{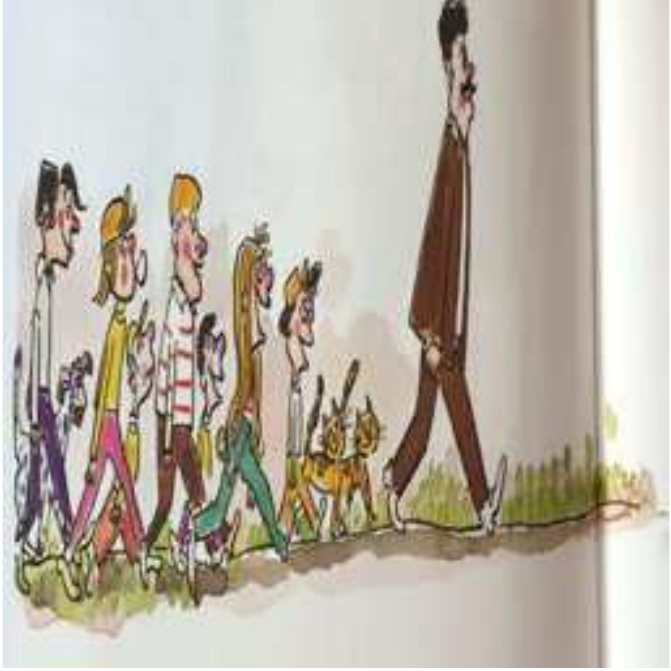}
\includegraphics[width=0.075\textwidth]{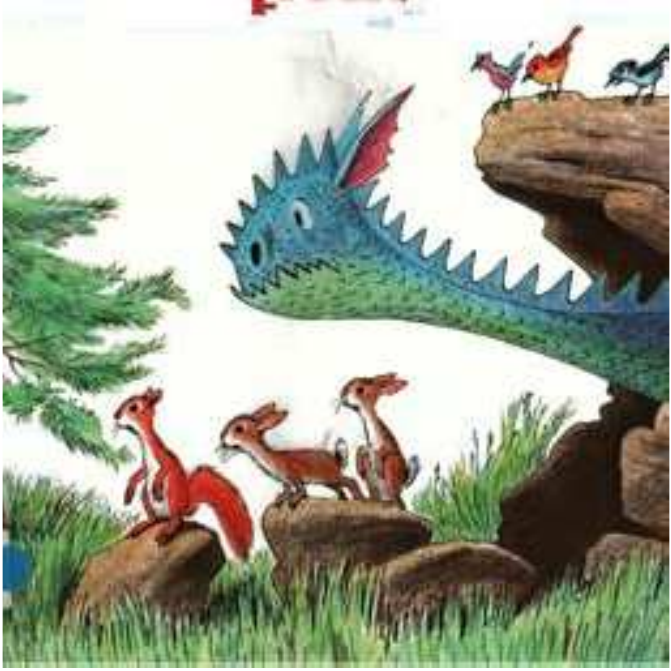}
\includegraphics[width=0.075\textwidth]{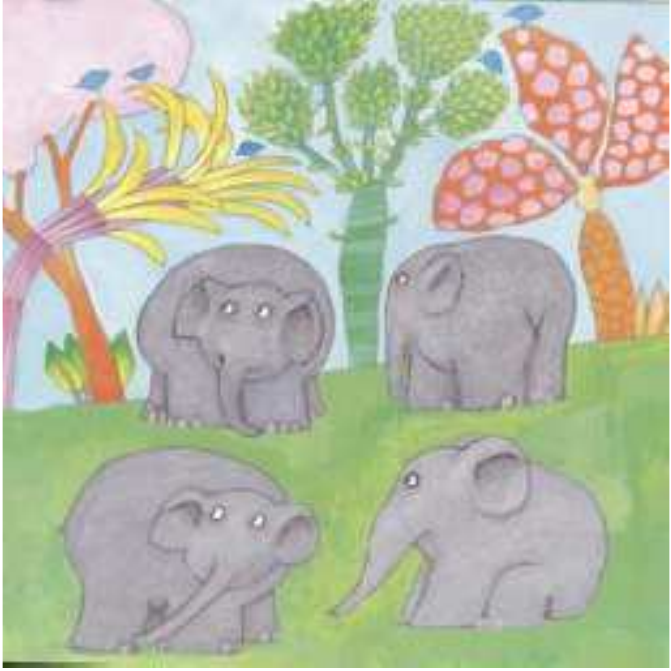}
\includegraphics[width=0.075\textwidth]{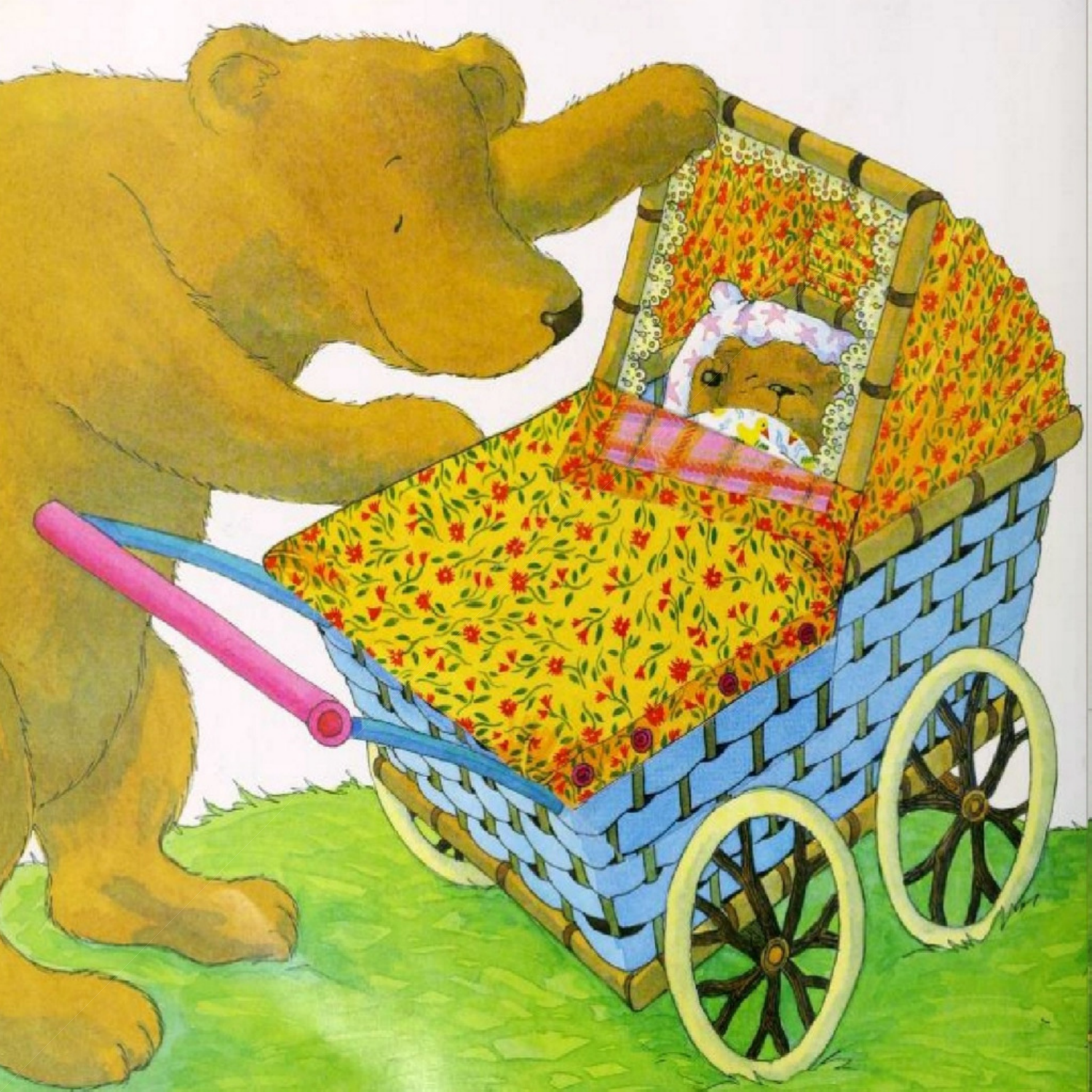}
\includegraphics[width=0.075\textwidth]{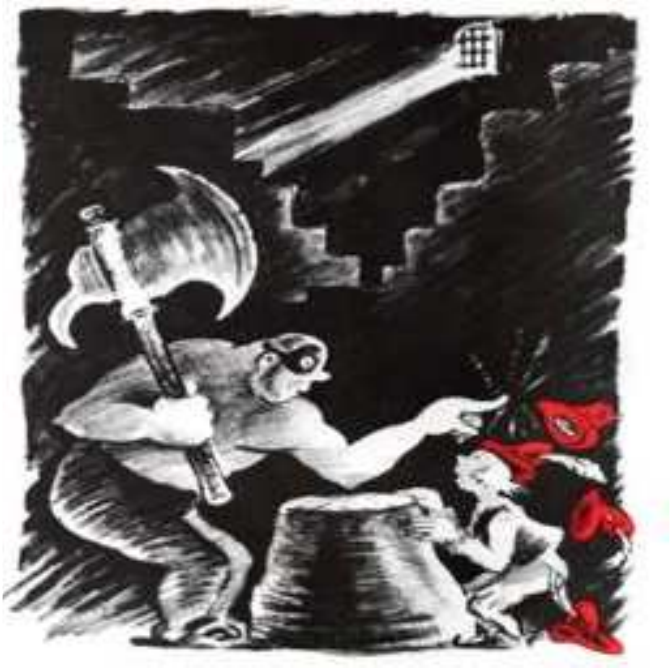}
\includegraphics[width=0.075\textwidth]{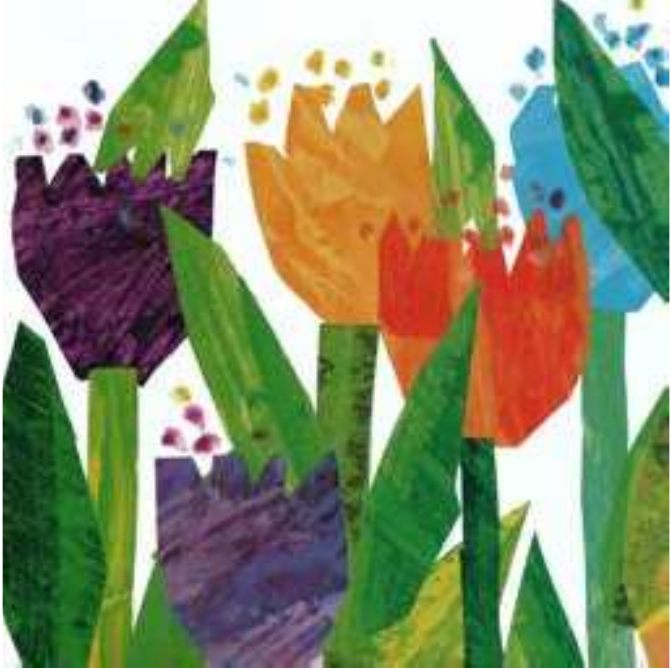}
\includegraphics[width=0.075\textwidth]{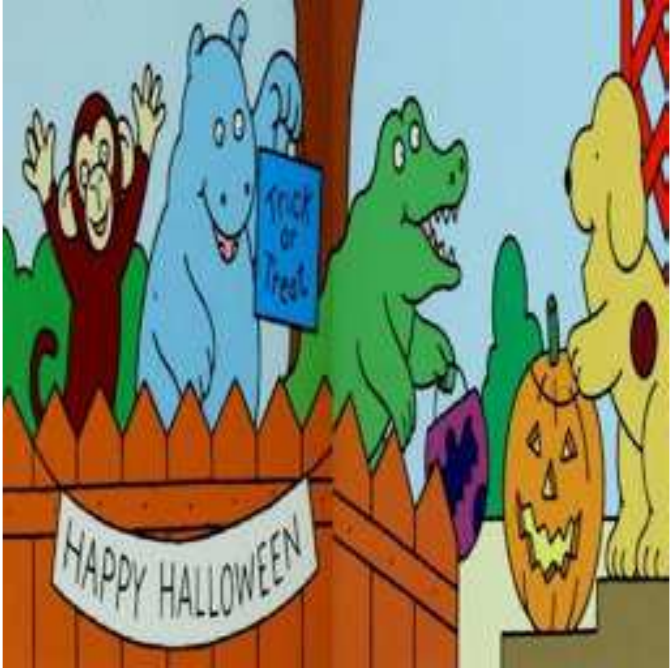}
\includegraphics[width=0.075\textwidth]{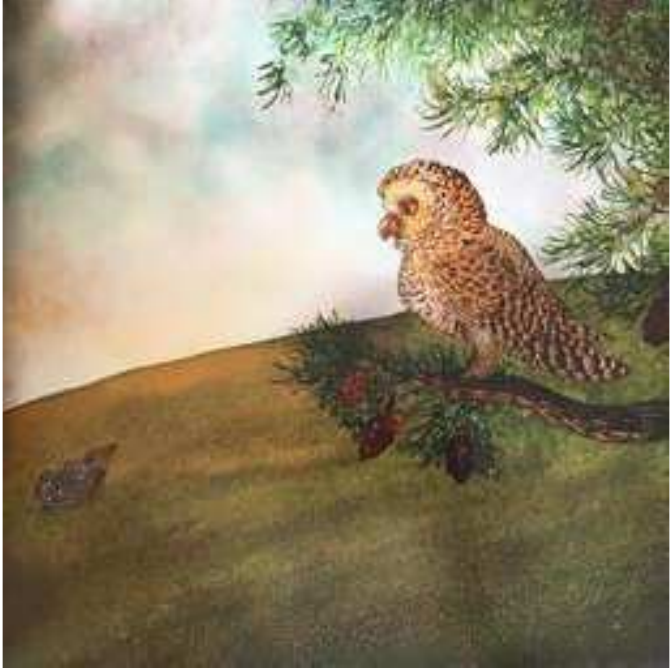}
\includegraphics[width=0.075\textwidth]{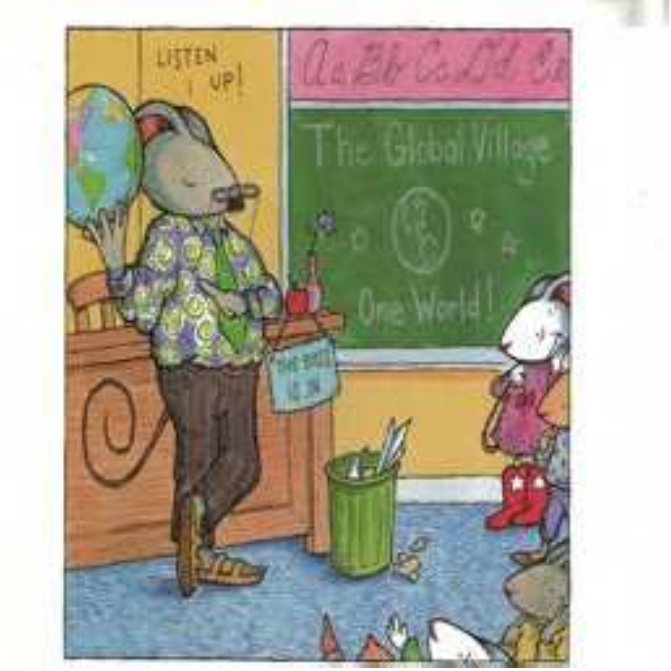} \\

\includegraphics[width=0.075\textwidth]{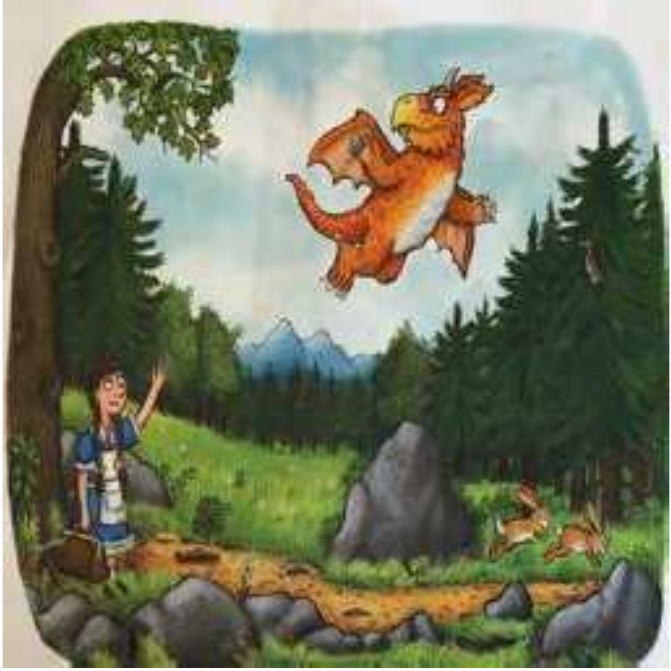}
\includegraphics[width=0.075\textwidth]{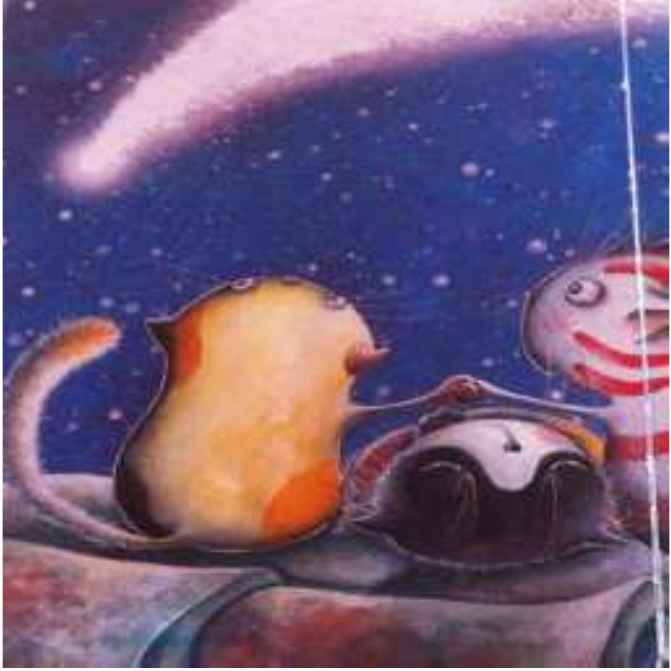}
\includegraphics[width=0.075\textwidth]{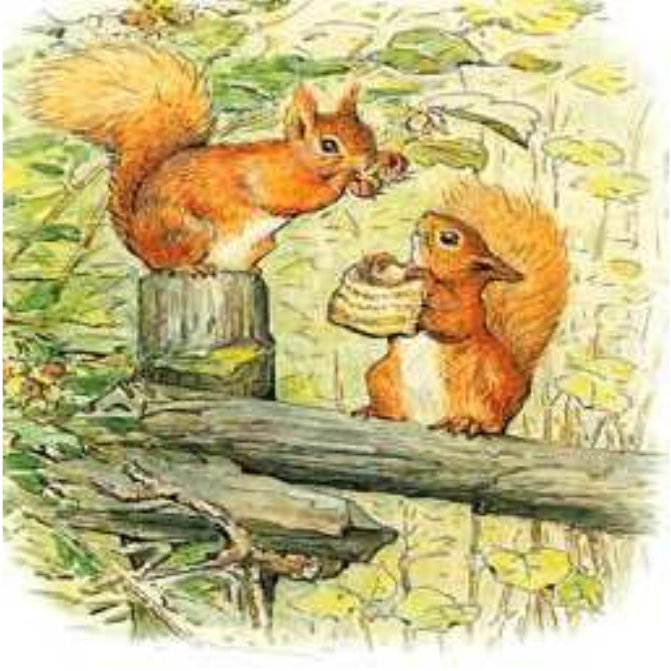}
\includegraphics[width=0.075\textwidth]{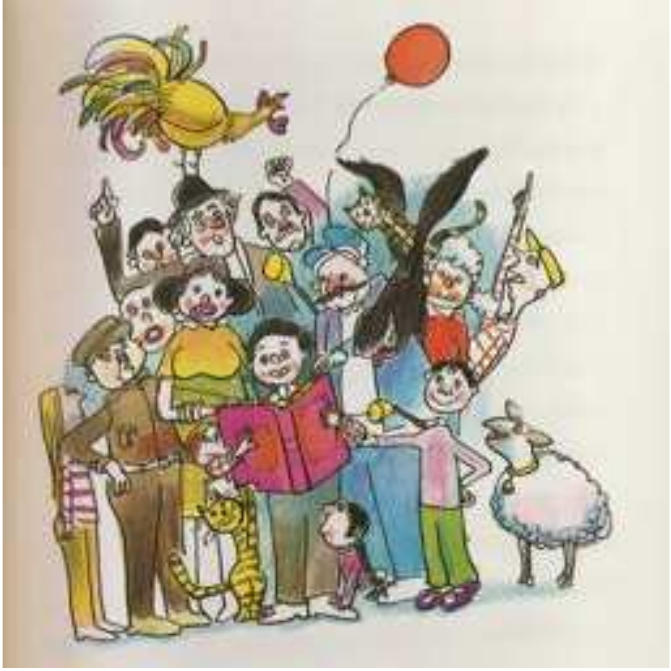}
\includegraphics[width=0.075\textwidth]{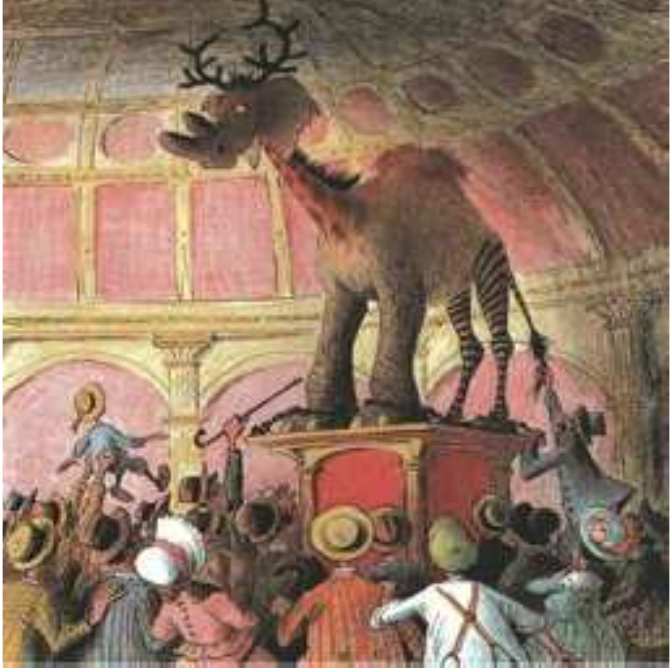}
\includegraphics[width=0.075\textwidth]{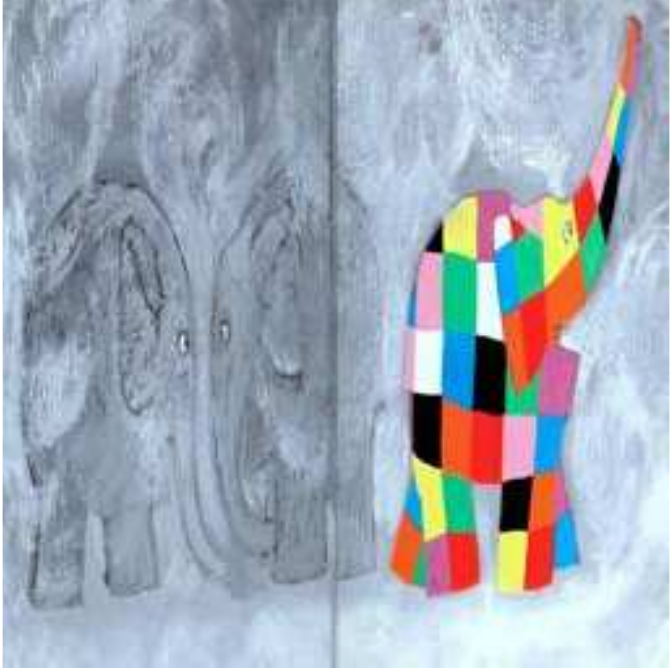}
\includegraphics[width=0.075\textwidth]{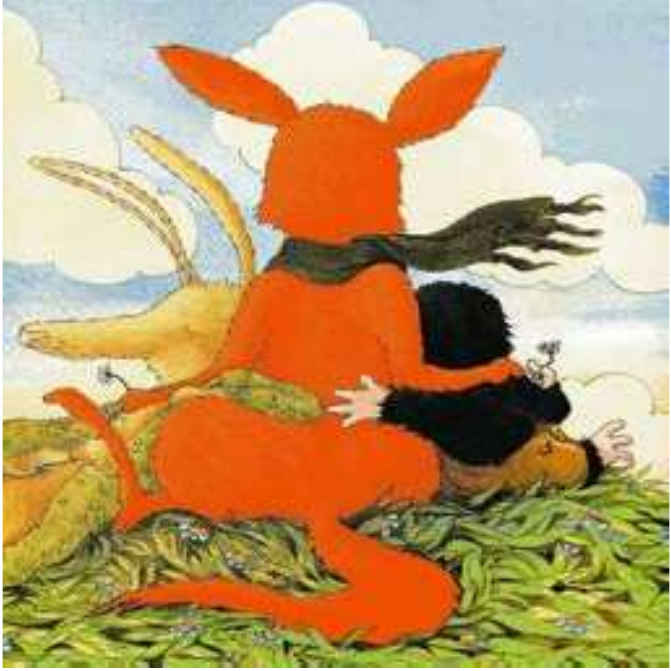}
\includegraphics[width=0.075\textwidth]{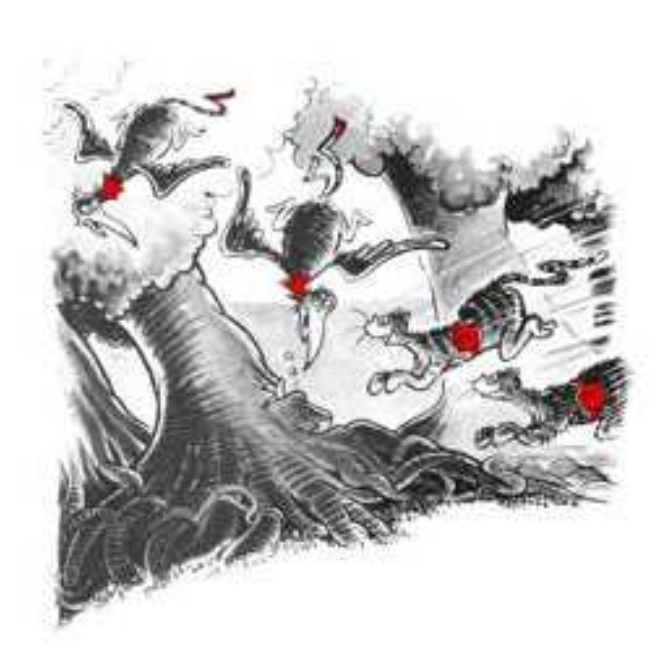}
\includegraphics[width=0.075\textwidth]{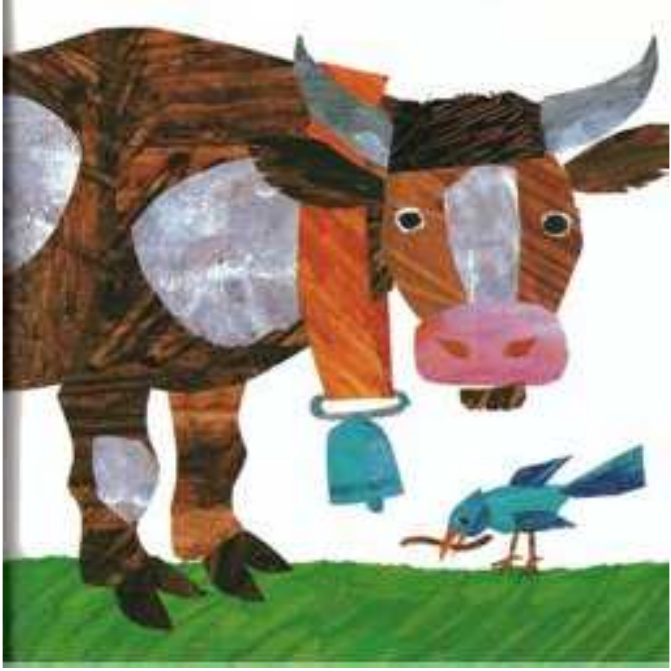}
\includegraphics[width=0.075\textwidth]{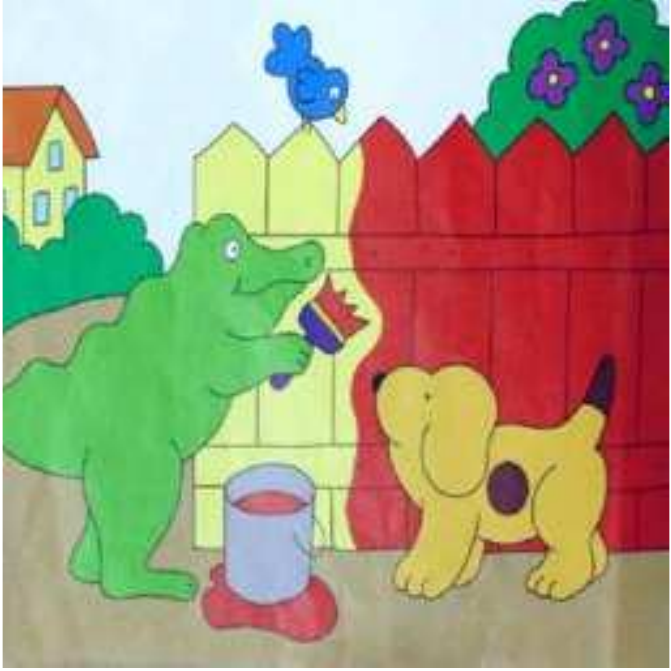}
\includegraphics[width=0.075\textwidth]{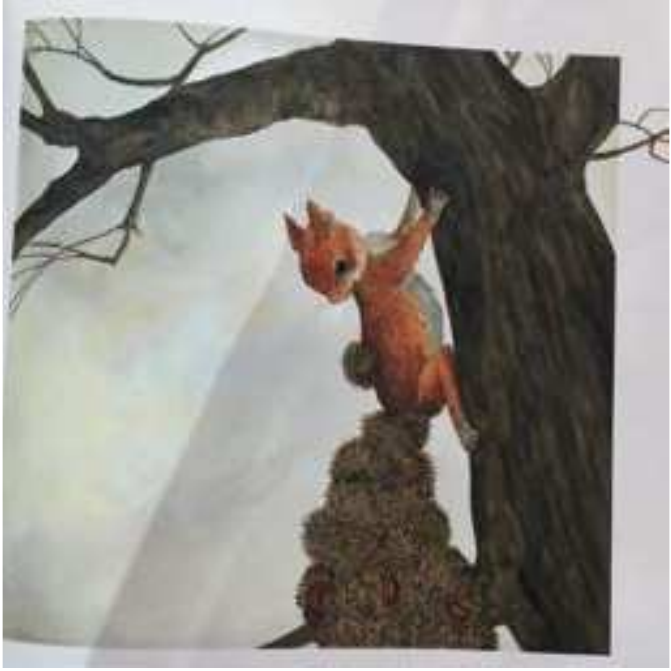}
\includegraphics[width=0.075\textwidth]{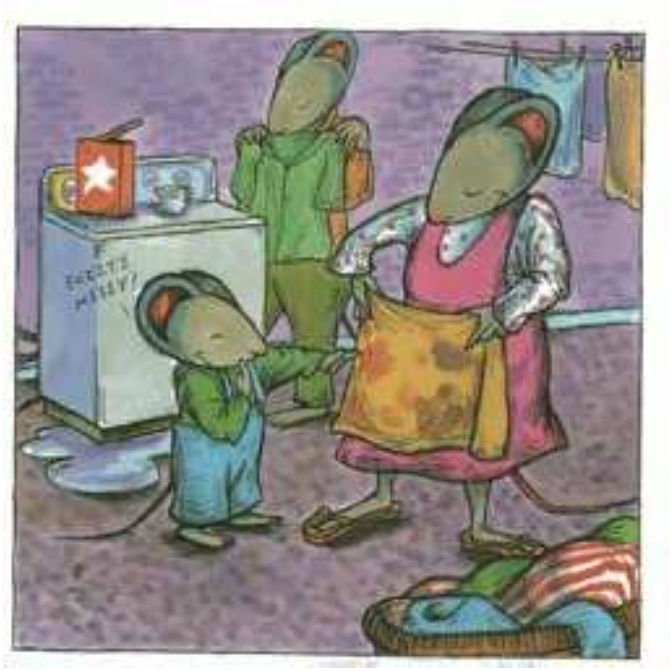} \\

\vspace{0.25cm} 

\includegraphics[width=0.075\textwidth]{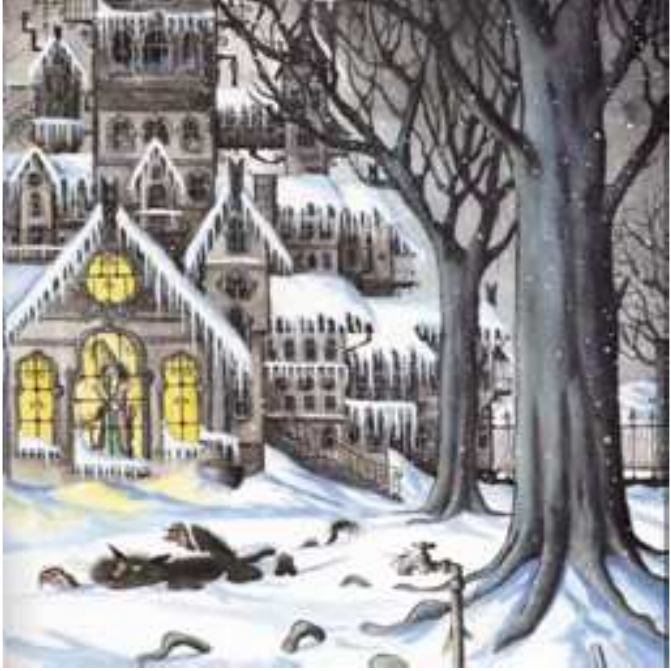}
\includegraphics[width=0.075\textwidth]{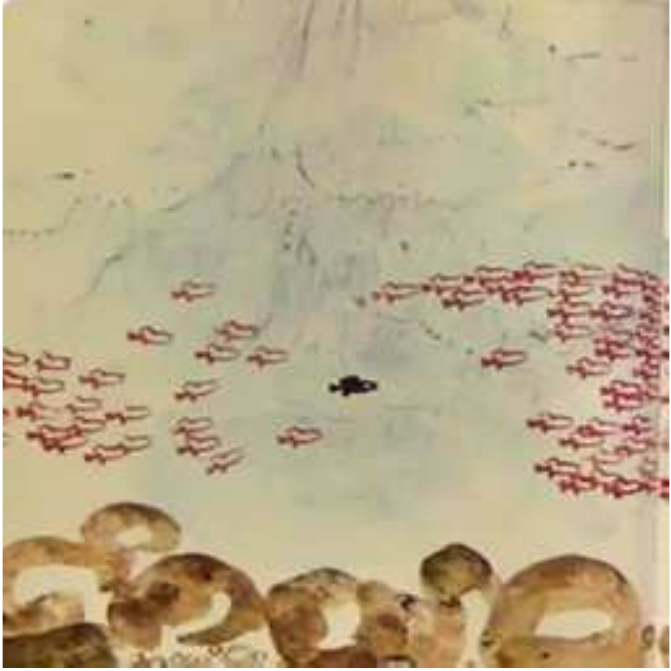} 
\includegraphics[width=0.075\textwidth]{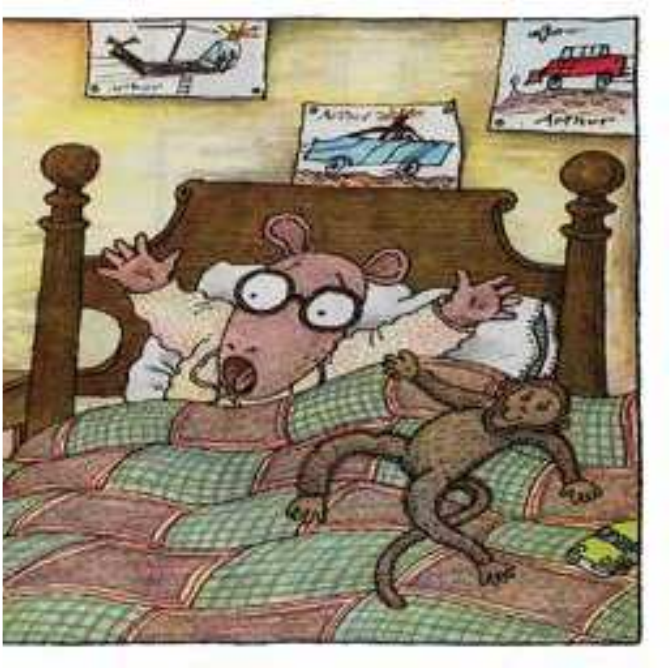} 
\includegraphics[width=0.075\textwidth]{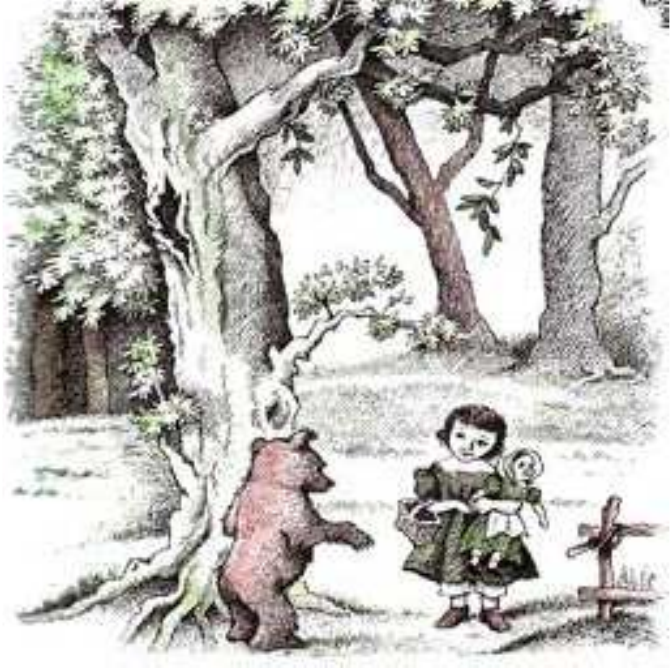} 
\includegraphics[width=0.075\textwidth]{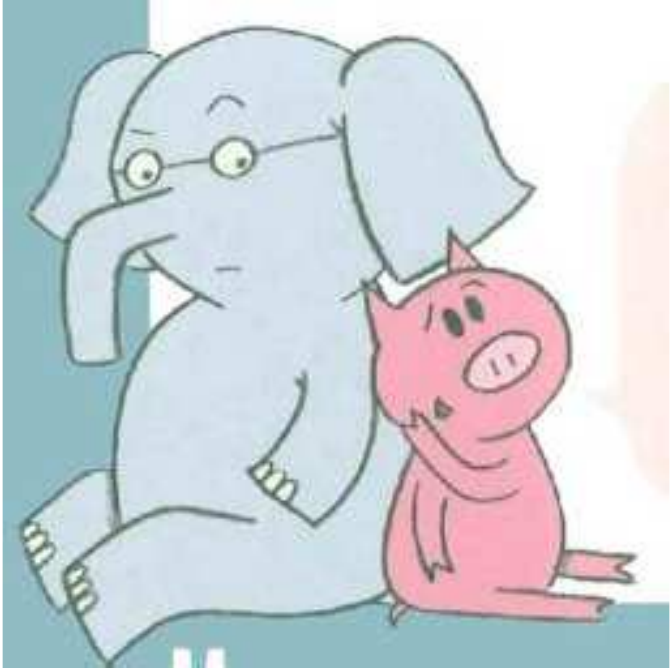} 
\includegraphics[width=0.075\textwidth]{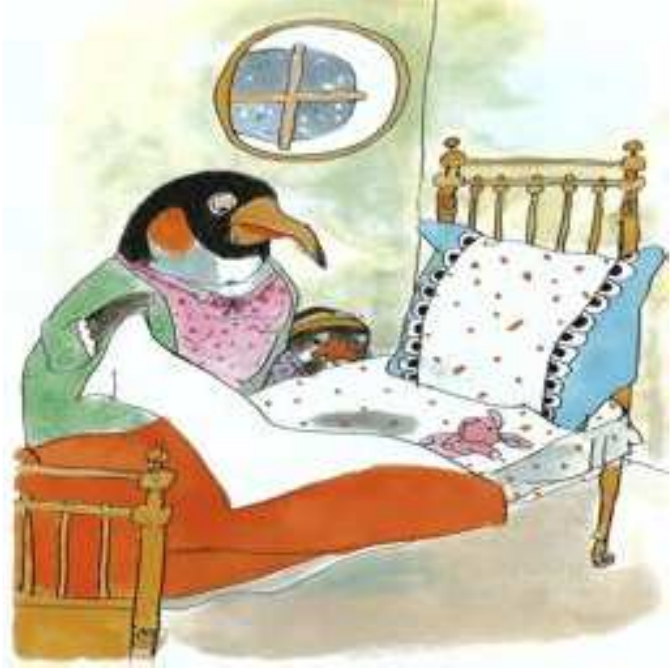} 
\includegraphics[width=0.075\textwidth]{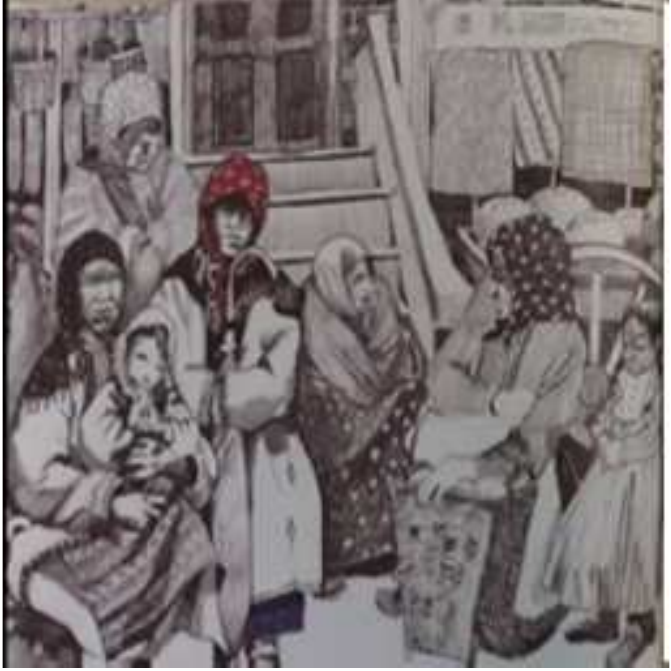} 
\includegraphics[width=0.075\textwidth]{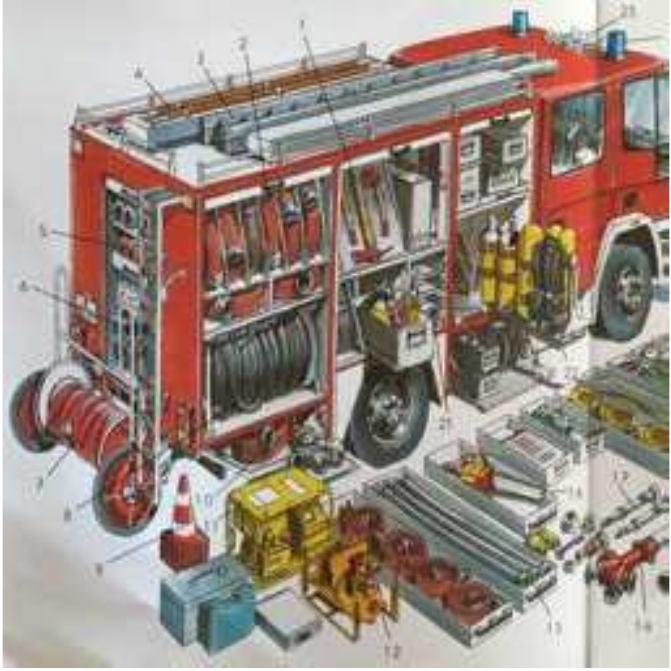}  
\includegraphics[width=0.075\textwidth]{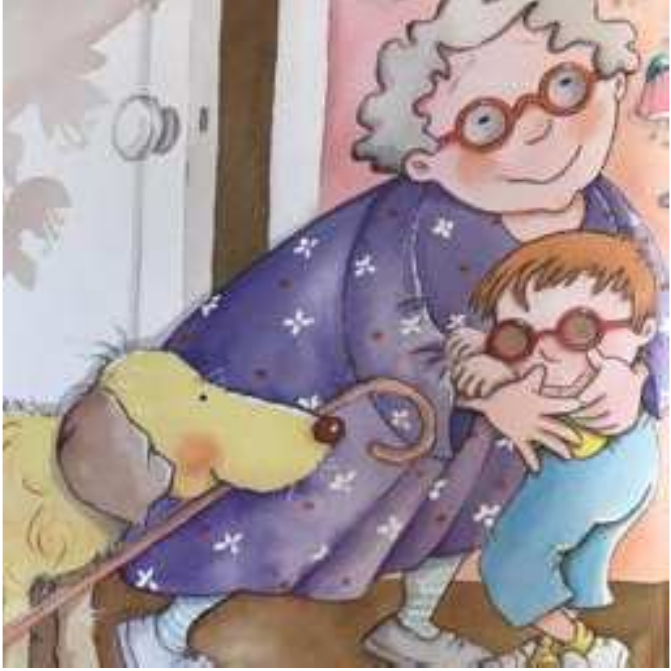} 
\includegraphics[width=0.075\textwidth]{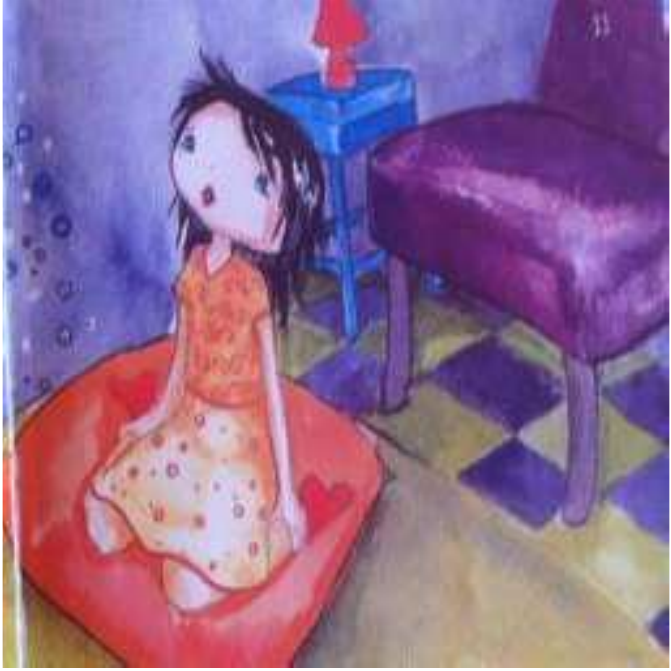} 
\includegraphics[width=0.075\textwidth]{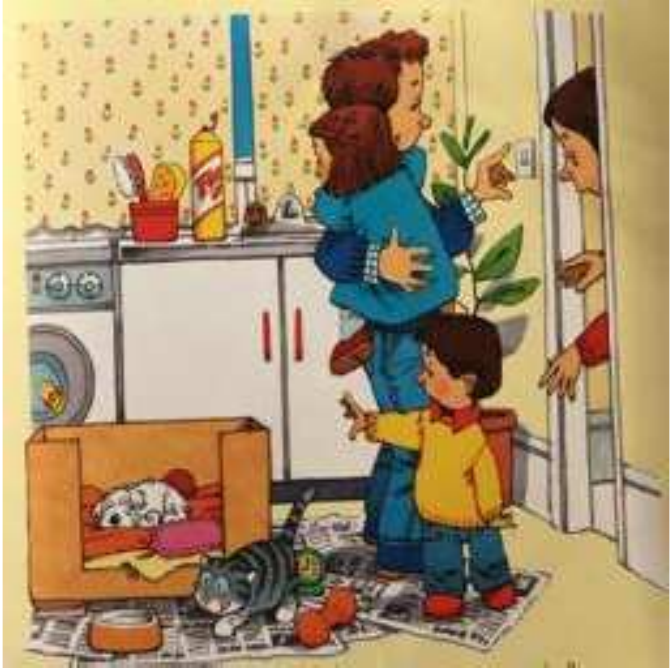} 
\includegraphics[width=0.075\textwidth]{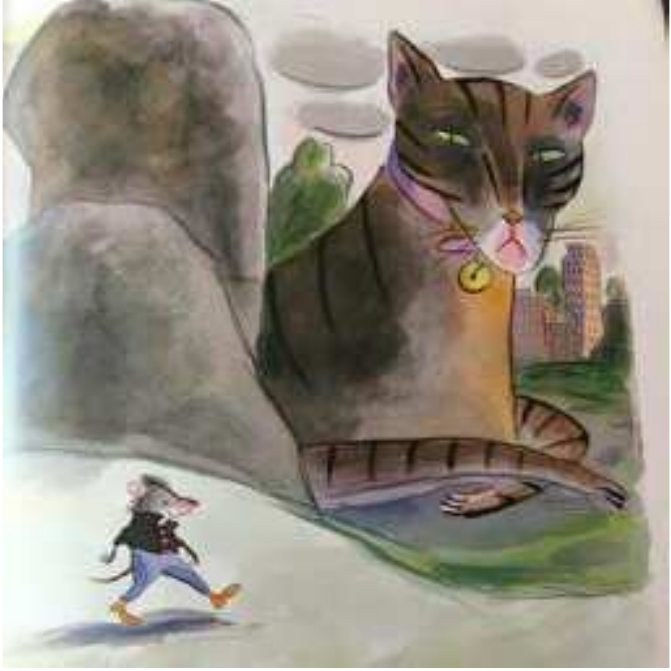} \\

\includegraphics[width=0.075\textwidth]{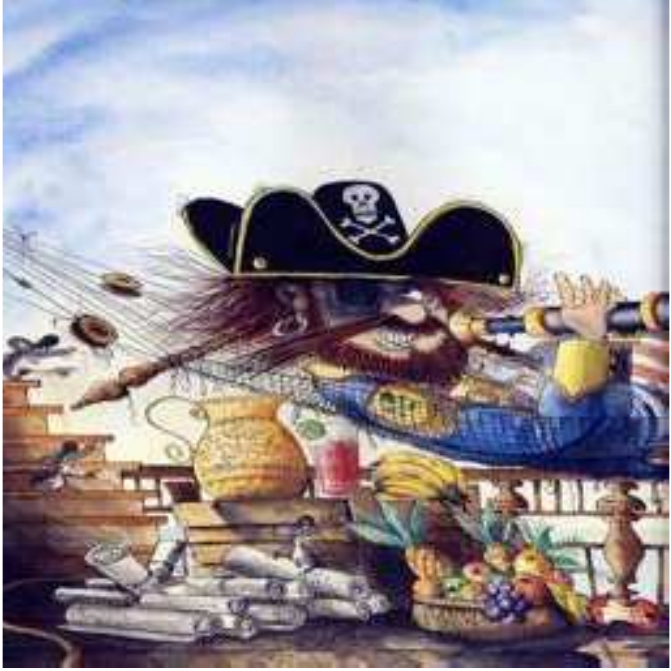} 
\includegraphics[width=0.075\textwidth]{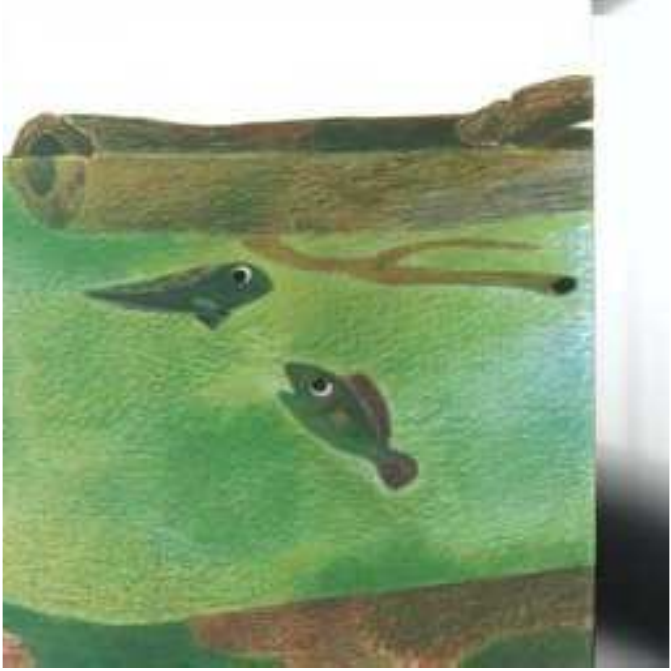} 
\includegraphics[width=0.075\textwidth]{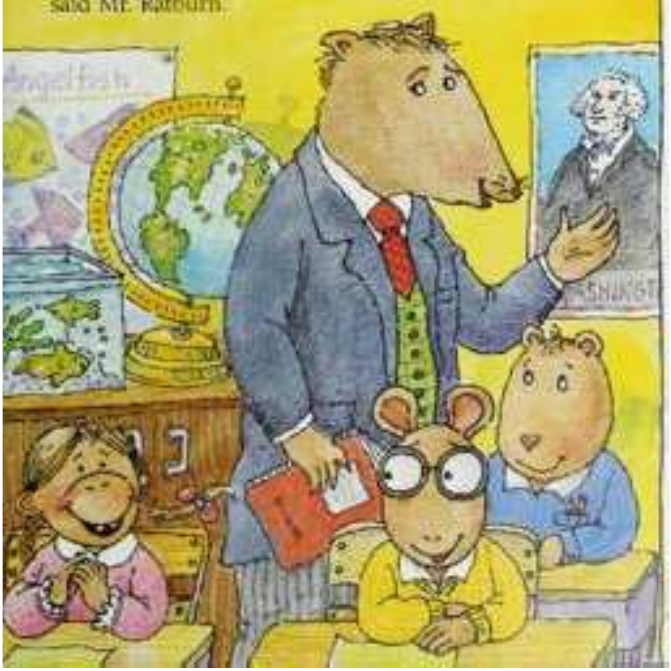} 
\includegraphics[width=0.075\textwidth]{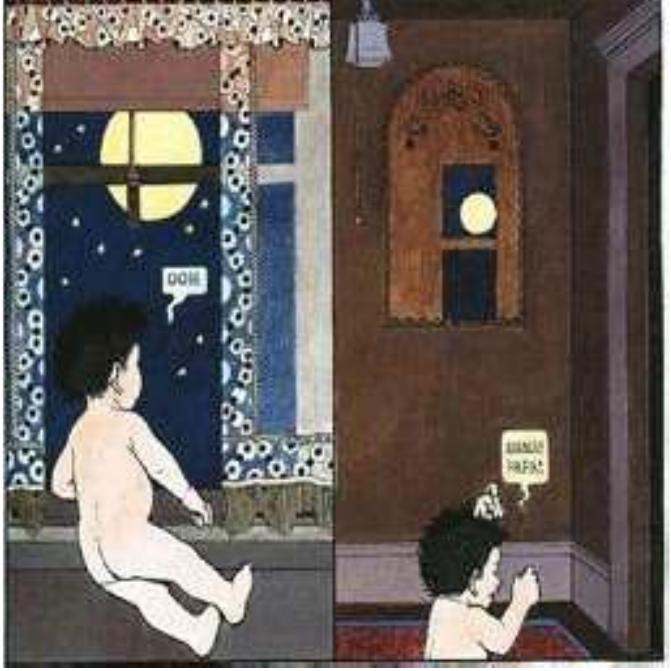} 
\includegraphics[width=0.075\textwidth]{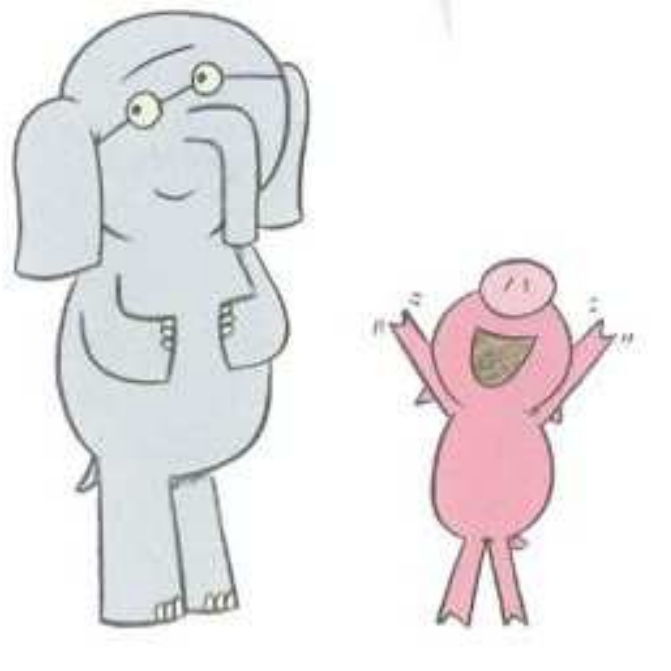} 
\includegraphics[width=0.075\textwidth]{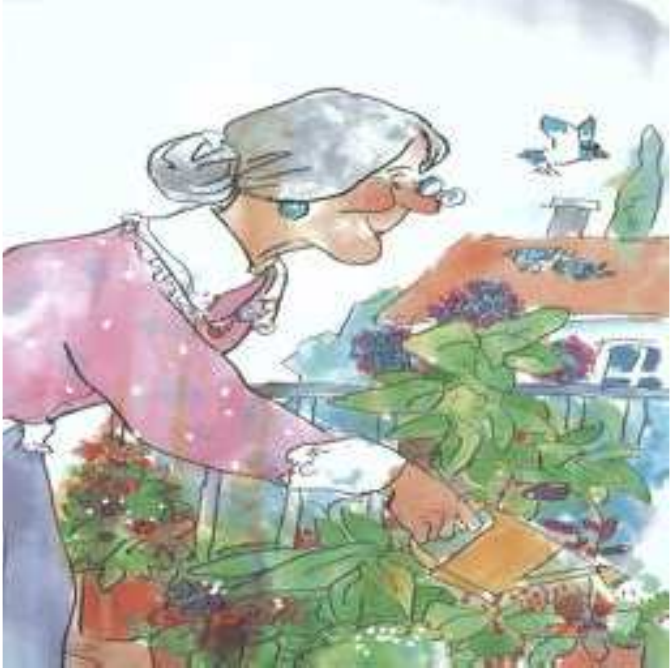} 
\includegraphics[width=0.075\textwidth]{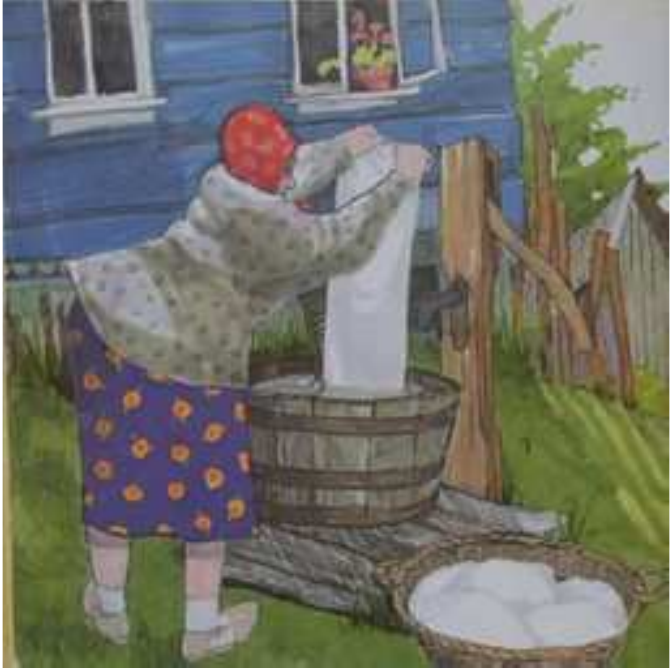} 
\includegraphics[width=0.075\textwidth]{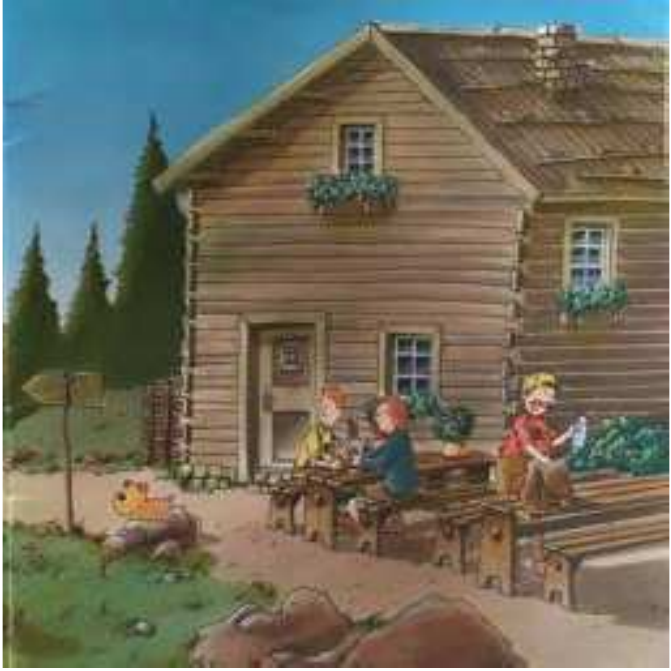}  
\includegraphics[width=0.075\textwidth]{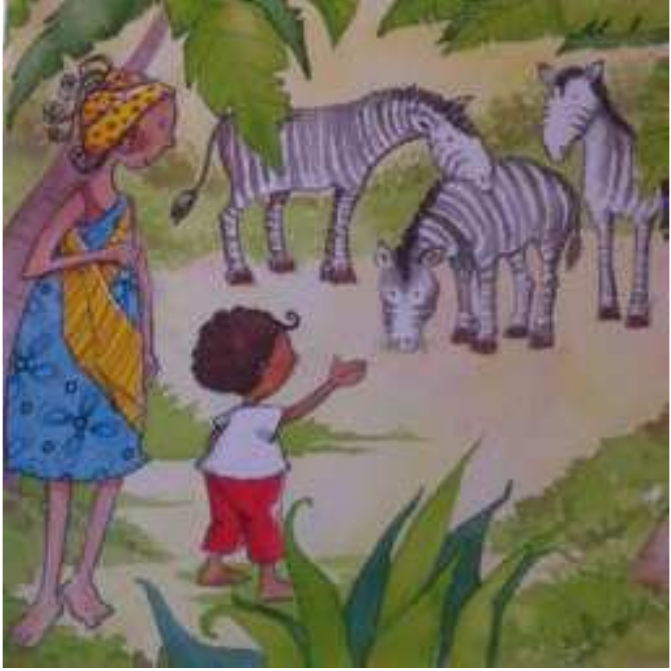} 
\includegraphics[width=0.075\textwidth]{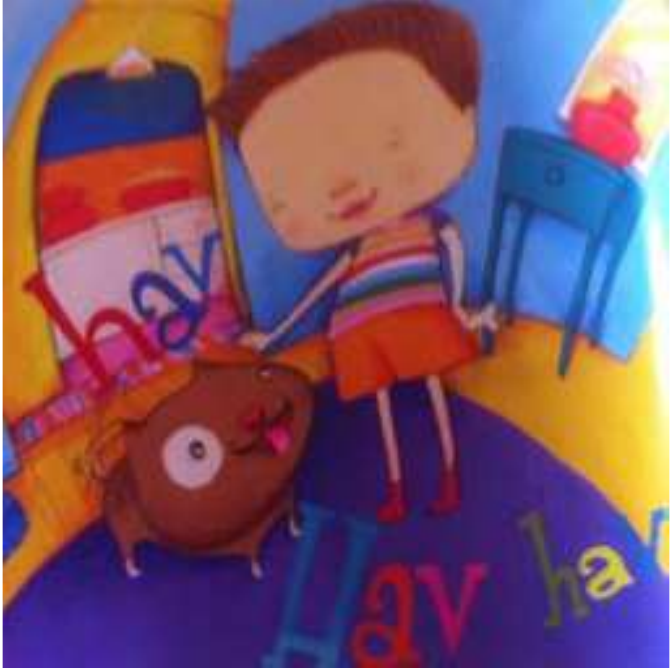} 
\includegraphics[width=0.075\textwidth]{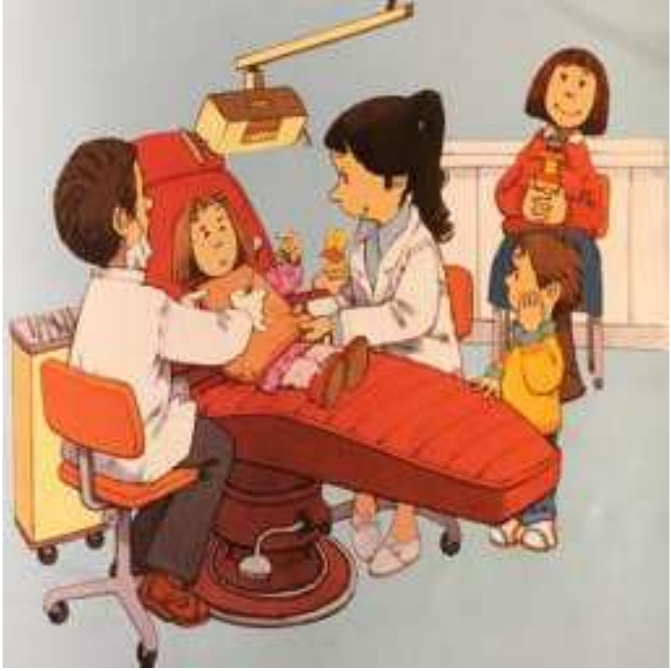} 
\includegraphics[width=0.075\textwidth]{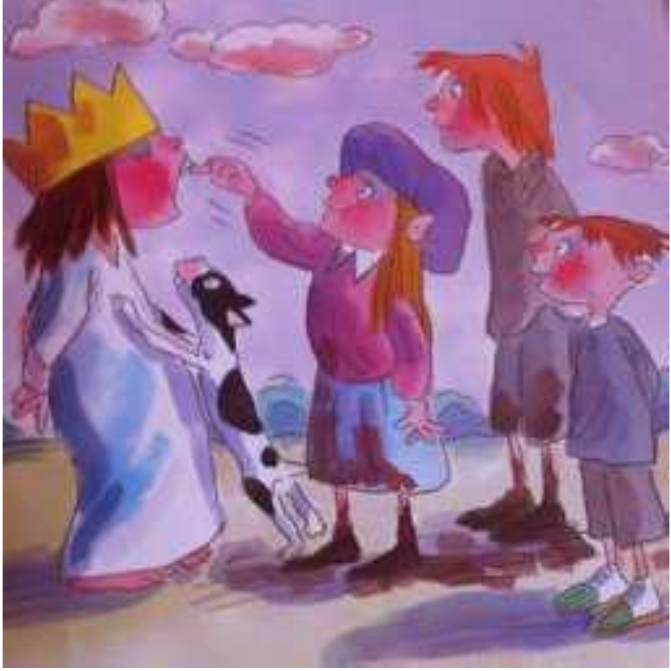} \\

\includegraphics[width=0.075\textwidth]{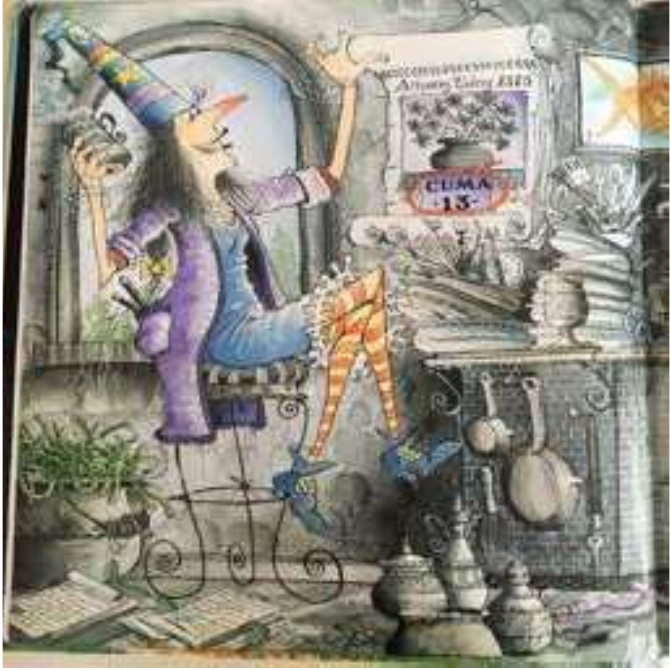} 
\includegraphics[width=0.075\textwidth]{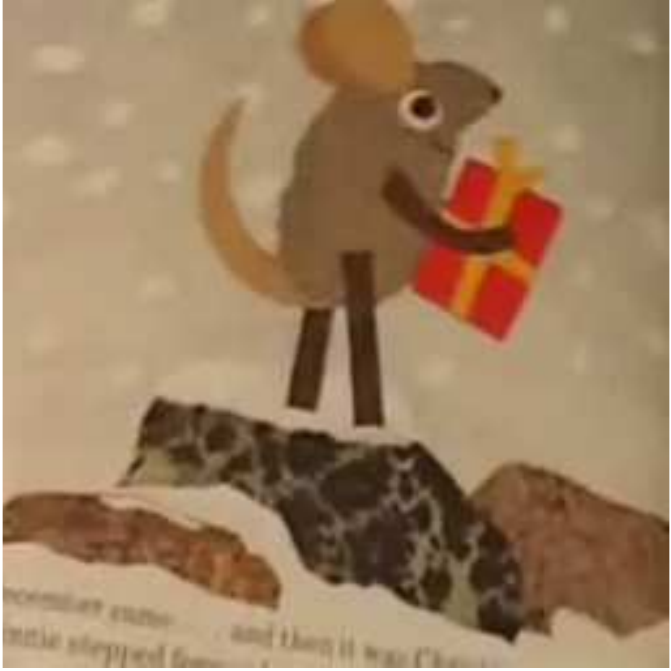} 
\includegraphics[width=0.075\textwidth]{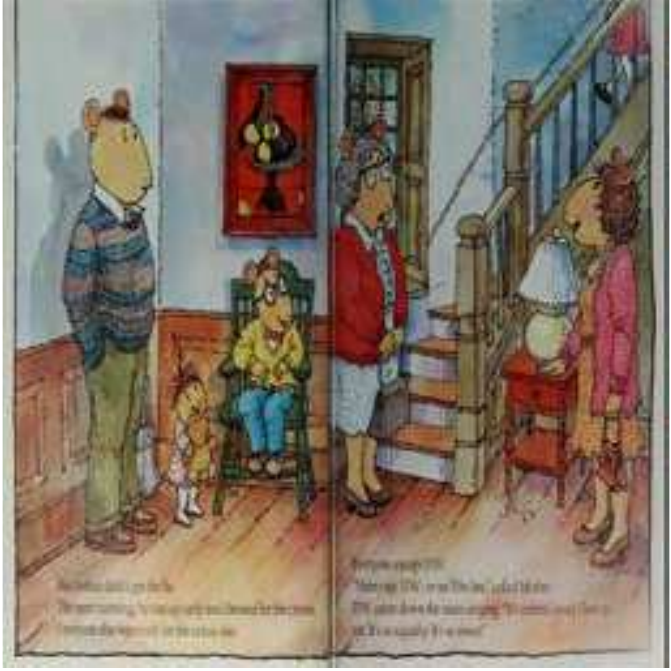} 
\includegraphics[width=0.075\textwidth]{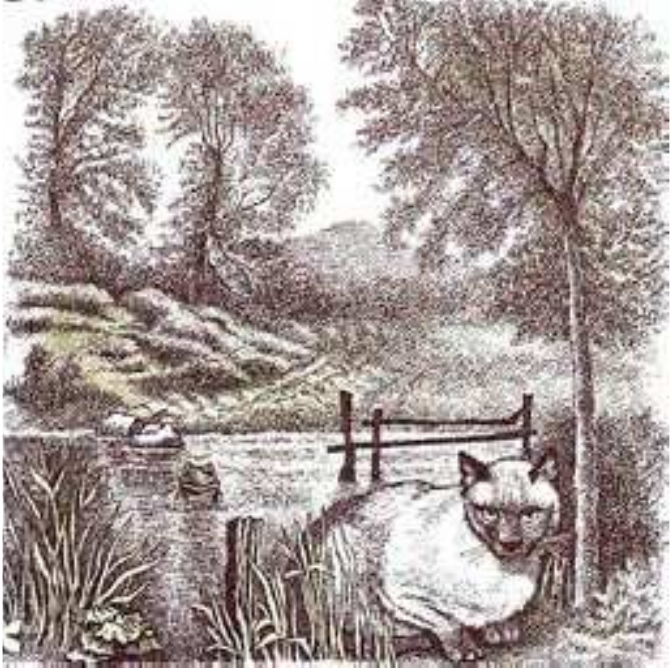} 
\includegraphics[width=0.075\textwidth]{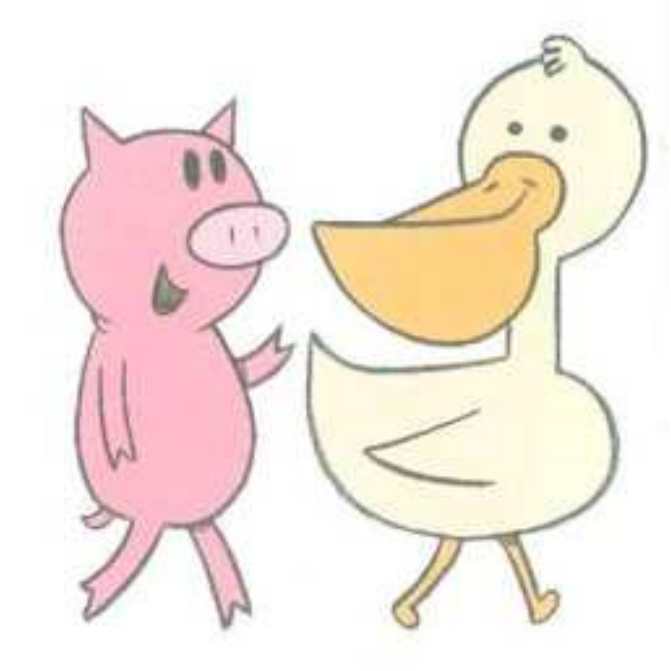} 
\includegraphics[width=0.075\textwidth]{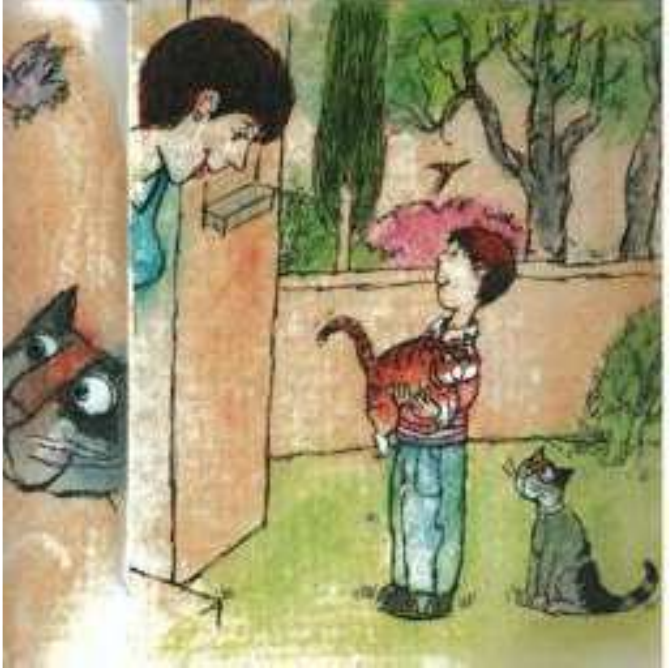} 
\includegraphics[width=0.075\textwidth]{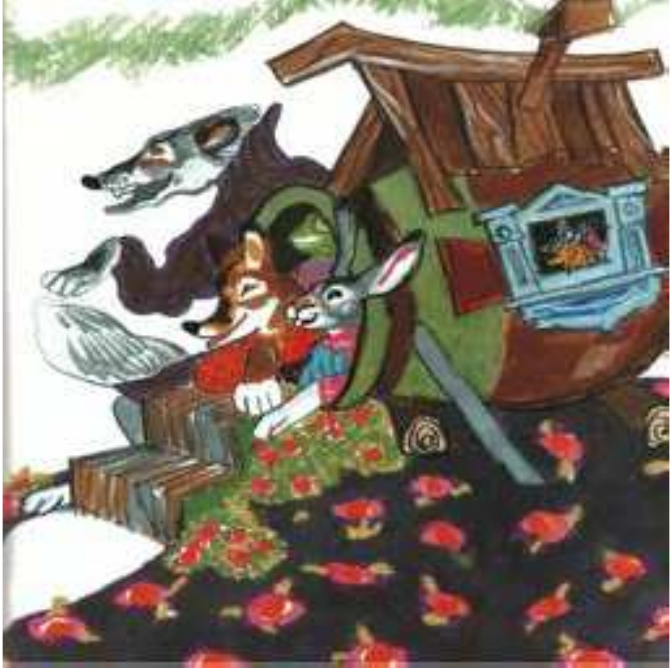} 
\includegraphics[width=0.075\textwidth]{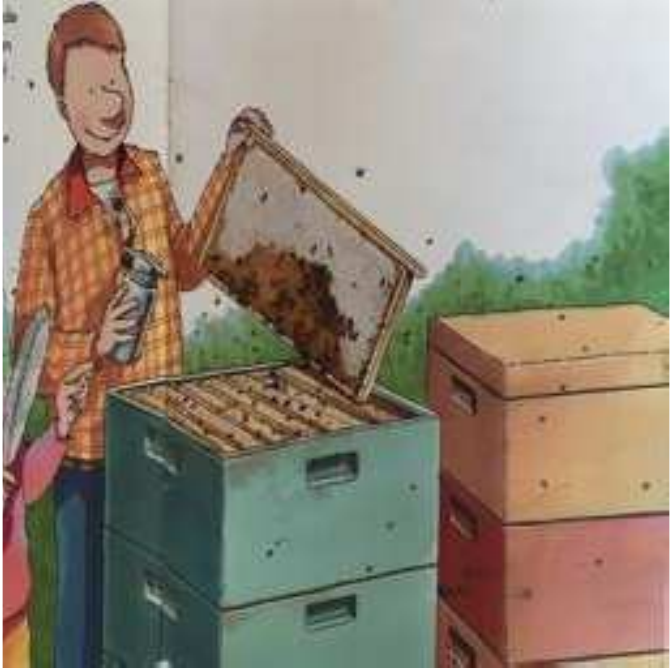}  
\includegraphics[width=0.075\textwidth]{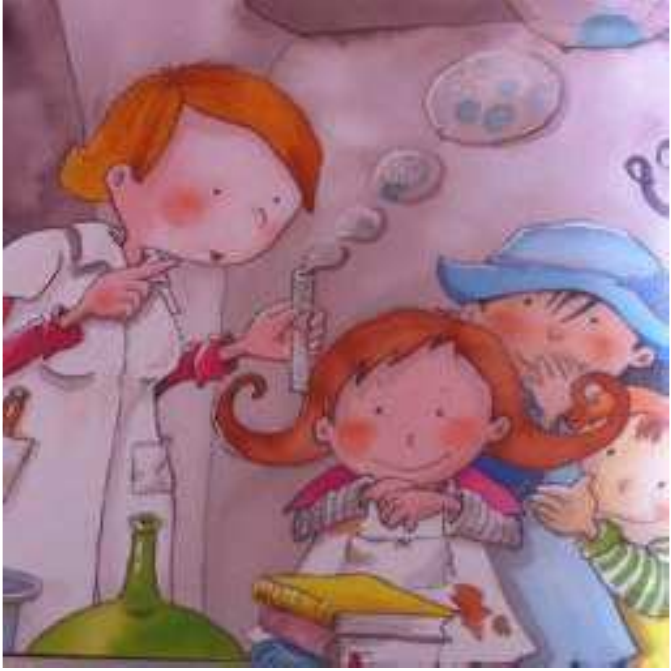} 
\includegraphics[width=0.075\textwidth]{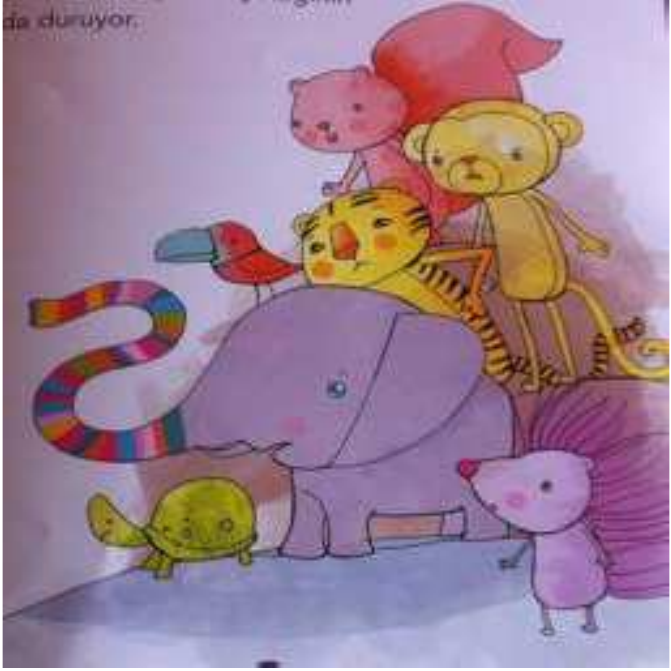} 
\includegraphics[width=0.075\textwidth]{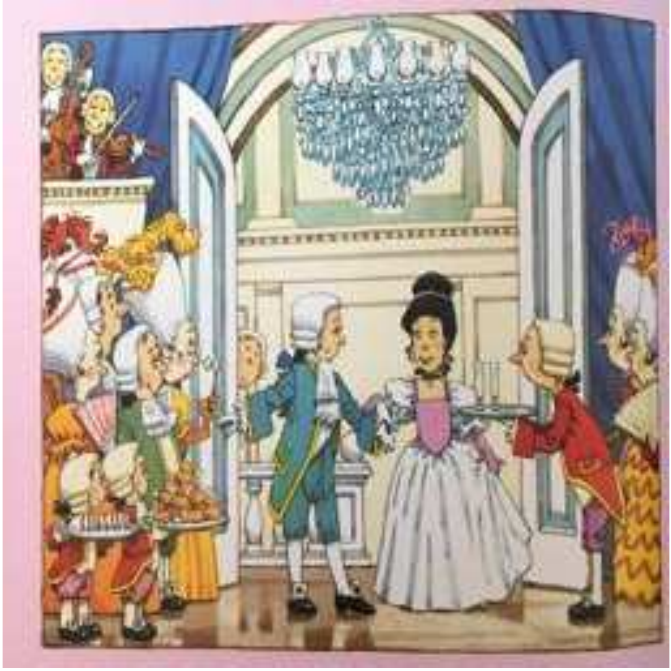} 
\includegraphics[width=0.075\textwidth]{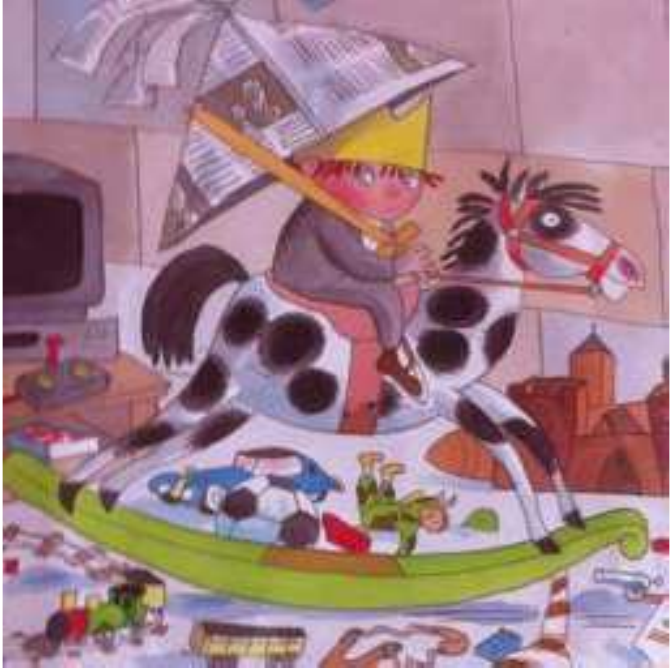} \\
%\end{tabular}
\captionof{figure}{Example illustrations from our data set. Three consecutive illustrations in a column correspond to a single illustrator. The order of illustrators is the same as in Table~\ref{tab:ill_data_set_info}. Note that, the styles are distinctive for illustrators. However, due to the variety in individual's styles some instances are difficult to categorise correctly. }
\label{fig:BookIllustrations}
\end{figure*}

\section{Dataset}
\label{sec:dataset}

We constructed a new data set consisting of 6468 distinct illustrations from 24 different illustrators. Focusing on the popular children's books, we mostly selected the illustrators who created more than a single basic character. The pages are collected either directly scanning from printed books or from publicly available e-books and read aloud videos over YouTube. 
%We pre-processed the gathered pages in order to get only the illustration by cropping and removing textual information. 
Table~\ref{tab:ill_data_set_info} shows summary of our dataset and Figure~\ref{fig:BookIllustrations} represents some example illustrations.  

In building the dataset, we are inspired from~\cite{sener2012identification} in which a dataset consisting of 248 illustrations of Axel Scheffler, 243 illustrations of
Debi Gliori, 249 illustrations of Korky Paul and 234 illustrations of Dr. Seuss was generated. We almost doubled the examples for three of the illustrators, and included 20 other illustrators.  Within its current form the dataset is unique: although large scale datasets exist for paintings~\cite{karayev2013recognizing, Crowley14, khan2014painting}, based on our knowledge this is the first comprehensive dataset for illustrations. \\

Note that, in the painting datasets there are a variety of artists following the same artistic style, and thus the dataset is deep in the sense that the number of examples per style is large. However, each illustrator has only a limited set of books and therefore the number of examples per category is not possible to reach to the numbers in painting datasets. Similarly, the number of categories can only be extended within some limits when we force each illustrator to have more than a single specific character or book series. We continue to extend the dataset and will make it publicly available within the copyright limitations. 
%---------------------------------------------------------------------------------------------------------------------------Methodology

\section{Discovering style of illustrators}
\label{sec:method}
In the following, we will first describe the details of our method in categorizing the style of illustrators using deep networks. Then, we will discuss about approaches to transfer style and to discover representative elements.

\subsection{Deep Learning For Style Recognition}
\label{sec:deep_lr}

Instead of creating a model from scratch, we used three well-known CNN models in training:  AlexNet~\cite{alexnet}, VGG-19~\cite{vgg19} and GoogLeNet~\cite{googlenet}. We used Caffe~\cite{caffe} framework to train deep networks on a Tesla K40 12GB GPU. We employed both end-to-end training and transfer learning. To  train an end-to-end model, we enlarged our data set which is comparably small, by applying data augmentation. 

For small data sets like ours, it is not practical and meaningful to fully train very deep networks. Thus, we fully trained only the AlexNet as being relatively shallow. We first subtracted the mean of RGB values over our illustrations dataset for each pixel and obtained the centered raw RGB values. We augmented our training and validation data using only horizontal reflections to reduce overfitting. The batch sizes are chosen as 128 and 40 for train and validation respectively. Base learning rate is set to 0.01 with a momentum of 0.9 and the learning rate is decreased by a factor of 10 after each 40K iterations. 
%Around 100K iterations loss converged to a fixed value. 

Considering the fact that our dataset is comparably small, alternatively we applied transfer learning. For this purpose, we used VGG-19, AlexNet and GoogLeNet models pre-trained on a large scale ImageNet dataset. 
Our hyper parameters are nearly the same for fine tuning on AlexNet and VGG-19 except learning rate and batch sizes. Due to the memory issues, we were able to train VGG-19 with train batch sizes of 32. We selected learning rate accordingly and set it to 0.0004. Base learning rate for AlexNet is 0.0001 and all other parameters for SGD are same as end-to-end training.  
%Both networks run for 5K iterations and validation loss and accuracy converged after 1K iterations. 
To train GoogLeNet we used quick solver~\cite{quick_solver} properties with initial learning rate of 0.01.

\subsection{Style transfer}
Inspired by the recent work on transfering artistic style of paintings~\cite{style_transfer_2}, we transfer the style from one illustration to another. Besides showing the ability of style generation, this task is also important to understand the capability of deep models to capture the style separated from the content.

Style transfer model~\cite{style_transfer_2} combines the appearance of a style image, e.g. an artwork,  with the content of another image, e.g. an arbitrary photograph,  by minimizing the loss of content and style. In our case, style is learned from an illustration of a particular illustrator, and transferred to another image. The target image could be a cartoon, a natural photograph, or another illustration from another artist. We expect the resulting image to contain the content of the target image drawn with the style of source illustration.

However, it is difficult and subjective to judge the quality of the resulting images. In this study, focusing on the style transfers from one illustration to another, we propose to compare the style of the resulting illustration with the original style from the categorisation perspective.  Our intuition here is that if we use the resulting image as a test instance on our deep networks, and classify them correctly then we could infer that deep models can capture styles.

\subsection{Discovering representatives}
\label{sec:disc_patch}

Here we try to understand style of illustrators in terms of discriminative and representative examples. We utilised two methods for this purpose. The first method~\cite{paris} was initially proposed for discovering architectural elements of different cities. It takes a positive set of images from which we want to extract discriminative patches, and a global negative set. It uses HOG features~\cite{dalal2005histograms} to represent the images. We have used this method both to find representative illustrations for different artists and also for discovering the discriminative parts in the illustrations. However, since this algorithm takes days to complete on a powerful laptop, we were able to run it only for a few of illustrators.

The second method that we utilised~\cite{fame} focuses on eliminating the outliers from a candidate set of positive examples to capture the representative elements in an iterative fashion. The method was proposed to recognise faces from noisy weakly labeled images collected from web. Being flexible, we exploited this method with HOG~\cite{dalal2005histograms},  color dense SIFT~\cite{sift}, and VGG~\cite{vgg19} features. 

%---------------------------------------------------------------------------------------------------------------------------Experiments
\section{Experiments}
\label{sec:exp}

In this section, we first present detailed experimental evaluations to recognize style of illustrators using deep networks. We also provide experimental results of conventional classification methods as a baseline to compare with deep architectures.  Then, we present our results on style transfer and representative element discovery. 

\subsection{Style recognition with deep networks}
\label{sec:deep_lr_eval}
We used two different settings for categorisation.  In the first setting, we treated each page as an independent instance and constructed training, validation and test sets by randomly selecting instances from the entire collection. In the second setting, we tested a more challenging case, and removed some of the books entirely from the training set. Results of both settings will be discussed in the following.

\begin{figure*}[t]
\begin{center}
\centering
\begin{tabular}{cccccc}
%%%%% 1st row 
\includegraphics[width=0.12\textwidth]{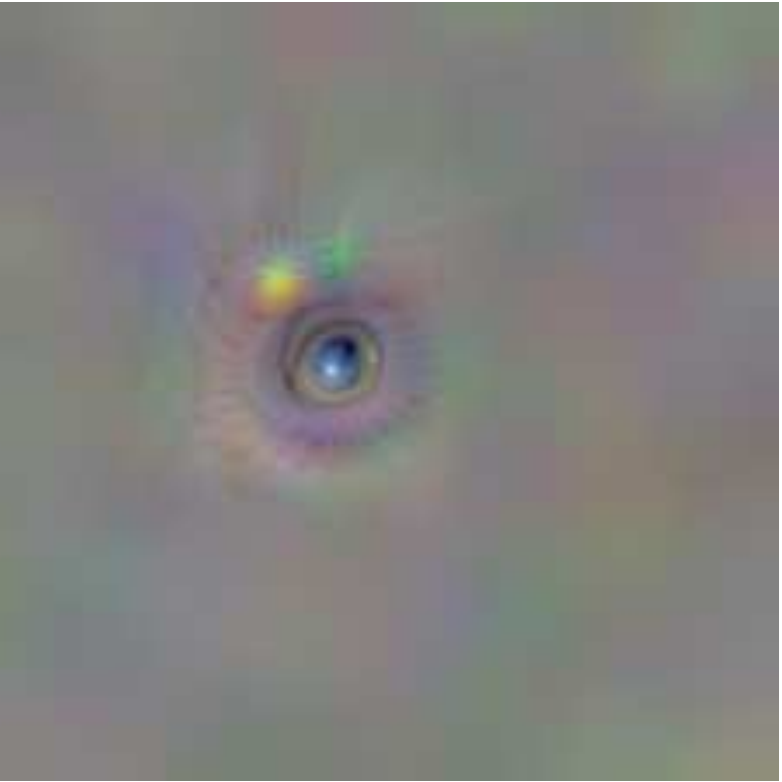} 
 \hspace*{0.001cm}
\includegraphics[width=0.12\textwidth]{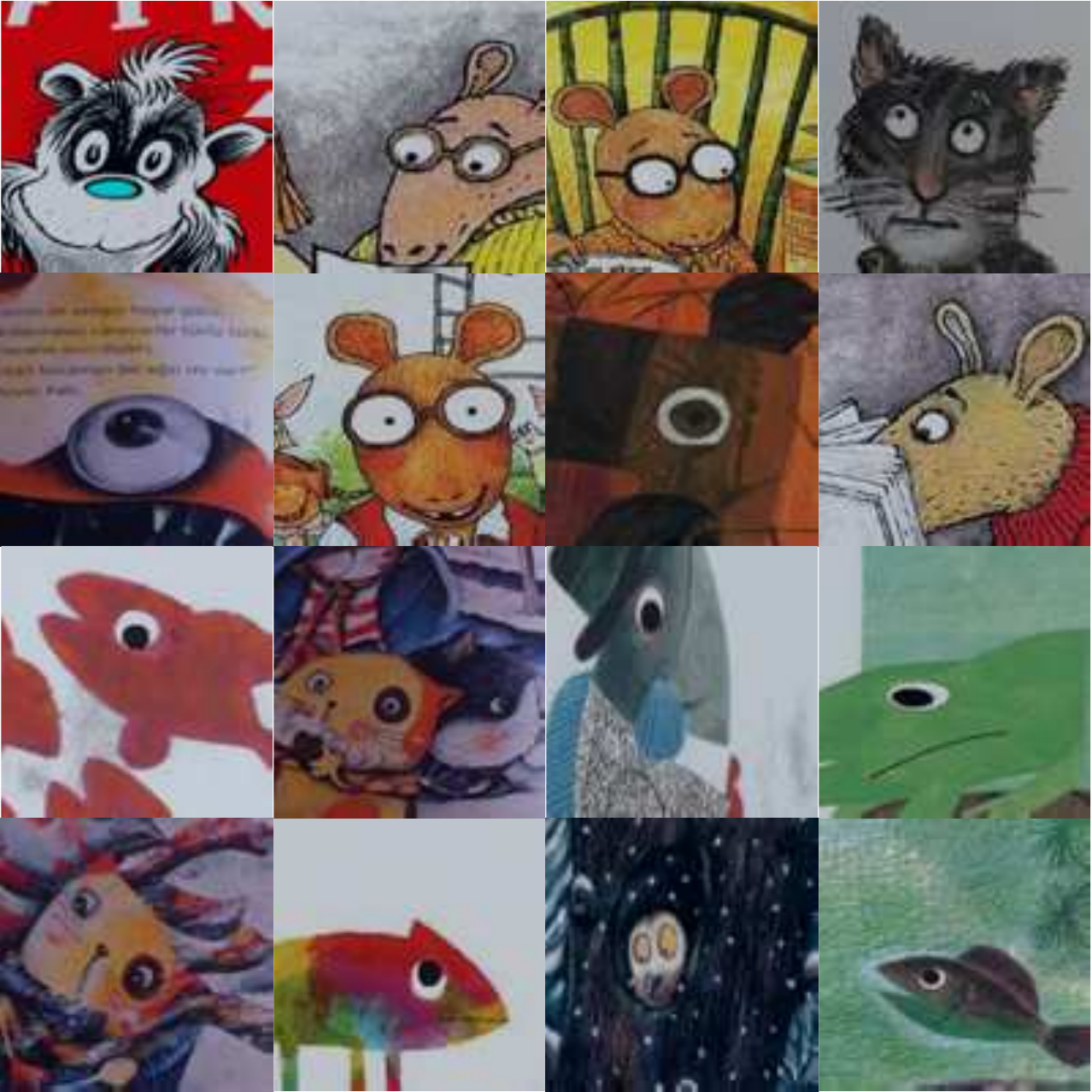} 
\hspace*{0.1cm}
\includegraphics[width=0.12\textwidth]{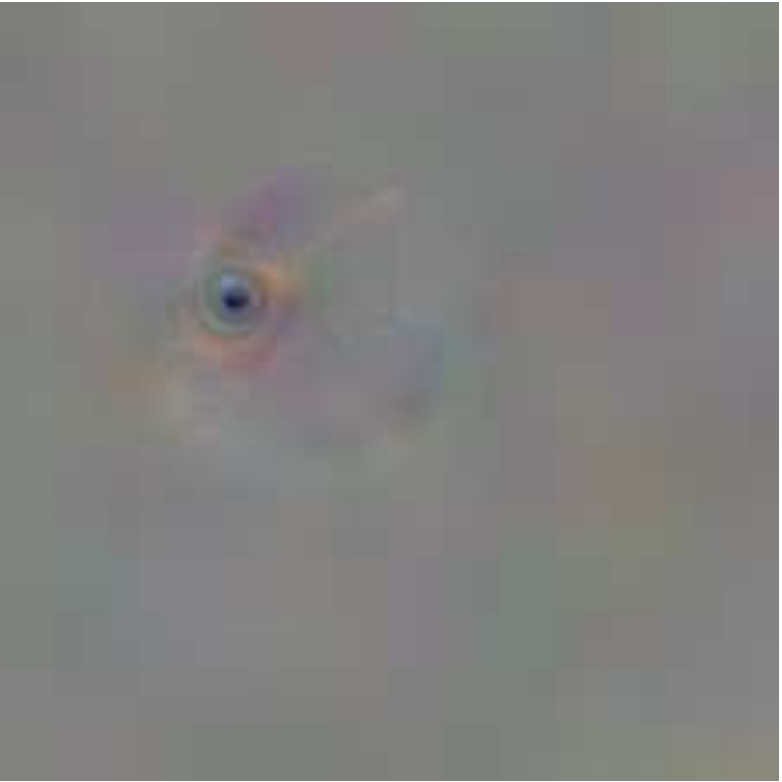} 
 \hspace*{0.001cm}
\includegraphics[width=0.12\textwidth]{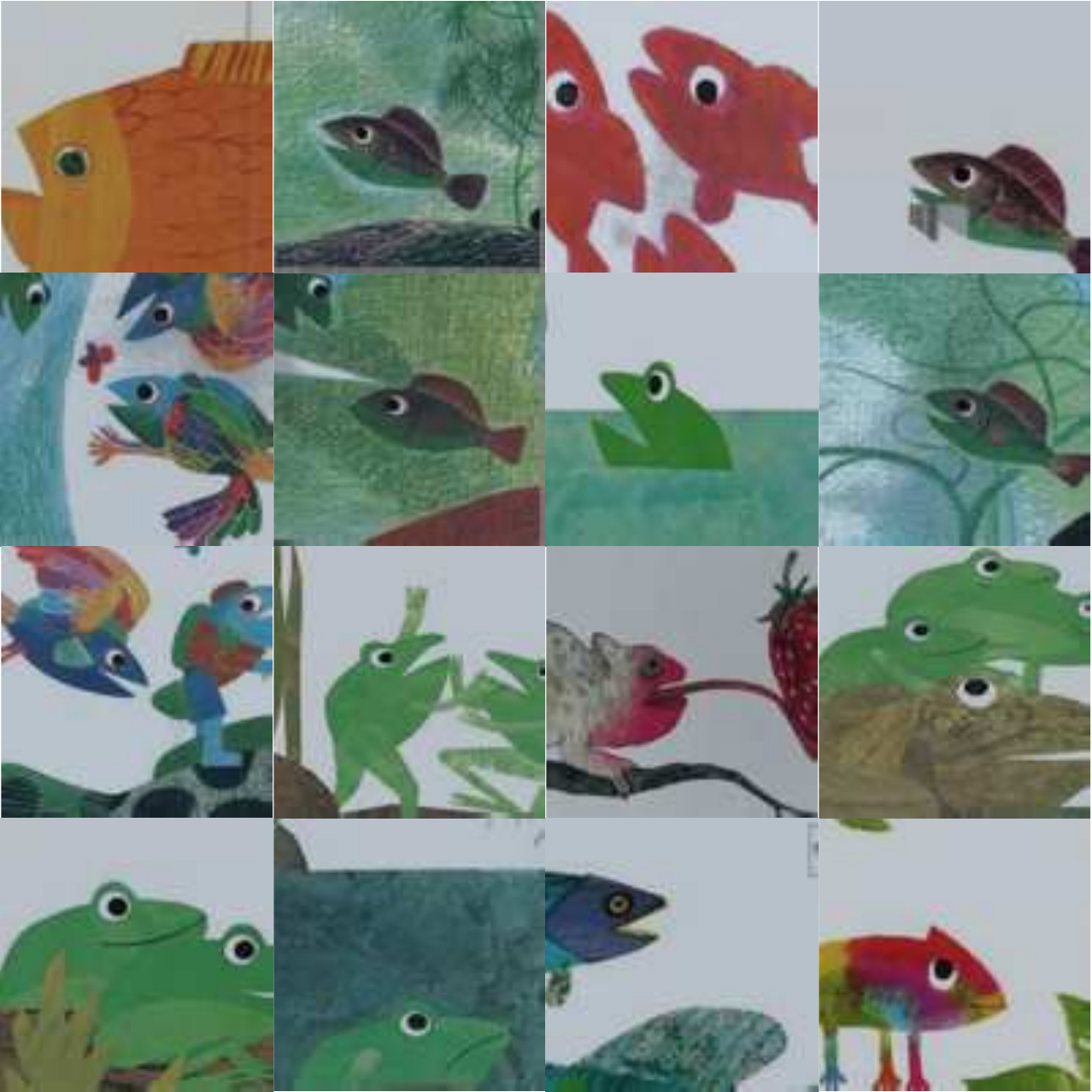} 
 \hspace*{0.1cm}
\includegraphics[width=0.12\textwidth]{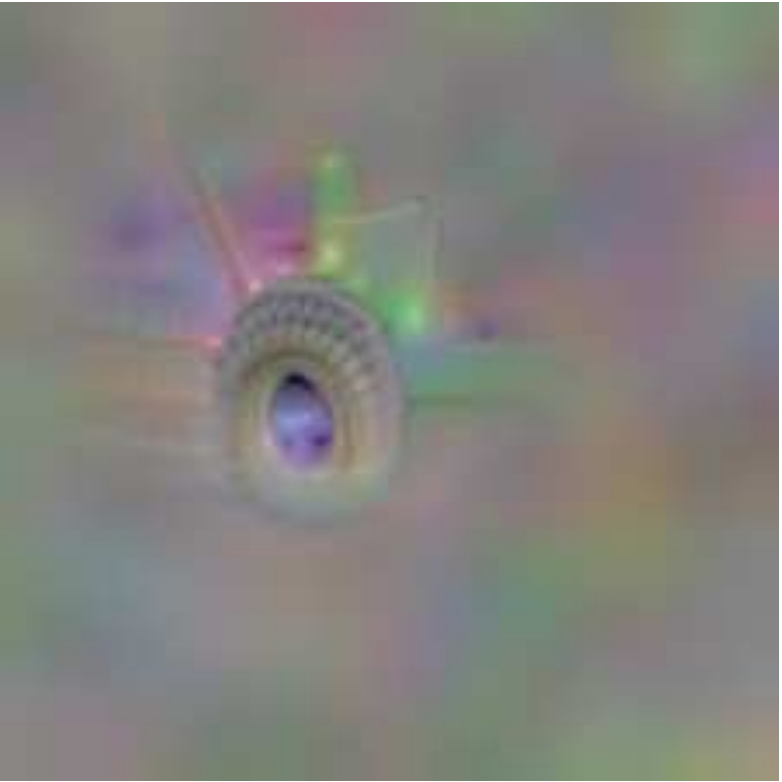} 
 \hspace*{0.001cm}
\includegraphics[width=0.12\textwidth]{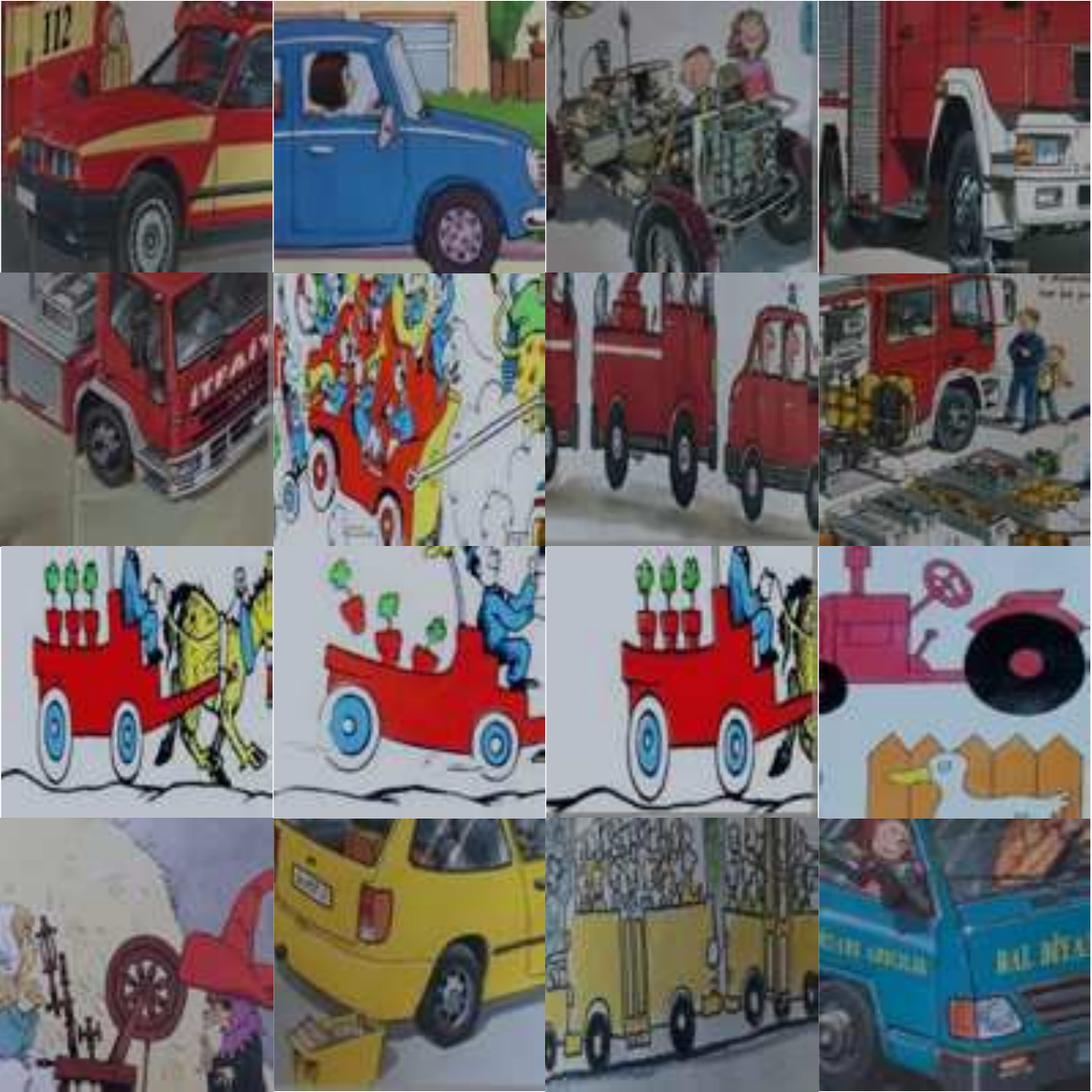}\\ 

\includegraphics[width=0.12\textwidth]{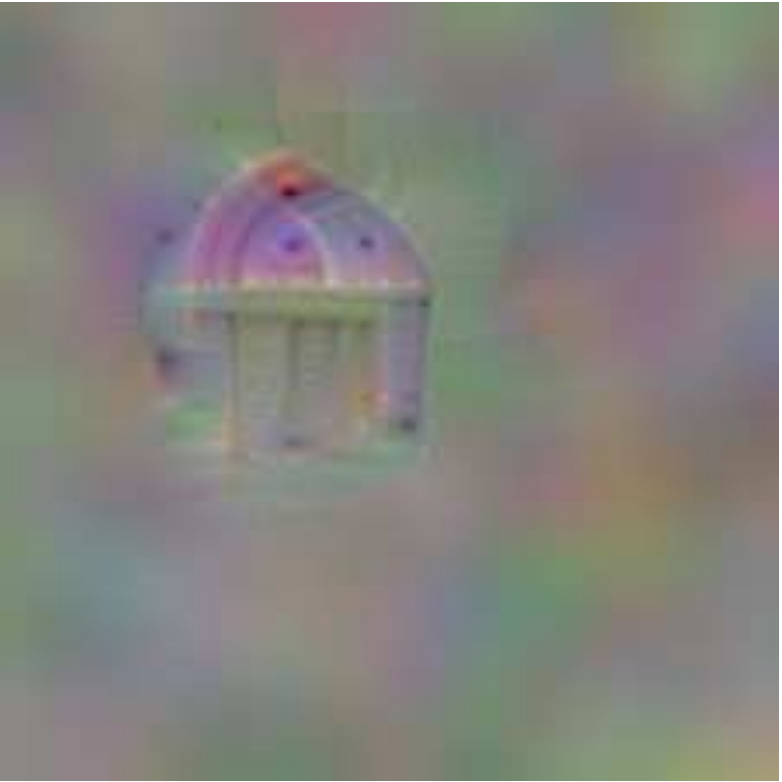} 
 \hspace*{0.001cm}
\includegraphics[width=0.12\textwidth]{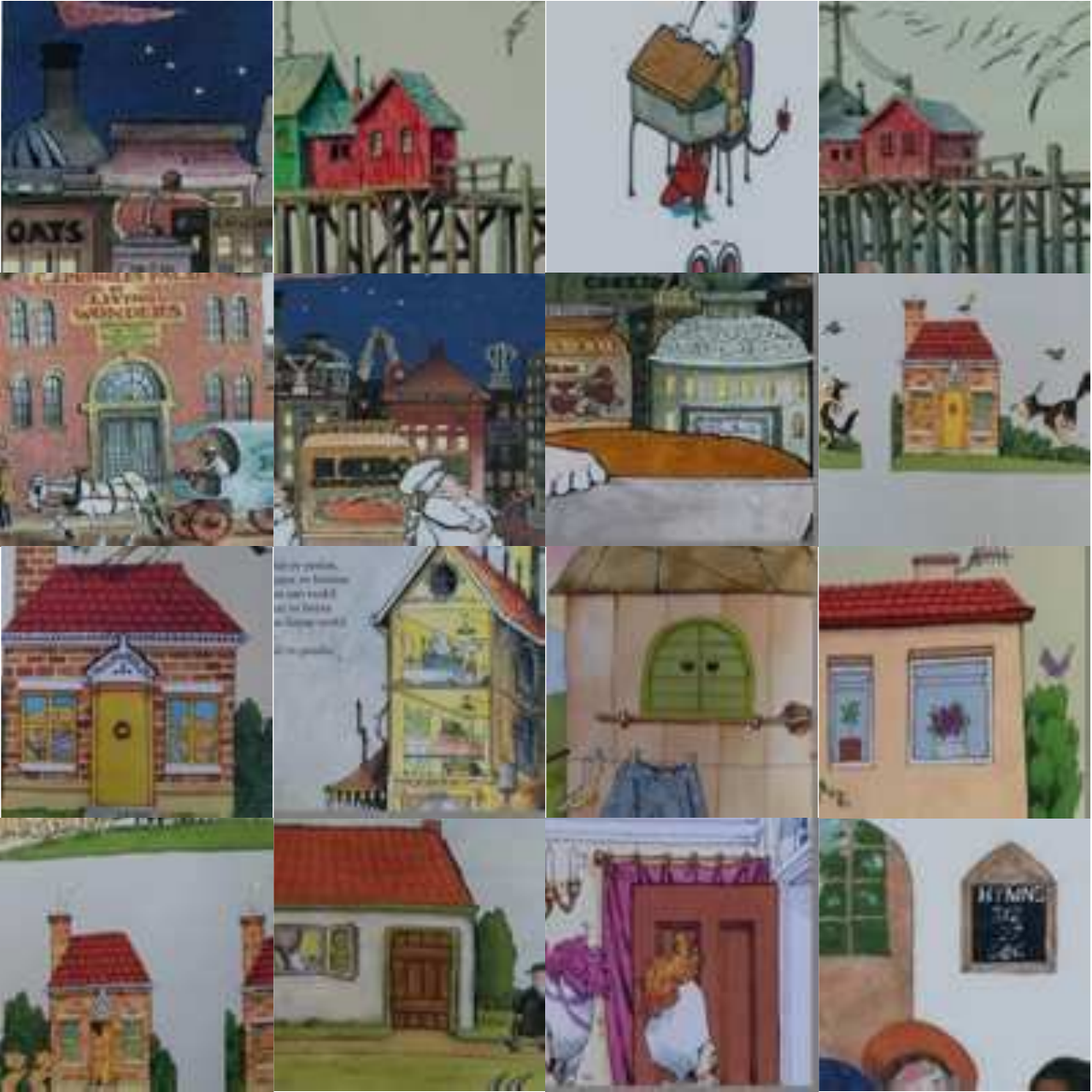} 
\hspace*{0.1cm}
\includegraphics[width=0.12\textwidth]{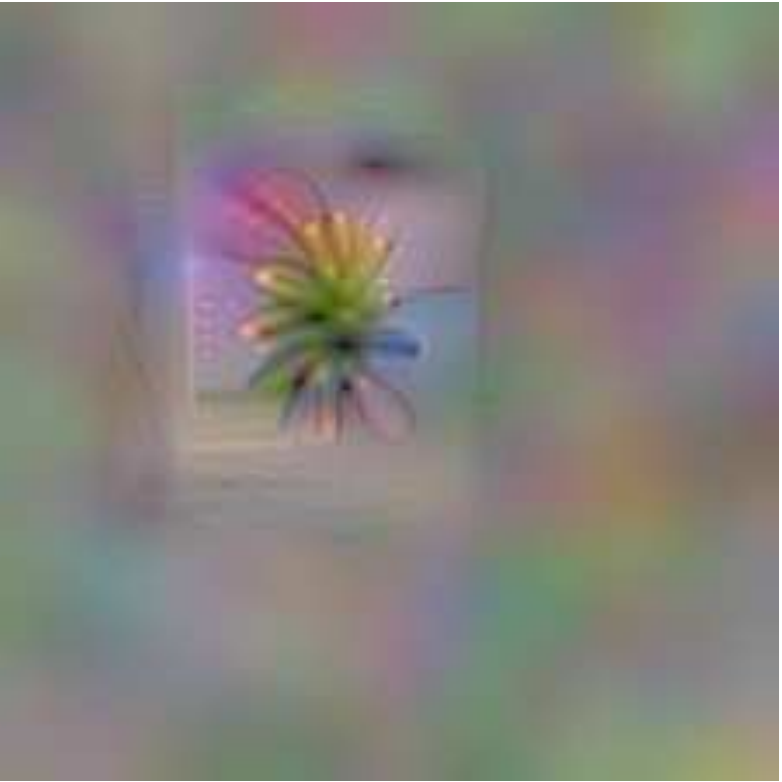} 
\hspace*{0.001cm}
\includegraphics[width=0.12\textwidth]{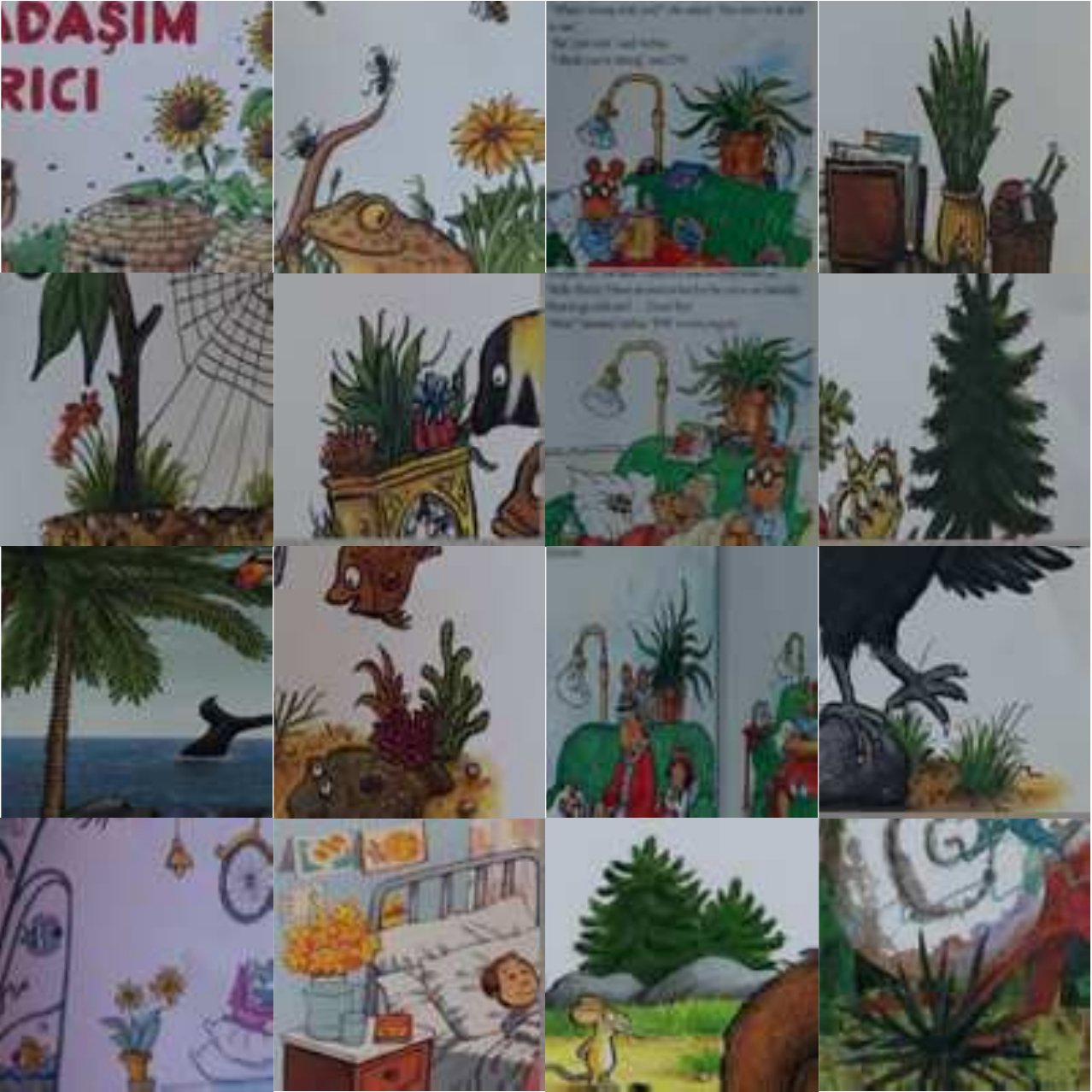} 
\hspace*{0.1cm}
\includegraphics[width=0.12\textwidth]{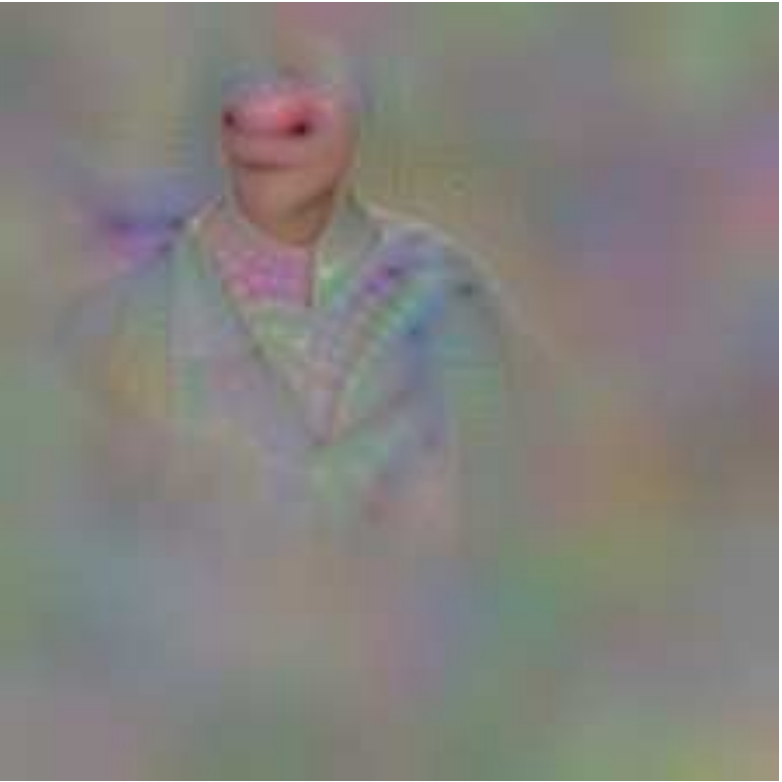} 
 \hspace*{0.001cm}
\includegraphics[width=0.12\textwidth]{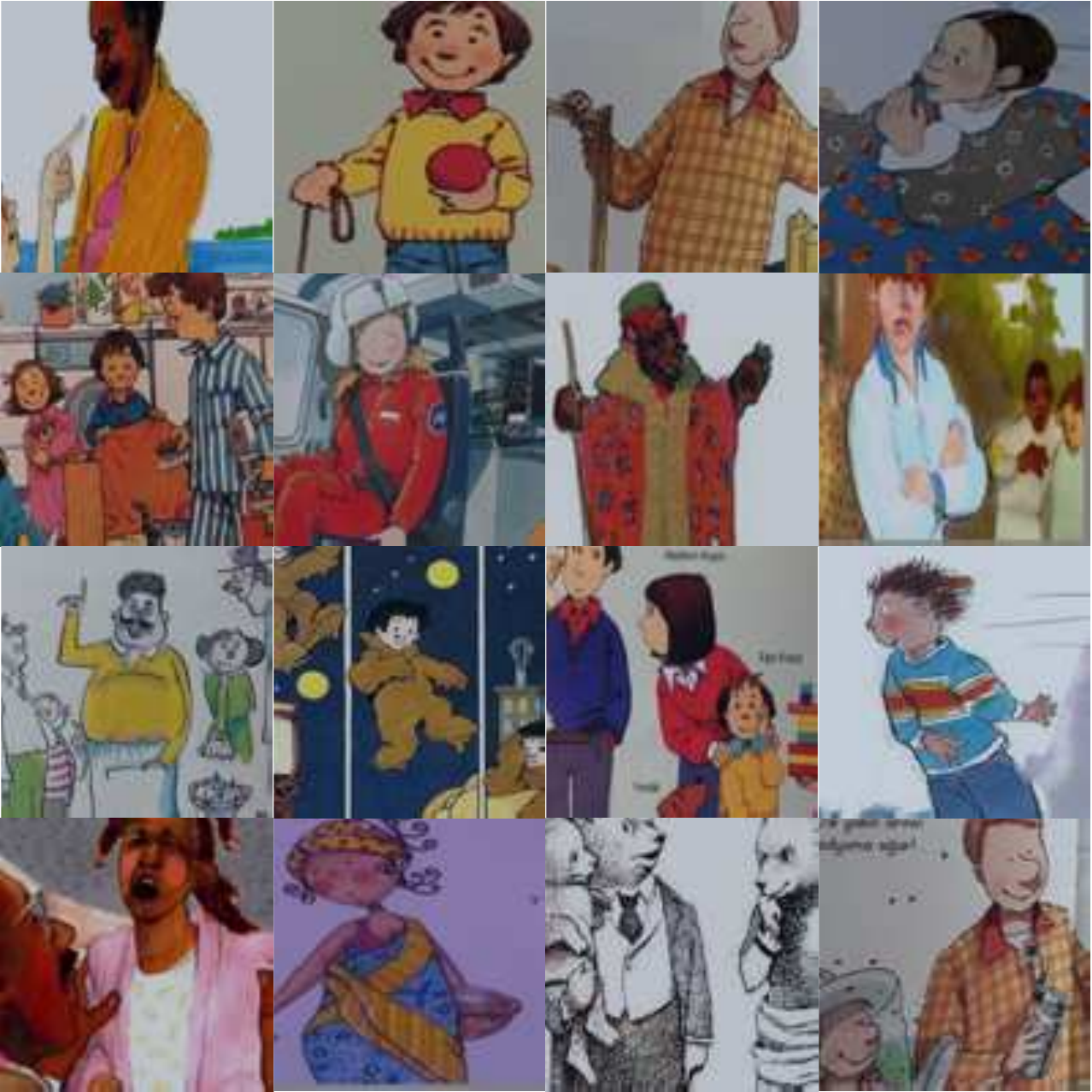}  \\

\includegraphics[width=0.12\textwidth]{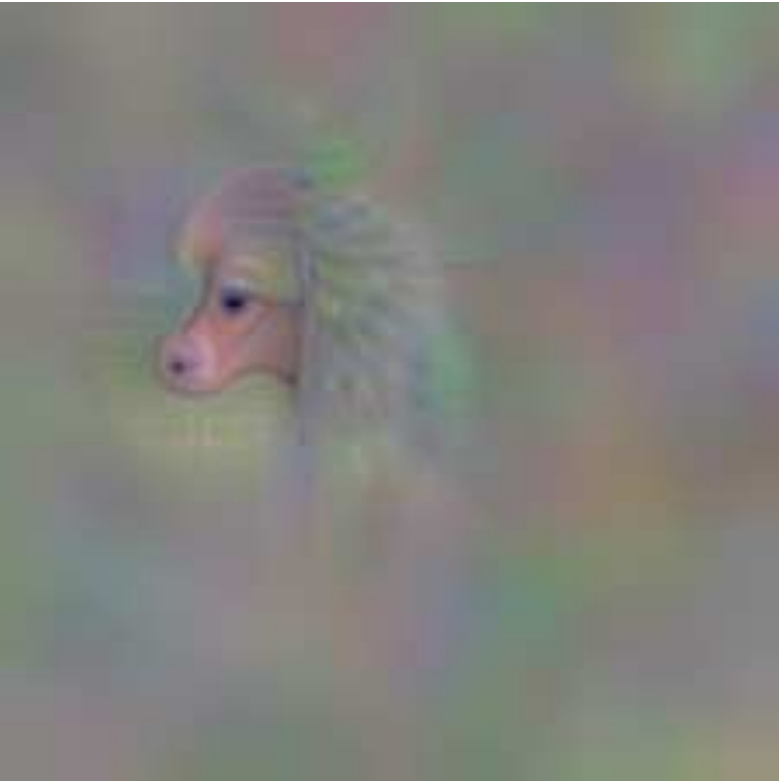} 
 \hspace*{0.001cm}
\includegraphics[width=0.12\textwidth]{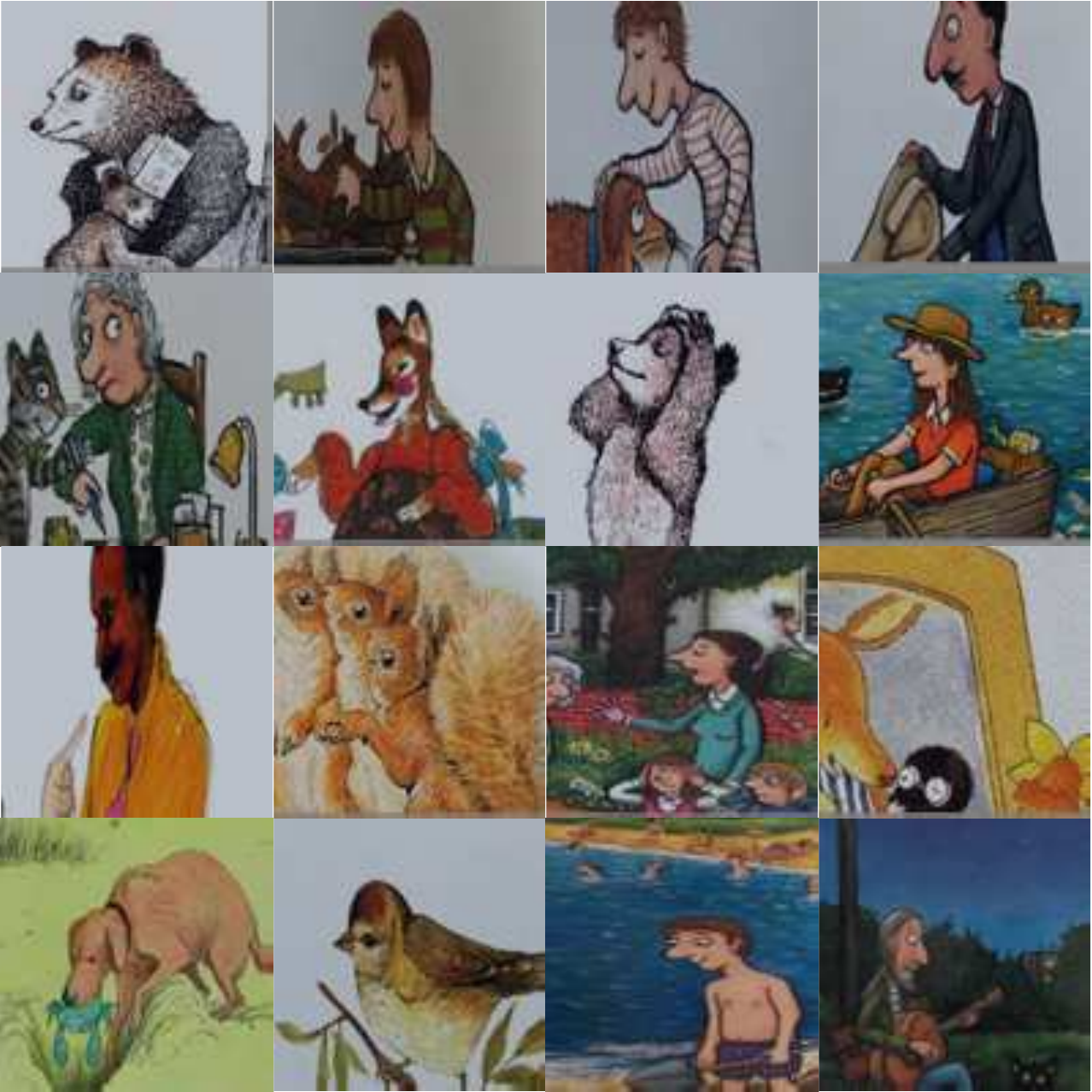} 
\hspace*{0.1cm}
\includegraphics[width=0.12\textwidth]{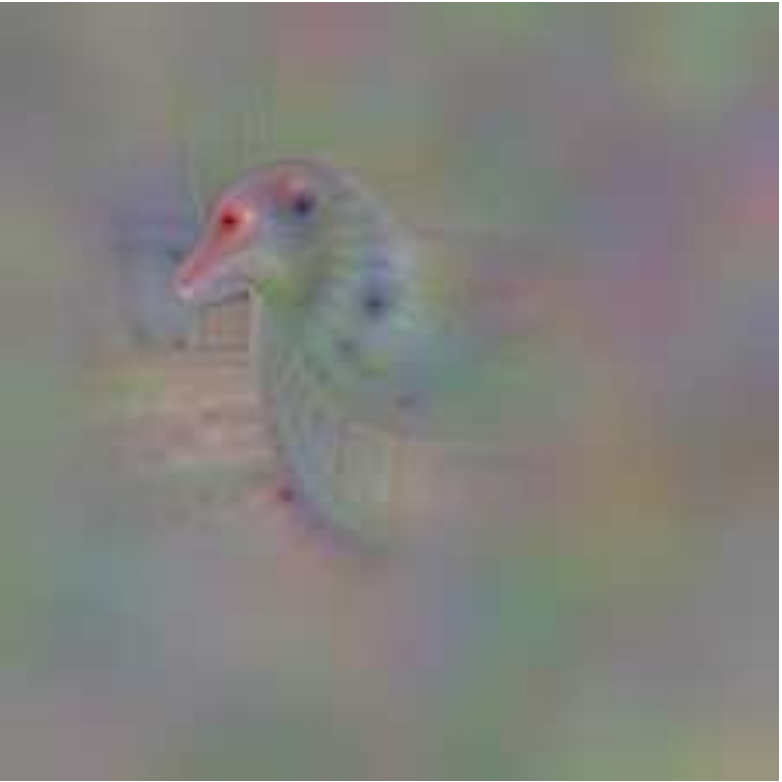} 
\hspace*{0.001cm}
\includegraphics[width=0.12\textwidth]{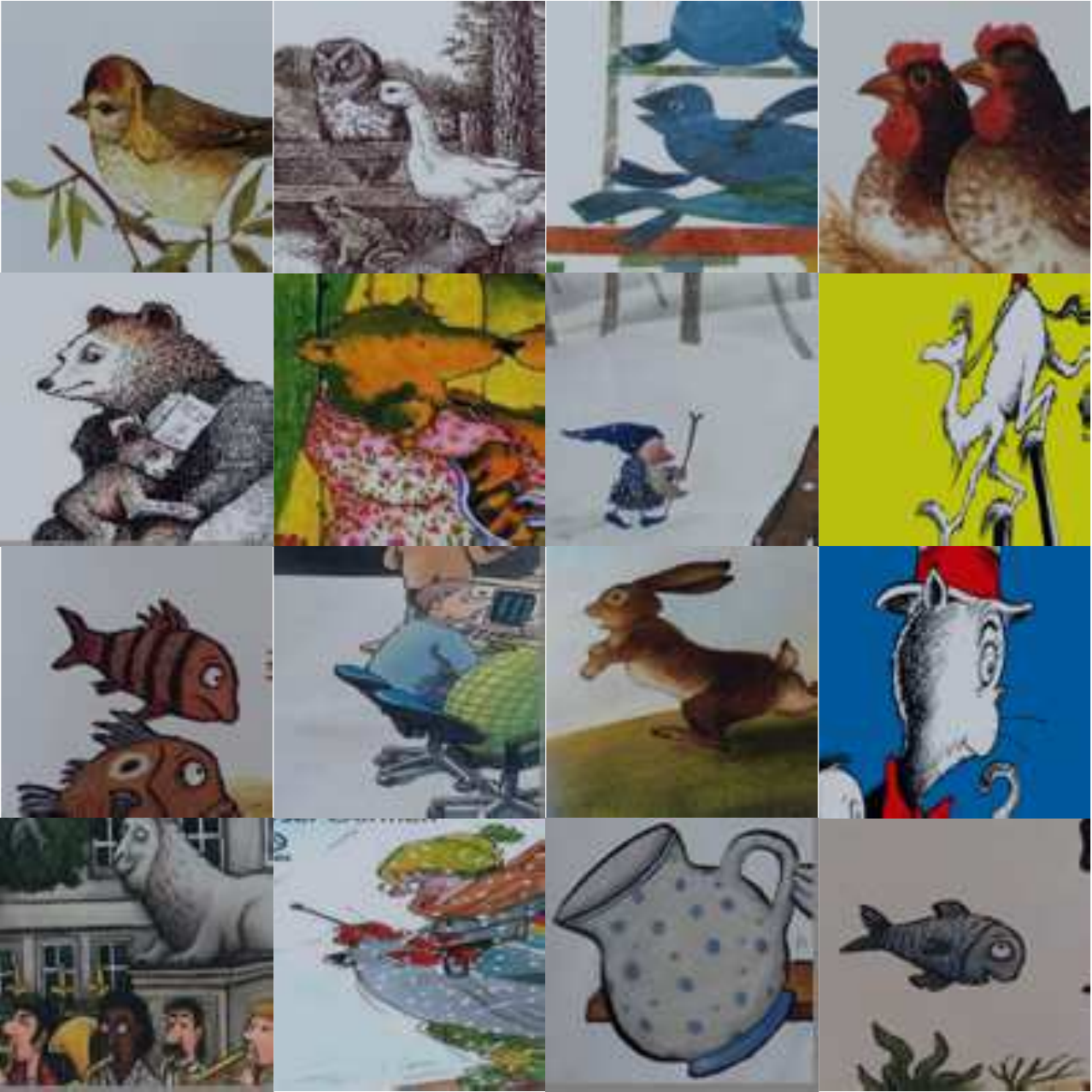}  
\hspace*{0.1cm}
\includegraphics[width=0.12\textwidth]{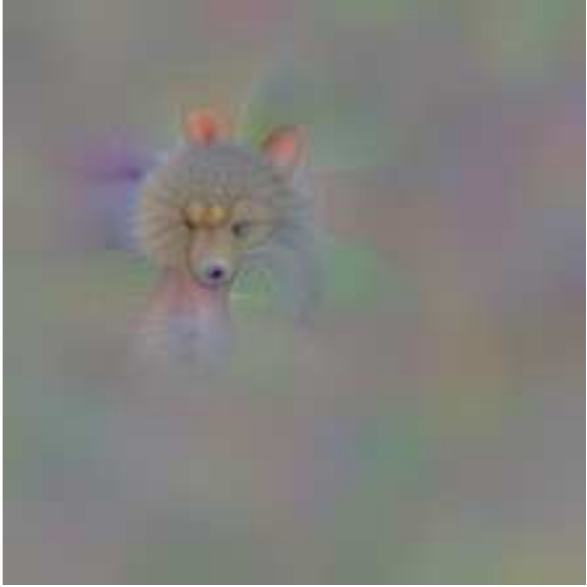} 
\hspace*{0.001cm}
\includegraphics[width=0.12\textwidth]{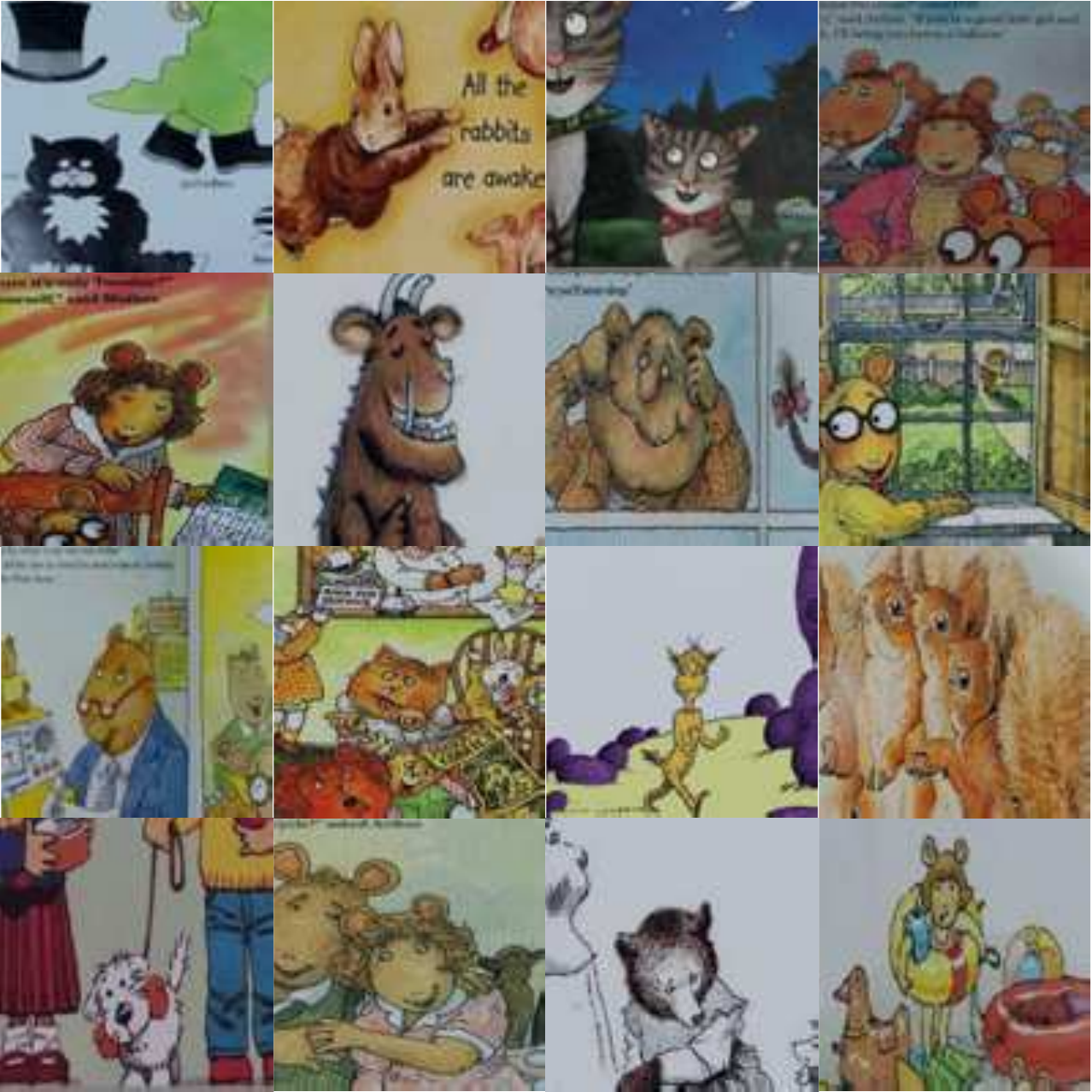}  \\
\end{tabular}
\caption{Visualizations from different layers. The network is able to capture parts and objects shared in drawings of different illustrators: eyes, wheels, buildings, pointy tree like structures, hairy or furry heads, big ears, human upper body, etc.  }
\label{fig:network_vis}
\end{center}%
\end{figure*}

To analyze and understand the results further, we exploited the method of~\cite{yosinski2015deepvis}. Figure~\ref{fig:network_vis} shows per-unit visualizations from different layers of VGG-19 network. In every image, first column corresponds to synthetic images which cause high activation using regularized optimization, and second column shows crops from our training dataset that cause highest activation for that unit. 
As it is shown, our network is able to find parts and objects such as eyes, fish, car/wheel, house, plant, people and clothes, and even discriminate poses such as side views of humans and animals, as well as hair, fur or ears.

\noindent{\bf Instance categorisation:}
In this setting, our goal is classify illustrations on a randomly carved data. Here, we don't care about the books and thus we put all the illustrations from all the books of an illustrator all together and then we construct training, validation and test sets by selecting fixed percentage of the instances randomly.

For this group of experiments we utilized several deep networks including end-to-end training of a network and fine tuning. Table~\ref{tab:exp1_results} summarizes the results in terms of the network architecture used, test type such as fully training or fine tuning the network, and whether data augmentation is used or not.
For all experiments on deep networks, we used 70\% of the data as training set and, 10\% of the data as validation set. The rest which is 20\% is used for testing.

%-----------------------Table Illustration Classification Result
\begin{table}
\centering
\caption{Illustration Classification Experiments. Fine Tuning is applied for VGG19 and GoogLeNet networks, while full training is performed for Alexnet. }
\label{tab:exp1_results}
\begin{tabular}{|c|c|c|}
\hline
Method 		& Augmentation 	& Accuracy(\%) 	\\ \hline
Alexnet 		& Yes  	& 68.75  			\\ \hline
VGG 19		& No 	& 93.47 			\\ \hline
VGG 19		 	& Yes 	& 93.24 	 		\\ \hline
GoogLeNet  	  	& No 	& 94.07 	 		\\ \hline
\hline
Dense SIFT  	  	& - 	&  82.71	 	\\ \hline
Color Dense SIFT   	& - 	& 84.35  	\\ \hline
\end{tabular}
\end{table}

\begin{figure}
\centering
%%%%% 1st row 
\includegraphics[width=0.9\linewidth]{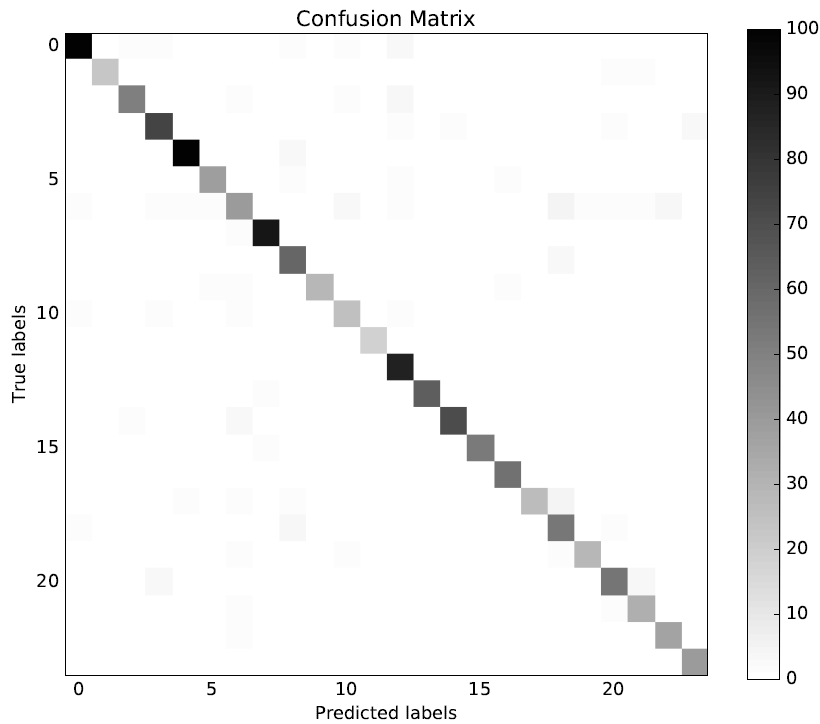} 
 \caption{Confusion Matrix for GoogleNet Finetune Test.}
\label{fig:conf_mat}
\end{figure}

As expected fully training a deep network gives less accuracy than fine tuning. Thus, in the next group of experiments we focused only on the fine tuning. Also note that, using augmented data for fine tuning a model doesn't improve the accuracy much. Thus, we preferred not to use augmented data while fine tuning a model.
GoogLeNet has much less parameters and less error rate than VGG-19 on ImageNet data set. Our results are in line with the same observation and GoogLeNet beats VGG-19 with a very small difference. Since GoogLeNet has the best performance, in the following experiments we report only the GoogLeNet results.  Figure~\ref{fig:conf_mat} and Table~\ref{tab:class_metrics} depicts confusion matrix and class-based F1 and accuracy results respectively.

\begin{table}
\centering
\caption{ Classification results using GoogLeNet finetuning.} 
\begin{tabular}{lcclcc}
\hline
& \multicolumn{2}{c}{Metrics}  &  & \multicolumn{2}{c}{Metrics}\\
\cline{2-3} \cline{5-6}
Id & F1 Score & Accuracy & Id & F1 Score & Accuracy \\
\hline
01	& 0.96 & 0.94 &  		13 & 0.95 & 1.0 \\ 
02 		& 0.96 & 0.91 & 			14 & 0.99 & 0.98 \\ 
03 	& 0.93 	& 0.91 & 		15 & 0.97 & 0.96 \\  
04 		& 0.94 & 0.93 & 			16 & 0.99 & 0.98 \\  
05 		& 0.98	& 0.98 & 		17 & 0.98 & 1.0 \\  
06 		& 0.94 & 0.92 & 			18 & 0.88 & 0.78 \\ 
07 		& 0.74 & 0.69 & 			19 & 0.87 & 0.91 \\  
08 		& 0.98 & 0.99 & 			20 & 0.93 & 0.90 \\  
09 		& 0.92 & 0.96 & 			21 & 0.92 & 0.91 \\  
10 		& 0.95 & 0.90 & 			22 & 0.90 & 0.94 \\  
11 	& 0.84 & 0.85 & 			23 & 0.95 & 0.97 \\ 
12 	& 1.0 & 1.0 & 			24 & 0.97 & 1.0 \\ 
\hline 
\end{tabular}
\label{tab:class_metrics}
\end{table}

%% Conf matrix figure

%% visualization filters

\noindent{\bf Book based instance categorisation:}
%\label{sec:book_reg}
Since illustrators are likely to have varying styles in different books, in this setup we attack a more challenging problem of recognizing the style on novel books. Instead of carving illustrations from one illustrator, we split our data in terms of books into training/validation and test sets. Thus, training and test sets do not share illustrations from the same book. Some illustrators have fewer books than others, but to measure the accuracy we make sure that every illustrator have at least one book in the test set. Note that, this setting is similar to recognizing unseen categories, and especially in the case of domain transfer problem. Leaving out some books mean having unseen characters and contents. Therefore, our recognition performance on this setting proves the capability of our method in recognizing the style but not the specific characters. Notice that the results are lower than the results of instance recognition as expected (see Table~\ref{tab:exp2_results}) . 

\noindent{\bf Book categorisation:}
We further used this network to predict the illustrator of each illustration book. Note that, in the previous settings our goal was to predict the illustrator of a single page. To predict the illustrator of a book, we used majority voting and selected the illustrator as the one having the largest number of pages assigned. 
We evaluated the performance of book categorisation with 60 different illustration books using results of VGG-19 model,  and obtained 90\% accuracy on predicting illustrator of a given book. Table~\ref{tab:exp2_results} presents the performance on book recognition.

% Experiments 2 book classification
\begin{table}
\centering
\caption{Experiments on unseen books.}
\label{tab:exp2_results}
\begin{tabular}{|c|c|c|}
\hline
Network 			 	& book based instance & book 	\\ 
                		 	& categorisation  &  categorisation	\\ \hline
 VGG 19		   	& 78.96 	 &	90.00		\\ \hline
 GoogLeNet  	 	& 79.27 	 &	88.33	\\ \hline
\hline
 Dense SIFT  		  		&  69.34  	& -  \\ \hline
 Color Dense SIFT 		&  70.00 	 & -	\\ \hline
\end{tabular}
\end{table}

\subsection{Style recognition with conventional methods}
\label{sec:clas_met}

As a baseline method, we utilized conventional feature extraction methods that are shown to have the highest accuracies in \cite{sener2012identification}.
We extracted Dense SIFT~\cite{sift} and Color Dense SIFT~\cite{sift} features from every illustration 
and then generated a code book for Bag-of-words~\cite{bow} representation using k-means clustering.
We use Support Vector Machines to train our model. In particular LIBSVM library~\cite{libsvm} is used for SVM classification. 
We use one versus all approach for training where to prepare the training set for a class, we provide the negative samples from all other classes. 
A test example is fed into multiple classifiers and it is assigned to the class with
the highest confidence value. 
%Several different kernels were used for each set of features, including chi-square kernel, linear kernel, histogram intersection kernel, Radial Basis Function kernel and Hellinger’s kernel.
Half of the data set is used to train SVMs, and the rest is used for testing the models. We observe Hellinger's kernel boosts the performance by almost 20\% over other kernels. 
As seen in Table~\ref{tab:exp1_results} and Table~\ref{tab:exp2_results} the results are much lower compared to the results of deep network architectures.

\subsection{Style Transfer on Illustration Dataset}
\label{sec:styletrans}

In style transfer experiments, we first selected a simple content image (cartoon image or a natural photograph) gathered from web and has no relation with our data set. Then, we randomly chose a group of illustrations from different illustrators as style images.  In our second experiment, we challenged the problem and selected an illustration from our data set as the content image. In this setting, style image is an illustration from our data set, and content image is again an illustration but belongs to a different illustrator. We performed style transfer using each style and content image, and looked for the recognizing performance of our deep model on the resulting images. We use fine tuned GoogLeNet in all style transfer experiments.

\begin{figure*}[t]
\begin{center}
\centering
%\begin{tabular}{ccccc}
%%%%% 1st row 

\includegraphics[width=0.11\textwidth]{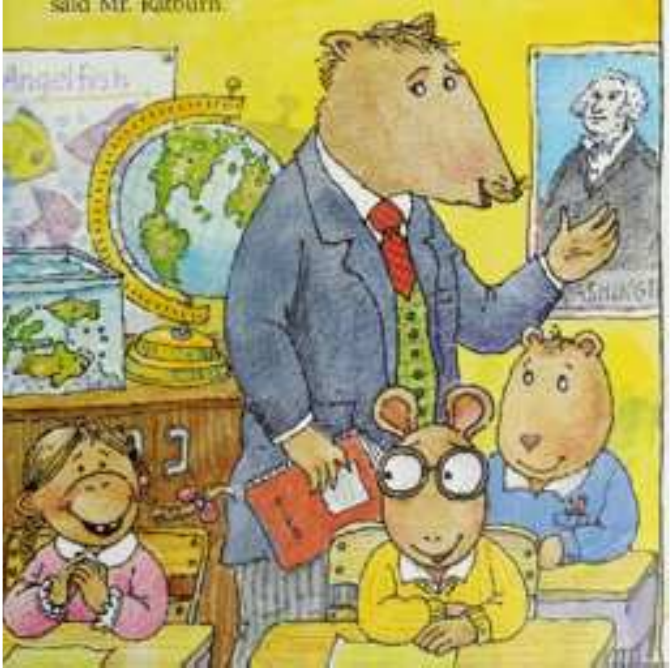} 
 \hspace*{0.1cm}
\includegraphics[width=0.11\textwidth]{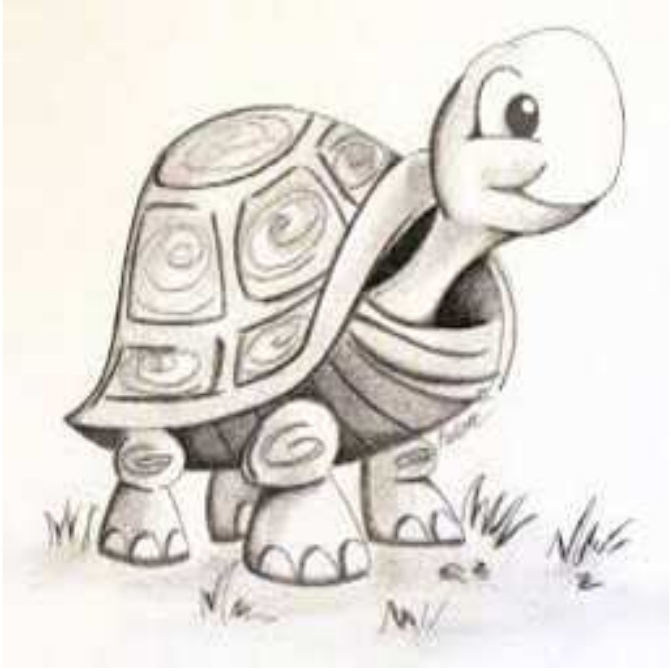} 
\hspace*{0.001cm}
\includegraphics[width=0.11\textwidth]{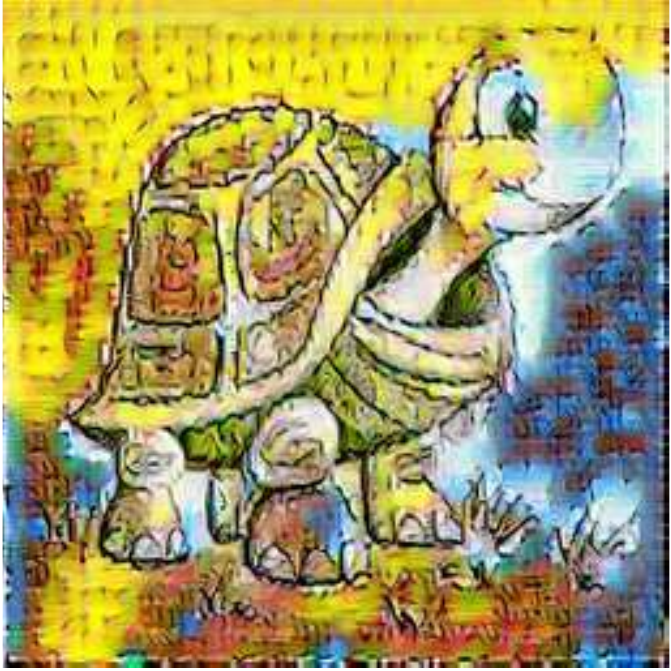} 
 \hspace*{0.1cm}
\includegraphics[width=0.11\textwidth]{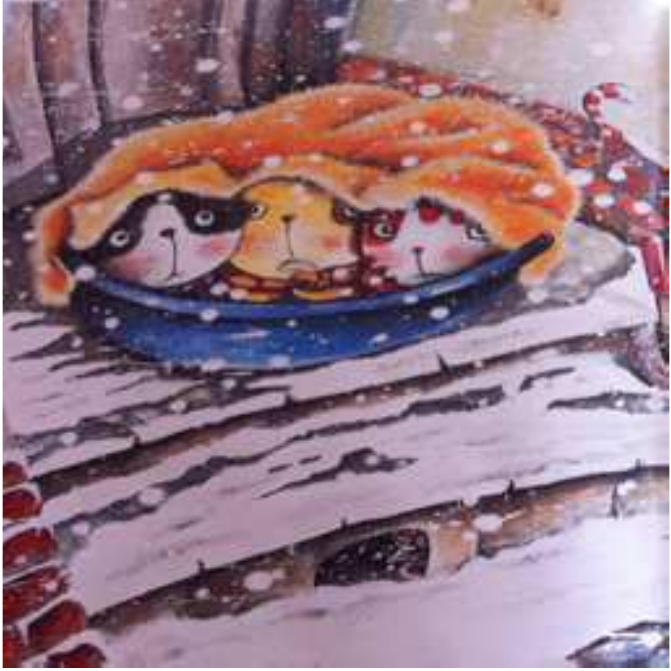} 
 \hspace*{0.001cm}
\includegraphics[width=0.11\textwidth]{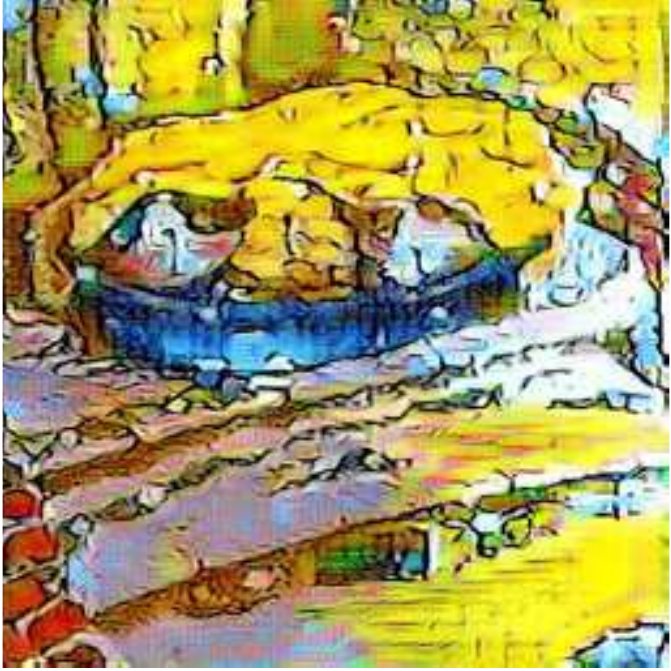} 
 \hspace*{0.1cm}
\includegraphics[width=0.11\textwidth]{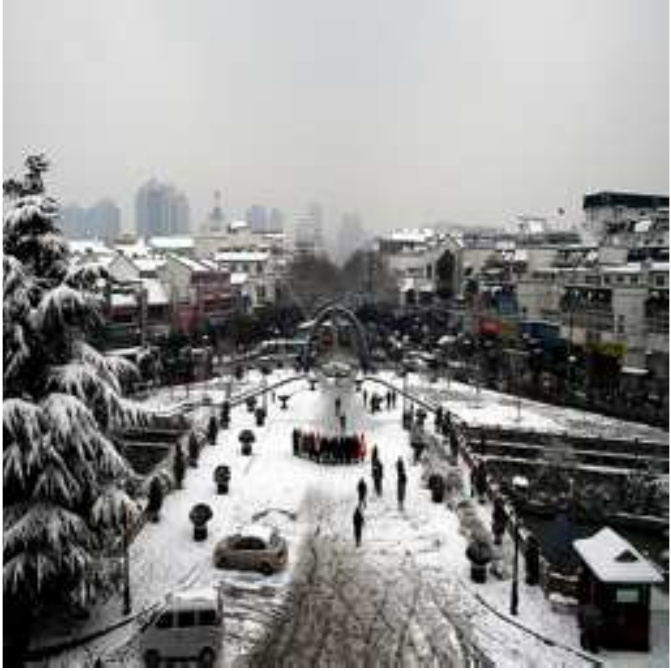} 
 \hspace*{0.001cm}
\includegraphics[width=0.11\textwidth]{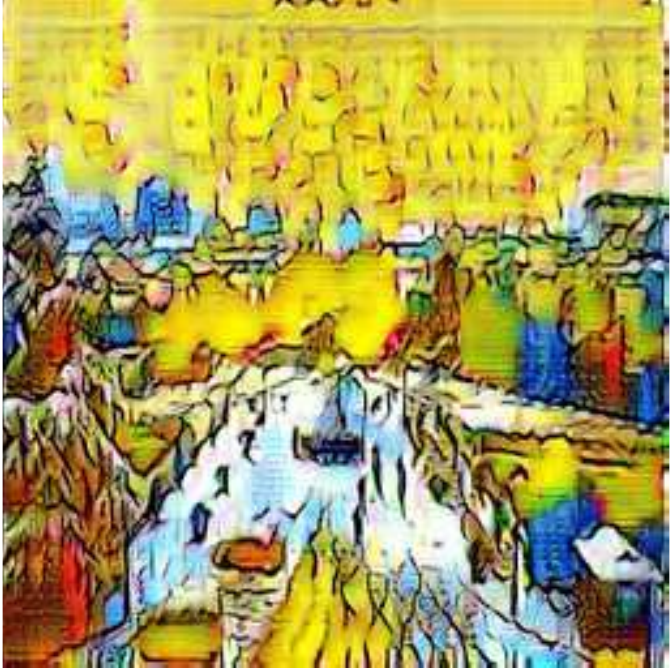} \\

\includegraphics[width=0.11\textwidth]{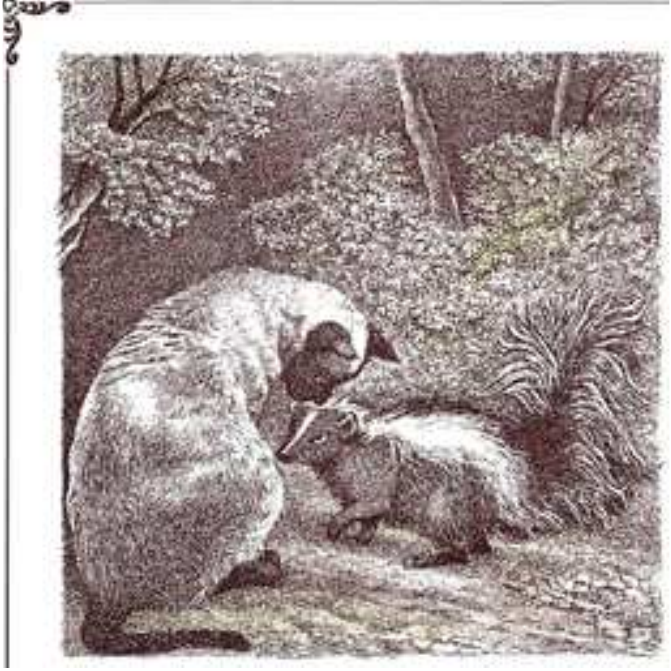} 
 \hspace*{0.1cm}
\includegraphics[width=0.11\textwidth]{./figures/style_transfer/content_mutual-eps-converted-to.pdf} 
\hspace*{0.001cm}
\includegraphics[width=0.11\textwidth]{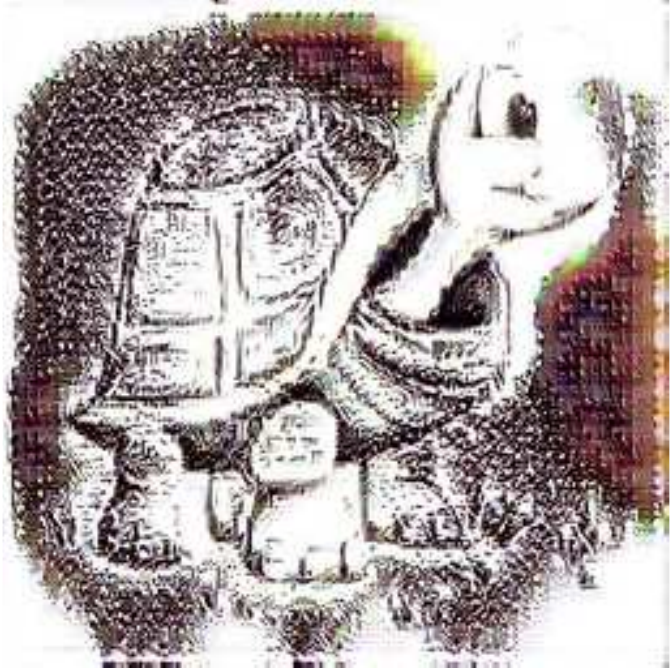} 
 \hspace*{0.1cm}
\includegraphics[width=0.11\textwidth]{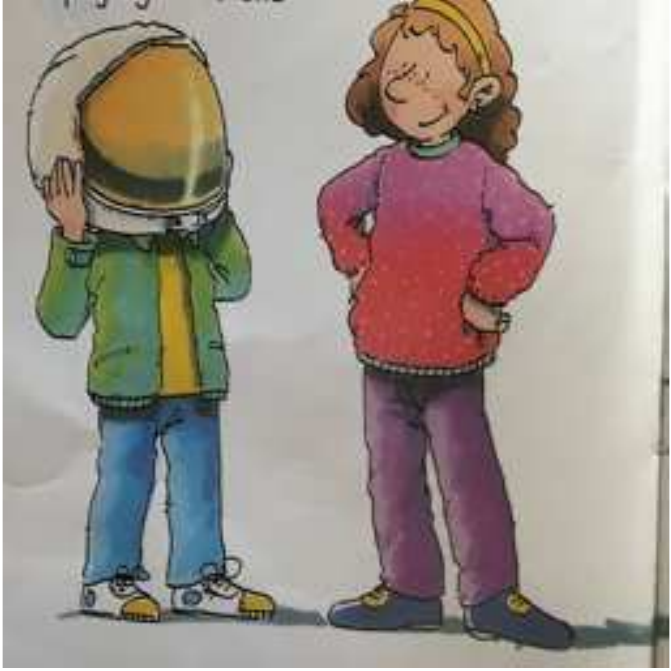} 
 \hspace*{0.001cm}
\includegraphics[width=0.11\textwidth]{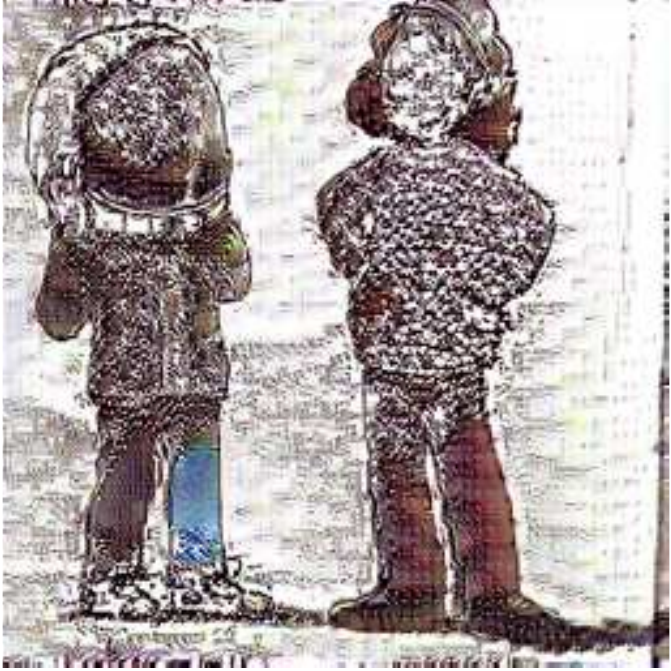} 
 \hspace*{0.1cm}
\includegraphics[width=0.11\textwidth]{./figures/style_transfer/content_nanjing-eps-converted-to.pdf} 
 \hspace*{0.001cm}
\includegraphics[width=0.11\textwidth]{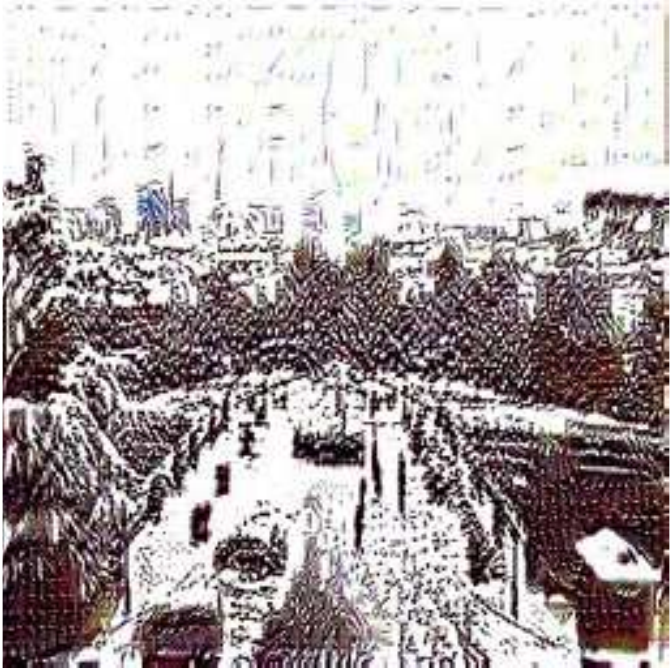} \\

\includegraphics[width=0.11\textwidth]{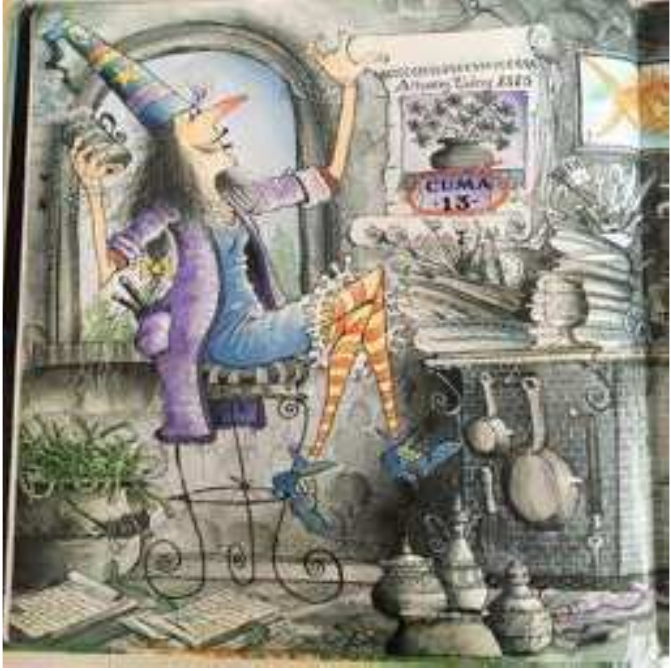} 
 \hspace*{0.1cm}
\includegraphics[width=0.11\textwidth]{./figures/style_transfer/content_mutual-eps-converted-to.pdf} 
\hspace*{0.001cm}
\includegraphics[width=0.11\textwidth]{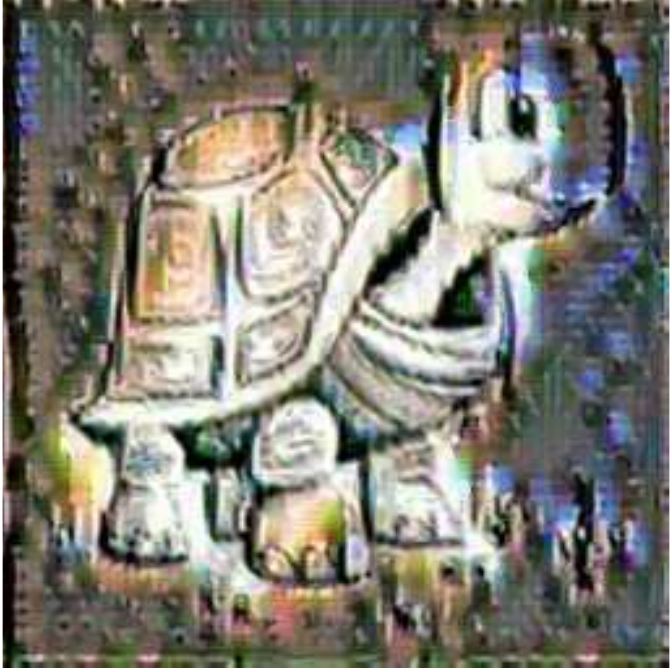} 
 \hspace*{0.1cm}
\includegraphics[width=0.11\textwidth]{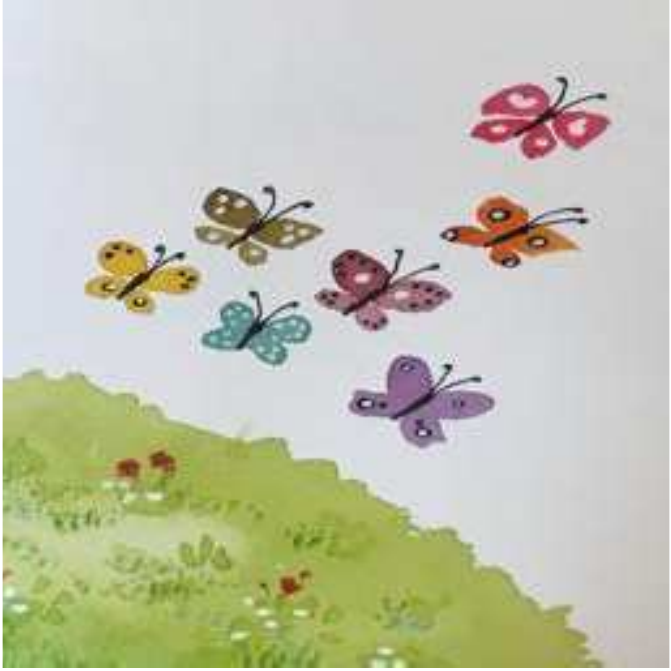} 
 \hspace*{0.001cm}
\includegraphics[width=0.11\textwidth]{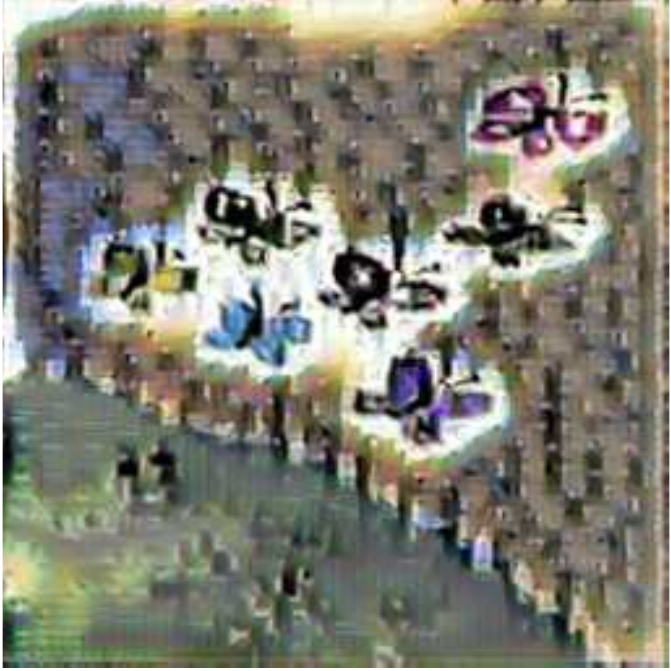} 
 \hspace*{0.1cm}
\includegraphics[width=0.11\textwidth]{./figures/style_transfer/content_nanjing-eps-converted-to.pdf} 
 \hspace*{0.001cm}
\includegraphics[width=0.11\textwidth]{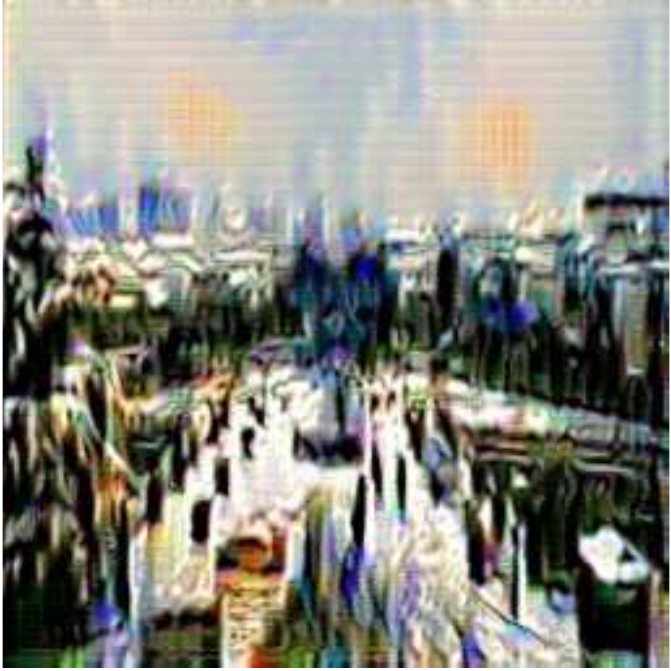} \\

\includegraphics[width=0.11\textwidth]{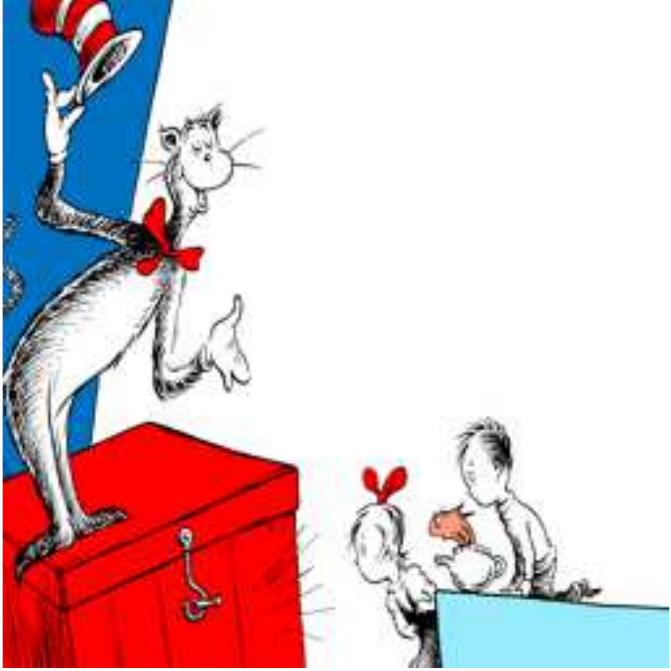} 
 \hspace*{0.1cm}
\includegraphics[width=0.11\textwidth]{./figures/style_transfer/content_mutual-eps-converted-to.pdf} 
\hspace*{0.001cm}
\includegraphics[width=0.11\textwidth]{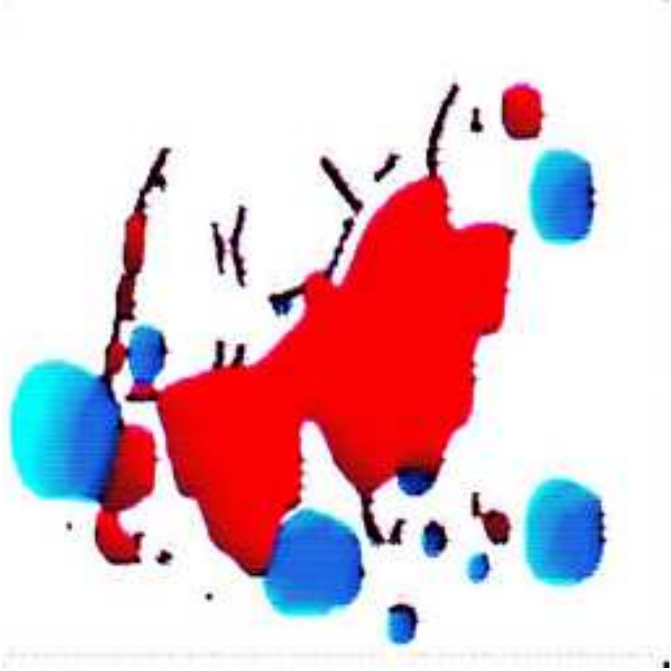}
 \hspace*{0.1cm}
\includegraphics[width=0.11\textwidth]{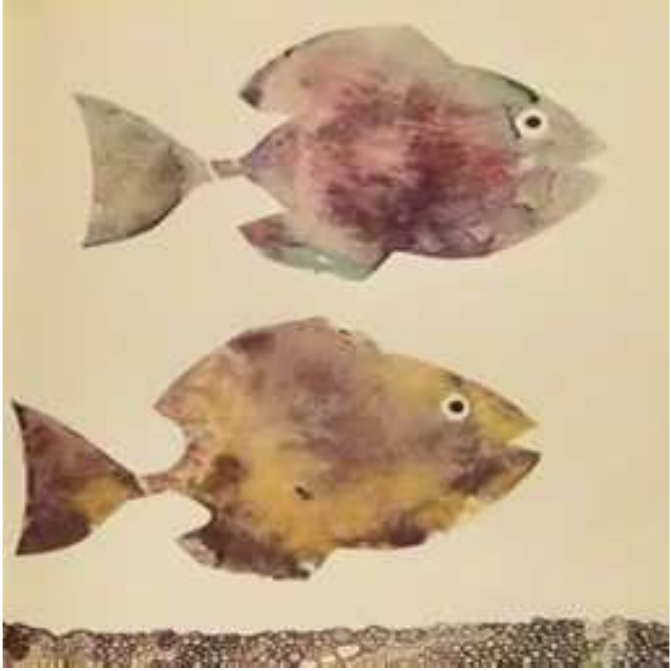} 
 \hspace*{0.001cm}
\includegraphics[width=0.11\textwidth]{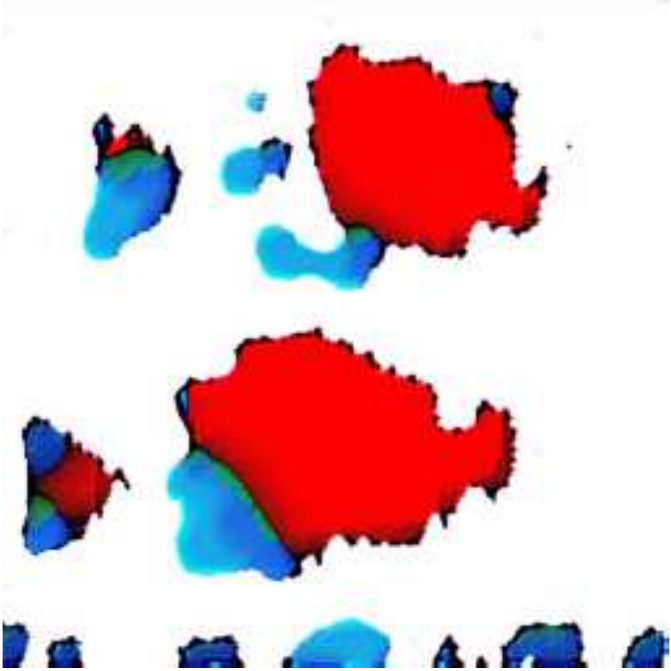} 
 \hspace*{0.1cm}
\includegraphics[width=0.11\textwidth]{./figures/style_transfer/content_nanjing-eps-converted-to.pdf} 
 \hspace*{0.001cm}
\includegraphics[width=0.11\textwidth]{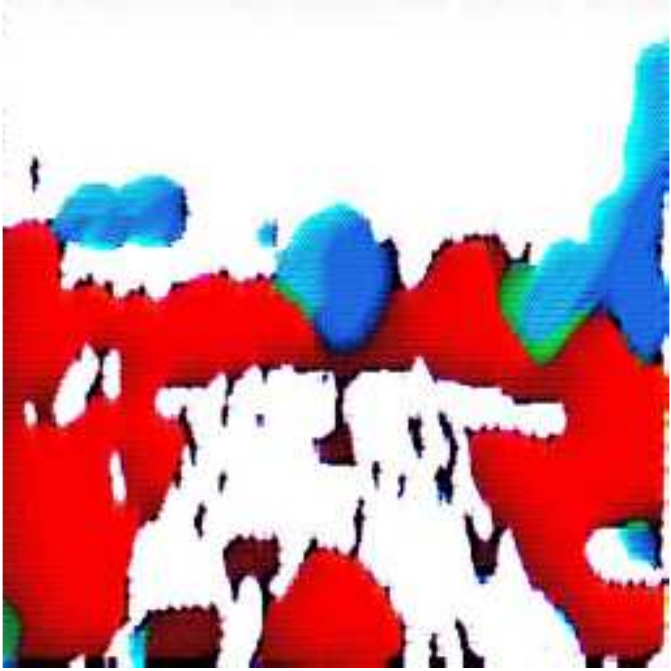} \\ 

%\includegraphics[width=0.12\textwidth]{./figures/style_transfer/style_beatrix-eps-converted-to.pdf} 
% \hspace*{0.1cm}
%\includegraphics[width=0.12\textwidth]{./figures/style_transfer/content_mutual-eps-converted-to.pdf} 
%\hspace*{0.001cm}
%\includegraphics[width=0.12\textwidth]{./figures/style_transfer/output_mutual_beatrix-eps-converted-to.pdf} 
% \hspace*{0.1cm}
%\includegraphics[width=0.12\textwidth]{./figures/style_transfer/content_feridun-eps-converted-to.pdf} 
% \hspace*{0.001cm}
%\fcolorbox{red}{red}{\includegraphics[width=0.12\textwidth]{./figures/style_transfer/output_diff_beatrix-eps-converted-to.pdf}} 
% \hspace*{0.1cm}
%\includegraphics[width=0.12\textwidth]{./figures/style_transfer/content_nanjing-eps-converted-to.pdf} 
% \hspace*{0.001cm}
%\fcolorbox{red}{red}{\includegraphics[width=0.12\textwidth]{./figures/style_transfer/output_nanjing-beatrix-eps-converted-to.pdf}}  \\

\includegraphics[width=0.10\textwidth]{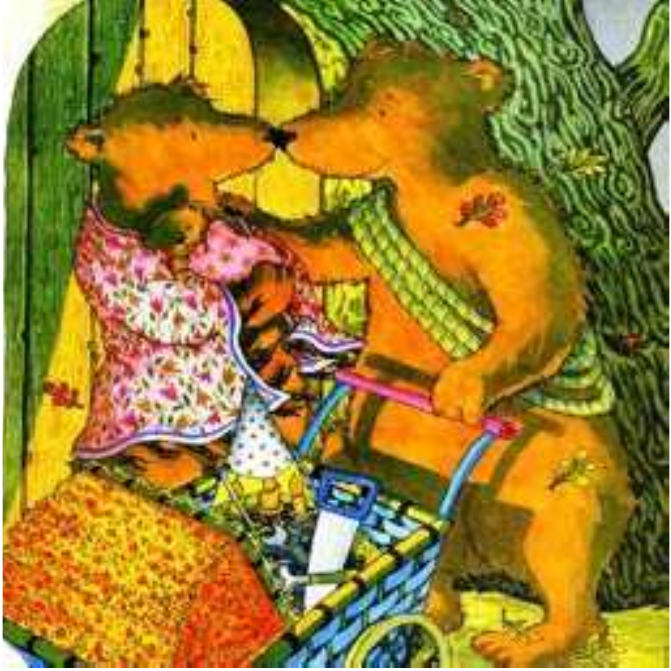} 
 \hspace*{0.1cm}
\includegraphics[width=0.10\textwidth]{./figures/style_transfer/content_mutual-eps-converted-to.pdf} 
\hspace*{0.001cm}
\fcolorbox{red}{red}{\includegraphics[width=0.10\textwidth]{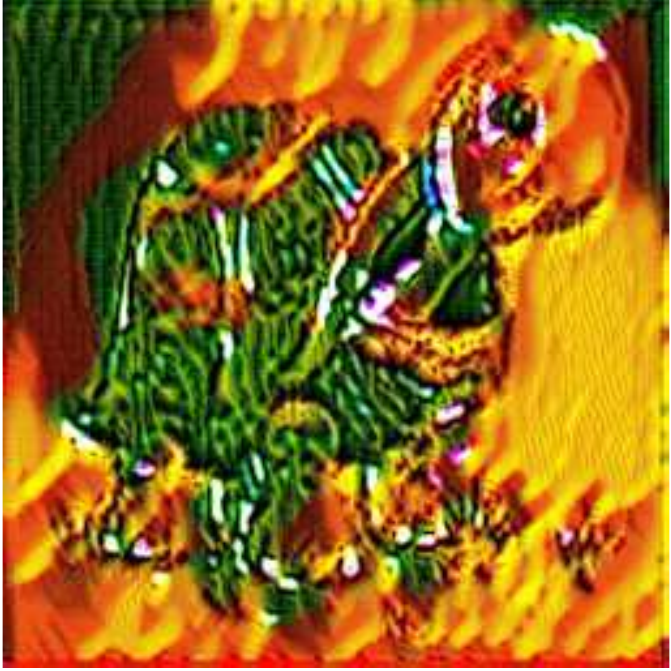}} 
 \hspace*{0.1cm}
\includegraphics[width=0.10\textwidth]{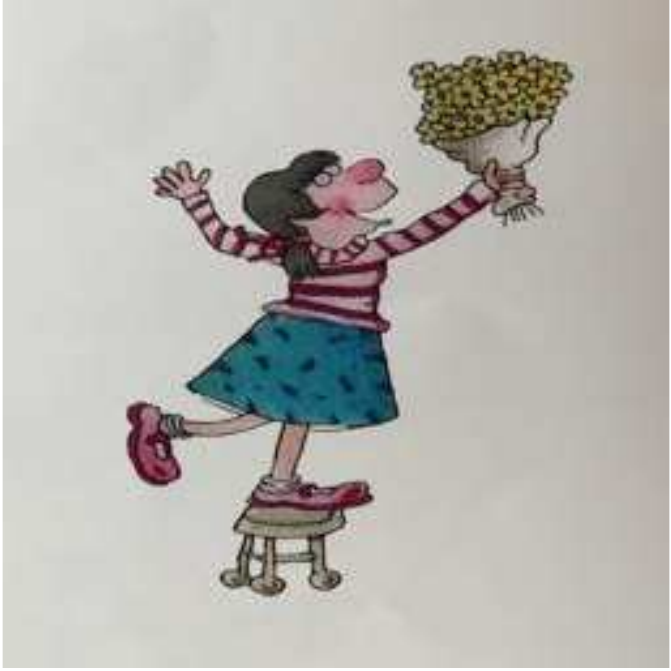} 
 \hspace*{0.001cm}
\fcolorbox{red}{red}{\includegraphics[width=0.10\textwidth]{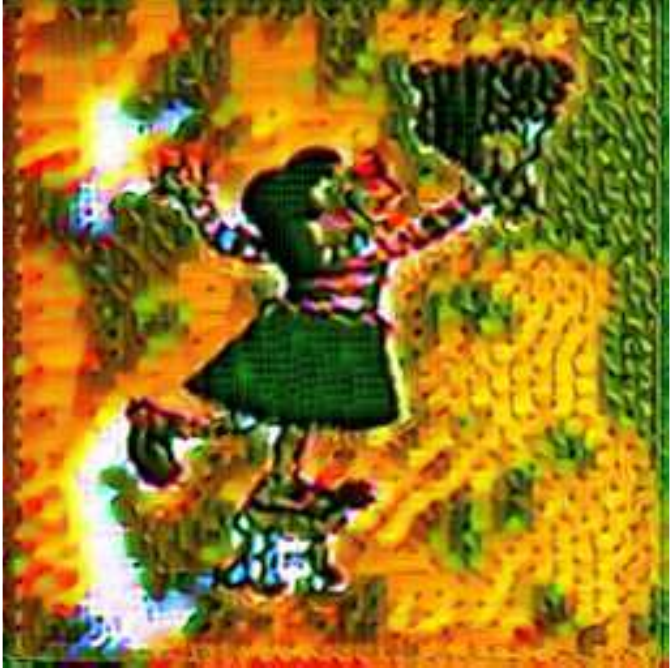}} 
 \hspace*{0.1cm}
\includegraphics[width=0.10\textwidth]{./figures/style_transfer/content_nanjing-eps-converted-to.pdf} 
 \hspace*{0.001cm}
\fcolorbox{red}{red}{\includegraphics[width=0.10\textwidth]{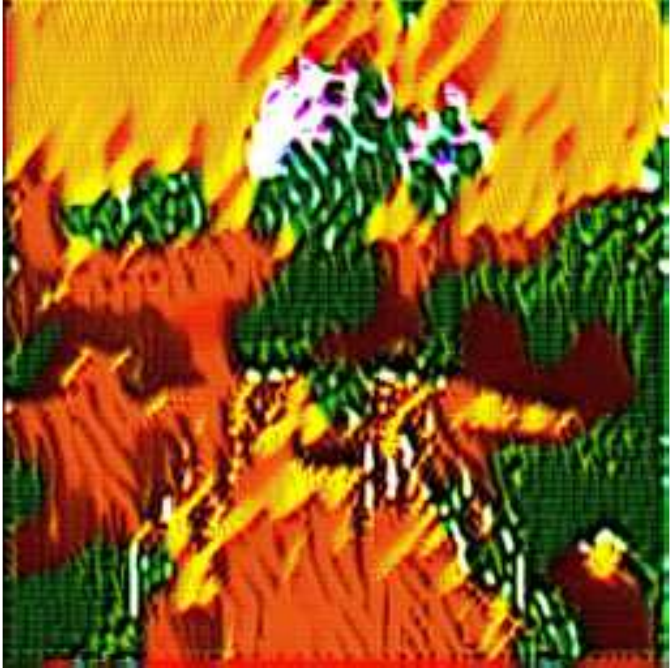}} 

%\includegraphics[width=0.13\textwidth]{./figures/style_transfer/style_dr_seuss-eps-converted-to.pdf} 
% \hspace*{0.1cm}
%\includegraphics[width=0.13\textwidth]{./figures/style_transfer/content_mutual-eps-converted-to.pdf} 
%\hspace*{0.001cm}
%\fcolorbox{red}{red}{\includegraphics[width=0.13\textwidth]{./figures/style_transfer/output_mutual_dr_seuss-eps-converted-to.pdf}} 
% \hspace*{0.1cm}
%\includegraphics[width=0.13\textwidth]{./figures/style_transfer/content_leo-eps-converted-to.pdf} 
% \hspace*{0.001cm}
%\fcolorbox{red}{red}{\includegraphics[width=0.13\textwidth]{./figures/style_transfer/output_diff_dr_seuss-eps-converted-to.pdf}} \\ 
%\end{tabular}
\caption{Example results for style transfer. 
First column shows selected style images. 
2nd, 4th and 6th columns present content images: a simple cartoon, an illustration from a different illustrator and a natural image respectively. 3rd, 5th and 7th columns belong to resulting images. The style images are from Marc Brown, Maurice Sendak, Korky Paul, Dr. Seuss, Debi Gliori and content images are from Ayse Inal, Ralf Butschkow, Rosa Curto, Leo Lionni, Behic Ak in the given order. Red boxes show the failure cases. }
\label{fig:style_trans}
\end{center}%
\end{figure*}

Figure~\ref{fig:style_trans} illustrates the style transfer results for the given style and content images. 
As it could be seen, our model mostly succeed to capture the styles, except for 'Debi Gliori' on both content images, who has the worst classification performance in the previous experiments as well due to large variations in her style.  
%Another fail happens, although difficult to observe visually, with the illustration from Beatrix Potter is selected as the style and the illustration from 'Feridun Oral' is selected as the content image. 
%Since the algorithm transferred the style mainly for colors, we see that color is an important feature for defining style of illustrators. Failure cases happened for repeating images.

%% style transfer

% disc patch results 1
\begin{figure*}
\centering
\begin{center}%{cc}
%%%%% 1st row 
\includegraphics[width=0.45\textwidth]{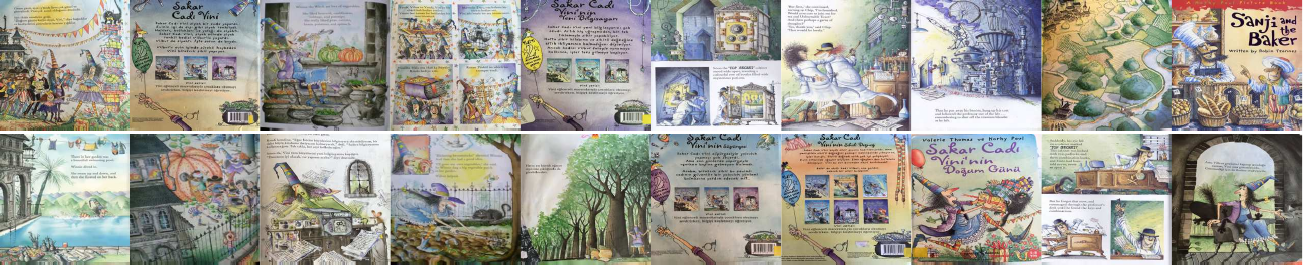} 
 \hspace*{0.001cm}
\includegraphics[width=0.45\textwidth]{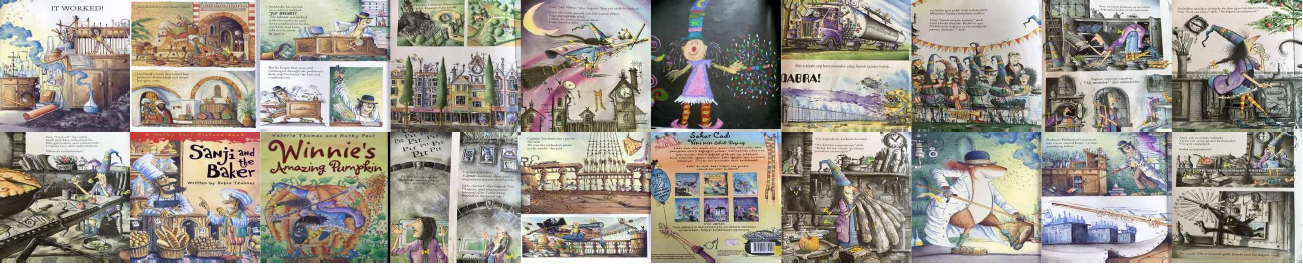}
\vspace*{0.25cm}
 %\hspace*{0.001cm}
\includegraphics[width=0.45\textwidth]{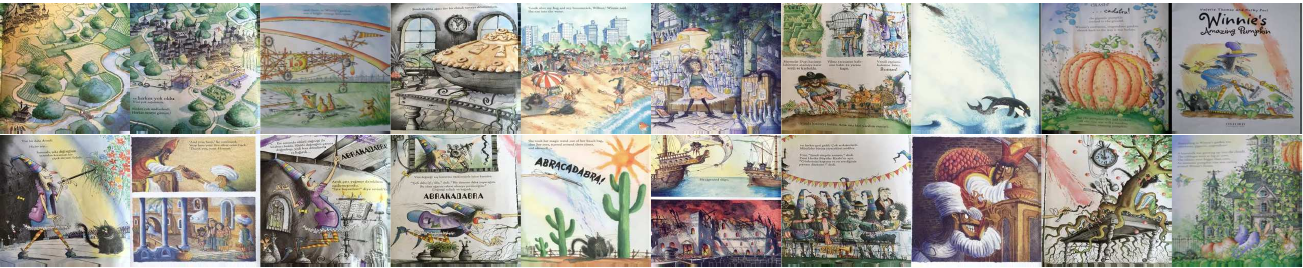} 
 \hspace*{0.001cm}
\includegraphics[width=0.45\textwidth]{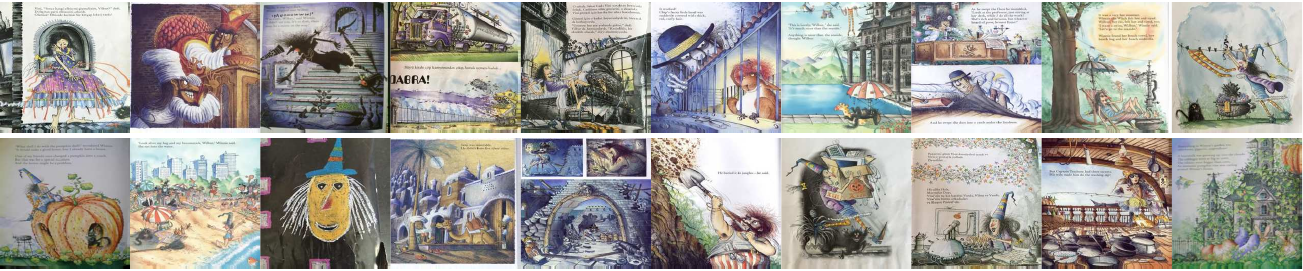}  
 \end{center}
 \caption{First 20 representative instances obtained by \cite{paris} (top-left), and by the method of \cite{fame} using HOG (top-right), color dense SIFT (bottom-left), and  VGG19 fine tuned (bottom-right) features.}
\label{fig:disc_patches}
\end{figure*}%

% disc patch results 2
\begin{figure*}
\centering
\begin{center}%{cc}
%%%%% 1st row 
\includegraphics[width=0.45\textwidth]{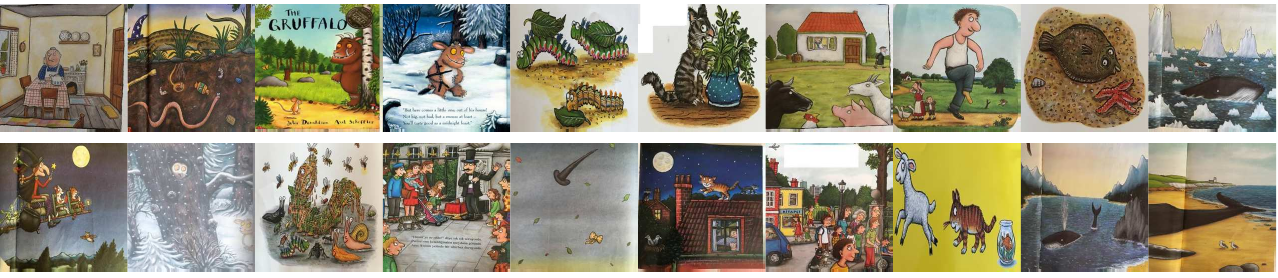} 
 \hspace*{0.001cm}
\includegraphics[width=0.45\textwidth]{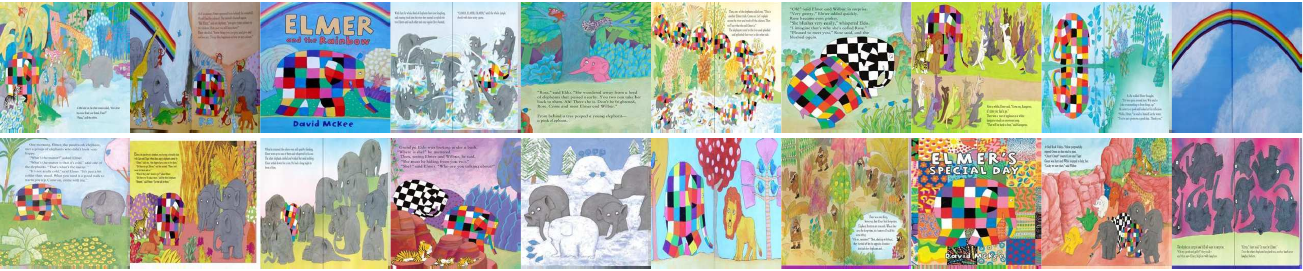}  \\
\vspace*{0.25cm}
\includegraphics[width=0.45\textwidth]{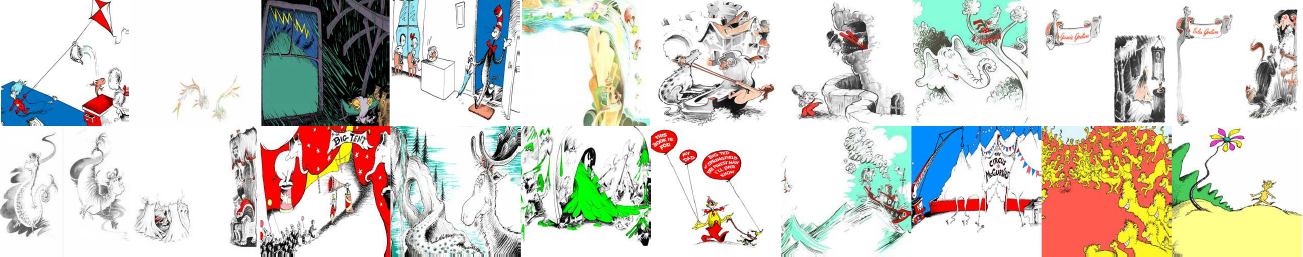} 
 \hspace*{0.001cm}
\includegraphics[width=0.45\textwidth]{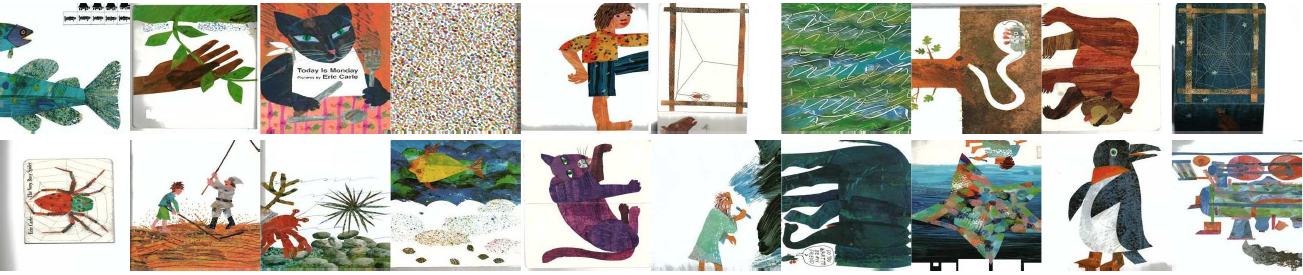} 
%\\
%\vspace*{0.25cm}
%\includegraphics[width=0.45\textwidth]{./figures/disc_patch/maurice-eps-converted-to.pdf} 
% \hspace*{0.001cm}
%\includegraphics[width=0.45\textwidth]{./figures/disc_patch/tony-eps-converted-to.pdf}
 \end{center}
 \caption{Representative instances for some other illustrators using the method in \cite{fame} by VGG19 fine tuned features.}
\label{fig:disc_patches_others}
\end{figure*}%

% Discriminative patches from Paris paper
\begin{figure}
\centering
\begin{center}%{cc}
\includegraphics[width=0.06\textwidth]{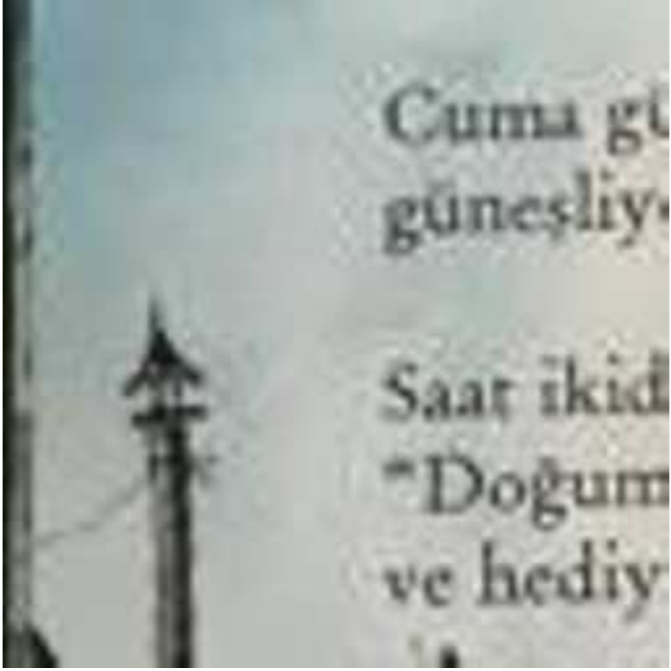} 
 \hspace*{0.001cm}
\includegraphics[width=0.06\textwidth]{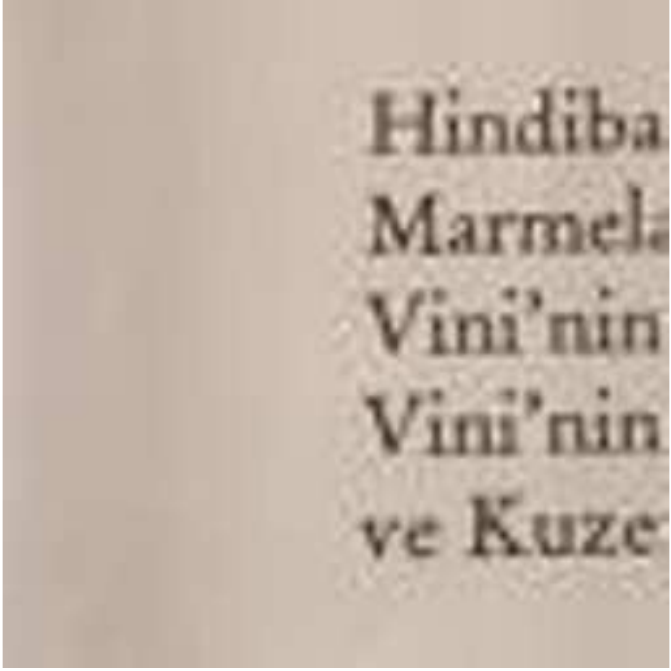}  
 \hspace*{0.001cm}
\includegraphics[width=0.06\textwidth]{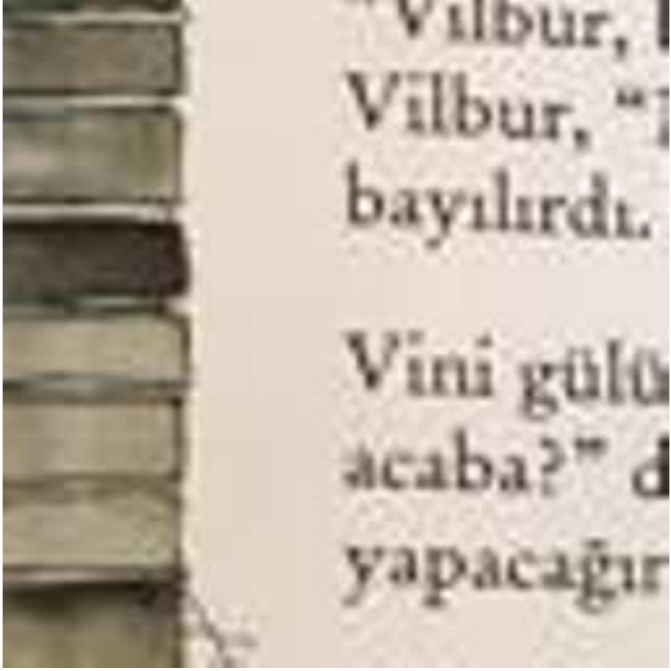} 
 \hspace*{0.001cm}
\includegraphics[width=0.06\textwidth]{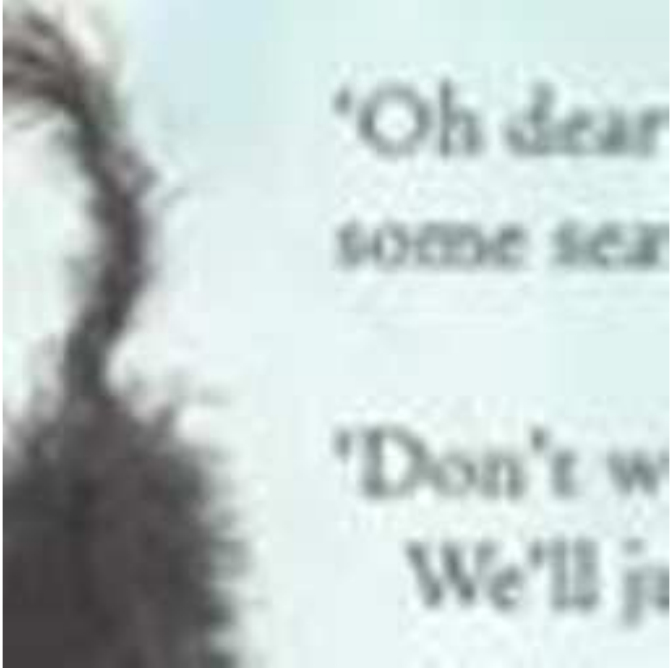} 
 \hspace*{0.001cm}
\includegraphics[width=0.06\textwidth]{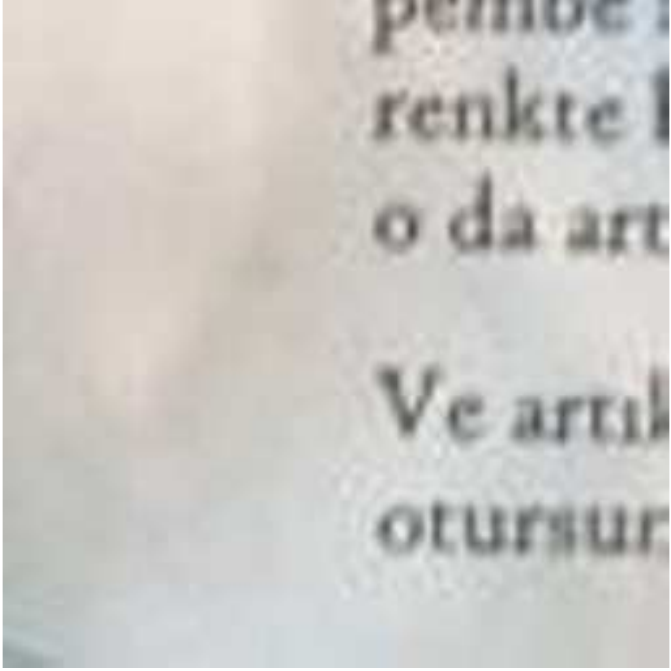} \\

\includegraphics[width=0.06\textwidth]{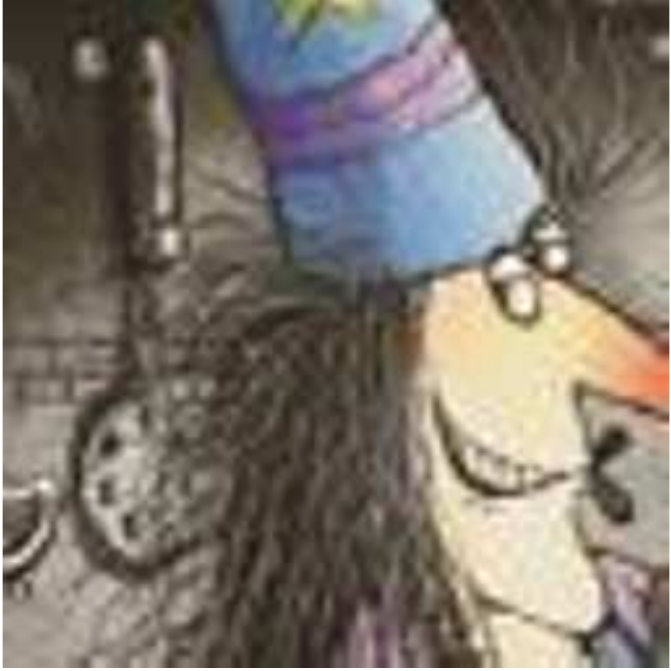} 
 \hspace*{0.001cm}
\includegraphics[width=0.06\textwidth]{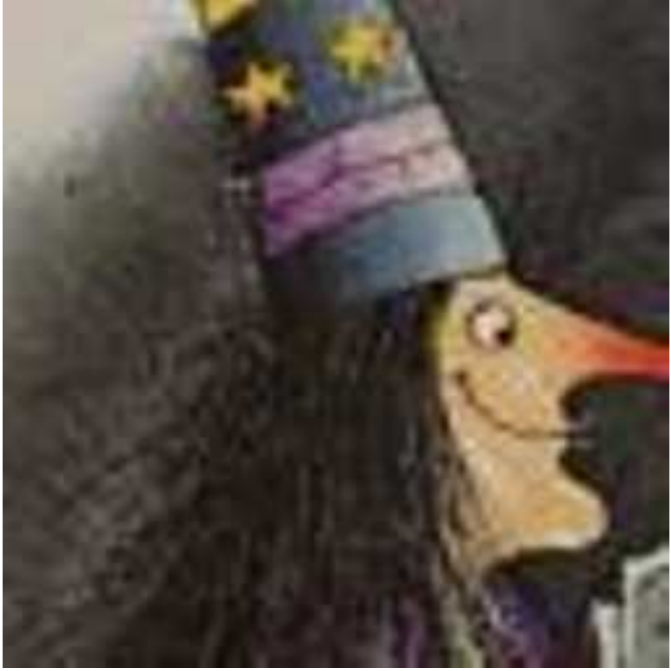}  
 \hspace*{0.001cm}
\includegraphics[width=0.06\textwidth]{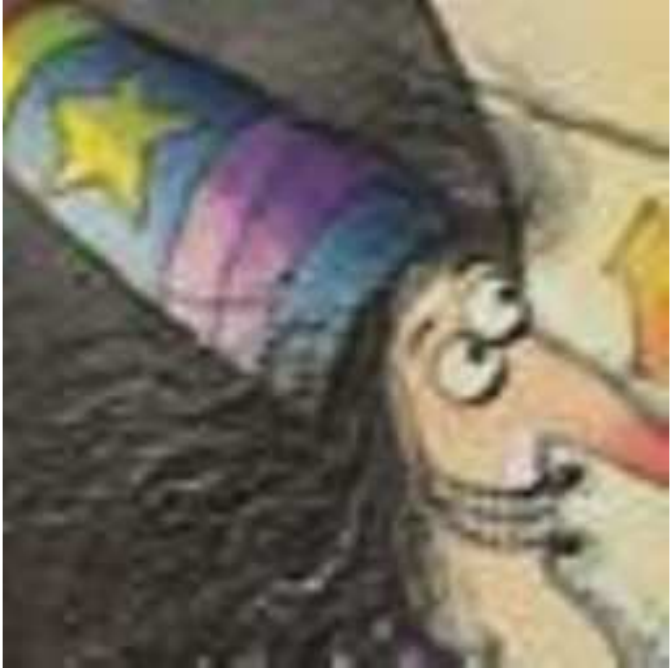} 
 \hspace*{0.001cm}
\includegraphics[width=0.06\textwidth]{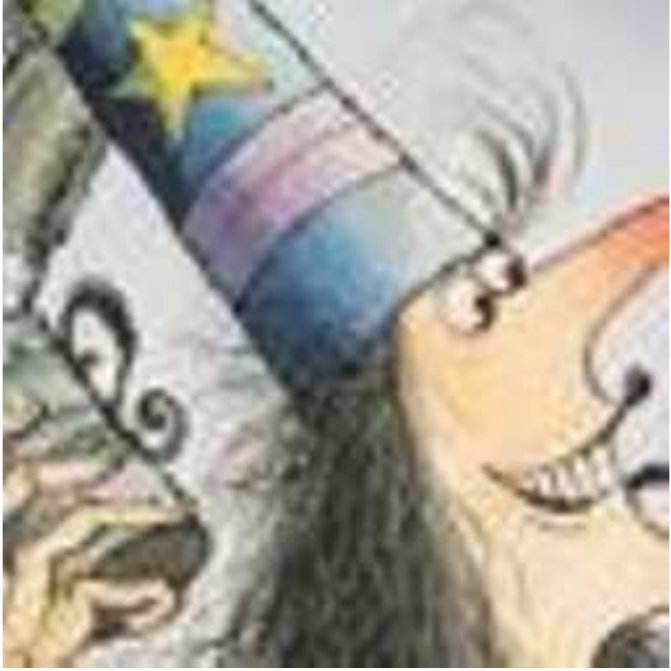} 
 \hspace*{0.001cm}
\includegraphics[width=0.06\textwidth]{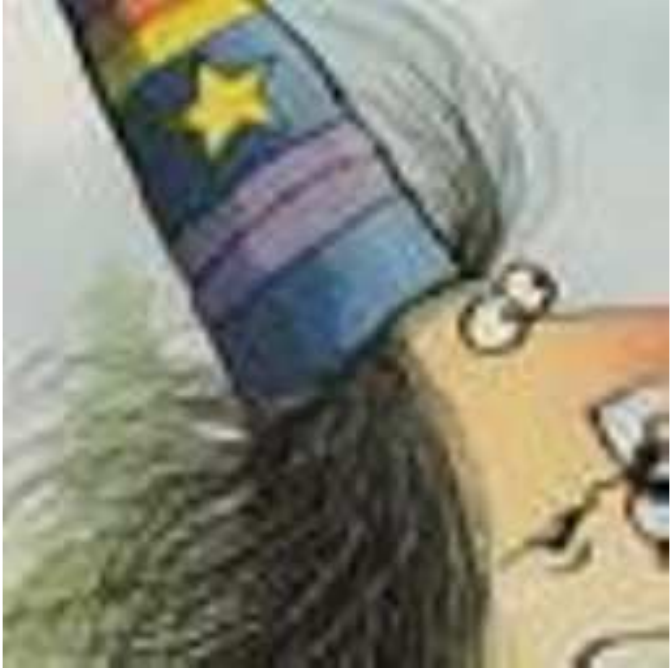} \\

%\includegraphics[width=0.06\textwidth]{./figures/paris_patch/6_1-eps-converted-to.pdf} 
 %\hspace*{0.001cm}
%\includegraphics[width=0.06\textwidth]{./figures/paris_patch/6_2-eps-converted-to.pdf}  
 %\hspace*{0.001cm}
%\includegraphics[width=0.06\textwidth]{./figures/paris_patch/6_3-eps-converted-to.pdf} 
 %\hspace*{0.001cm}
%\includegraphics[width=0.06\textwidth]{./figures/paris_patch/6_4-eps-converted-to.pdf} 
 %\hspace*{0.001cm}
%\includegraphics[width=0.06\textwidth]{./figures/paris_patch/6_5-eps-converted-to.pdf} \\

\includegraphics[width=0.06\textwidth]{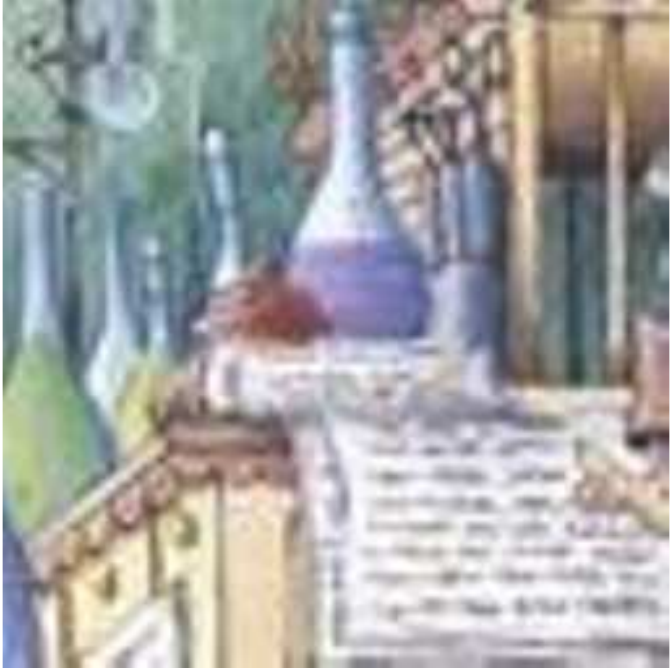} 
 \hspace*{0.001cm}
\includegraphics[width=0.06\textwidth]{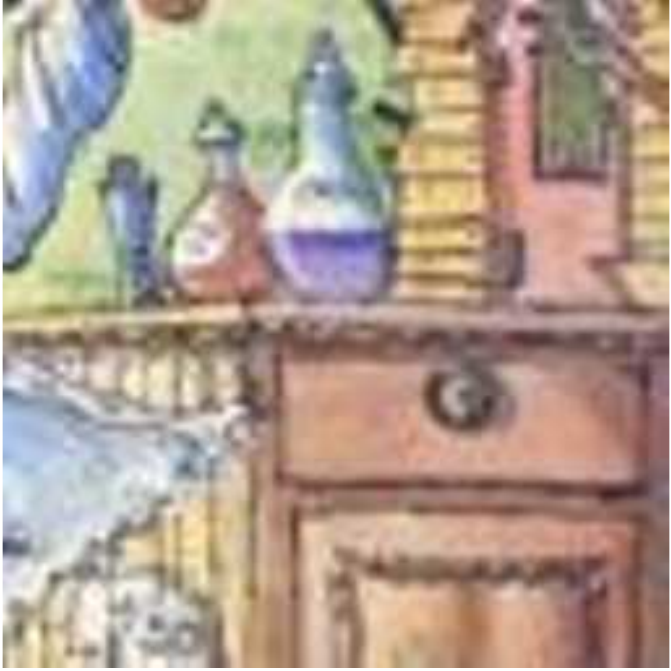}  
 \hspace*{0.001cm}
\includegraphics[width=0.06\textwidth]{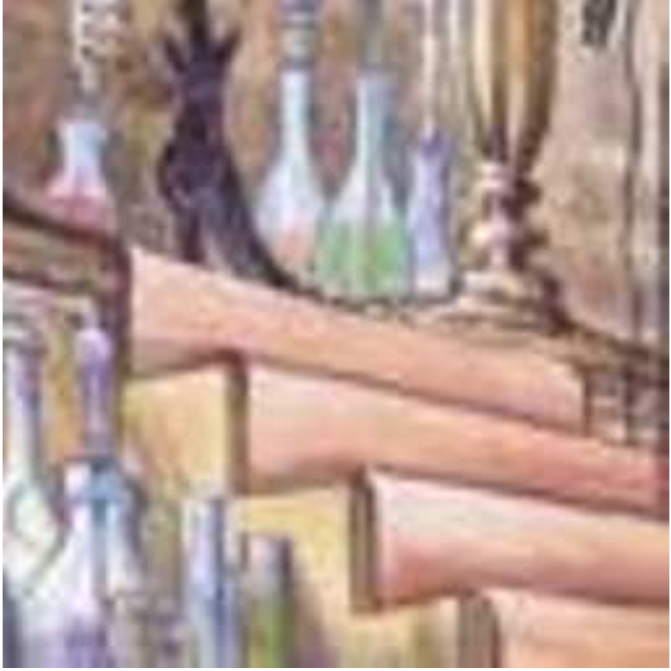} 
 \hspace*{0.001cm}
\includegraphics[width=0.06\textwidth]{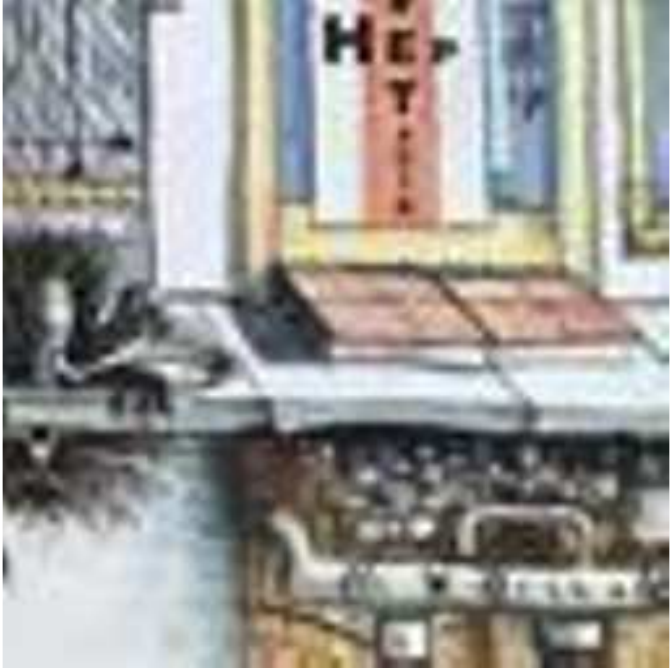} 
 \hspace*{0.001cm}
\includegraphics[width=0.06\textwidth]{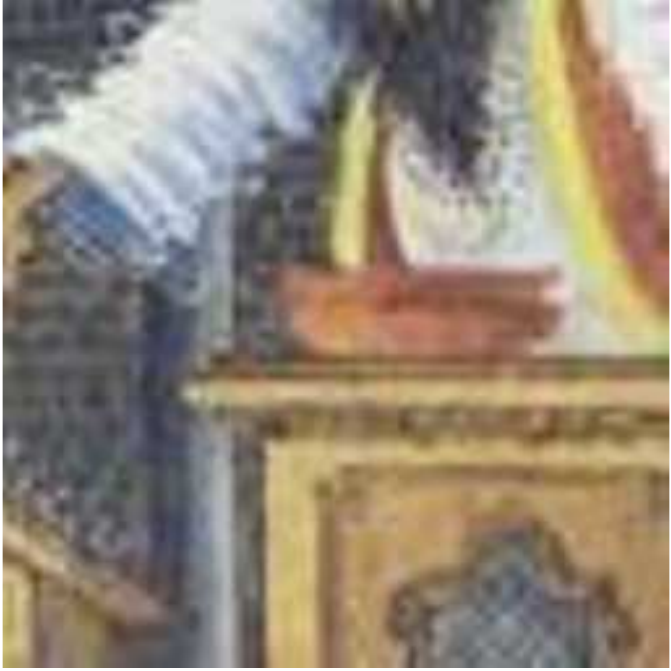} \\

\includegraphics[width=0.06\textwidth]{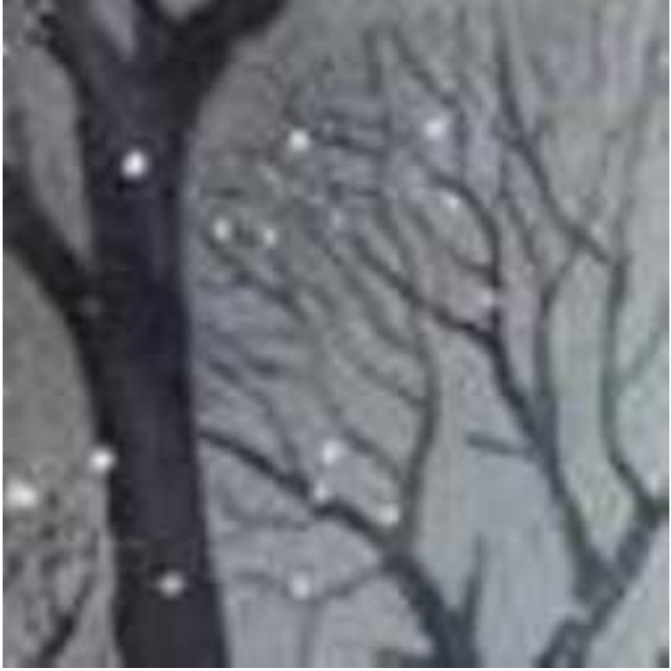} 
 \hspace*{0.001cm}
\includegraphics[width=0.06\textwidth]{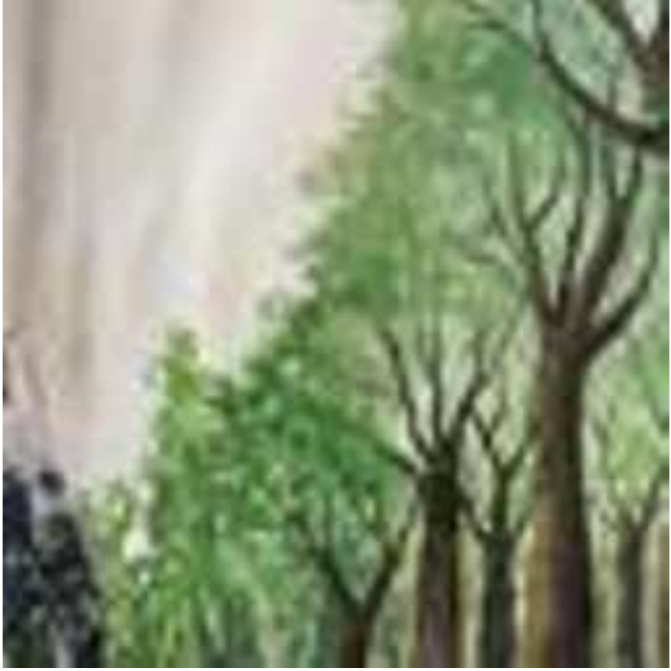}  
 \hspace*{0.001cm}
\includegraphics[width=0.06\textwidth]{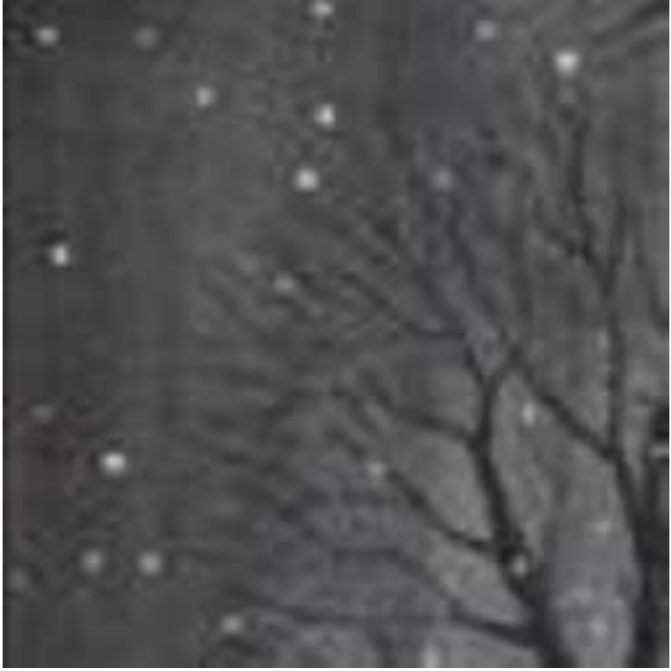} 
 \hspace*{0.001cm}
\includegraphics[width=0.06\textwidth]{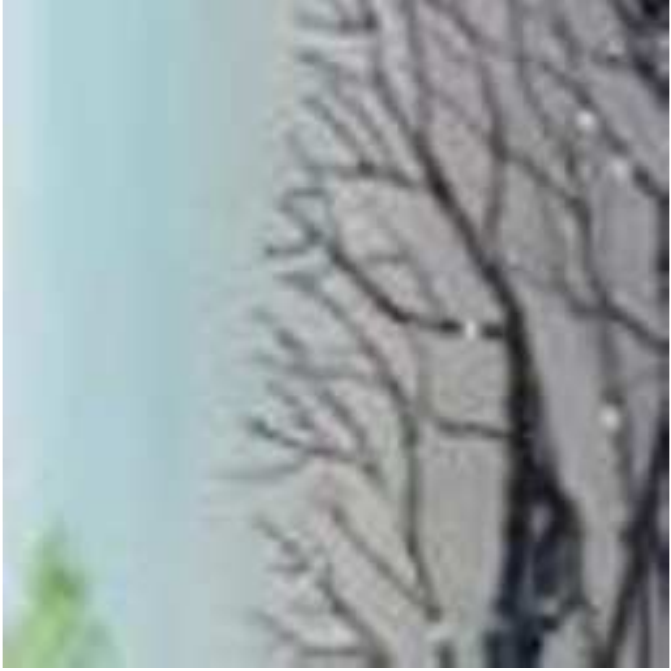} 
 \hspace*{0.001cm}
\includegraphics[width=0.06\textwidth]{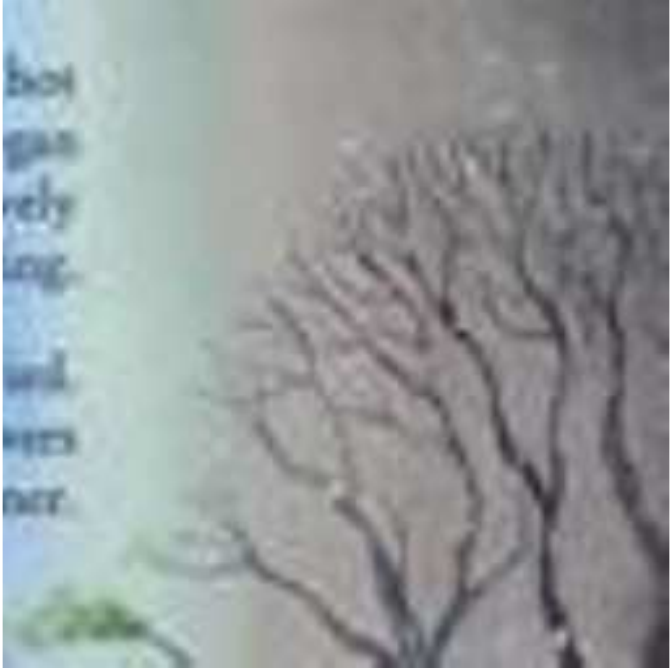} \\ 

%\includegraphics[width=0.06\textwidth]{./figures/paris_patch/2_1-eps-converted-to.pdf} 
 %\hspace*{0.001cm}
%\includegraphics[width=0.06\textwidth]{./figures/paris_patch/2_2-eps-converted-to.pdf}  
 %\hspace*{0.001cm}
%\includegraphics[width=0.06\textwidth]{./figures/paris_patch/2_3-eps-converted-to.pdf} 
 %\hspace*{0.001cm}
%\includegraphics[width=0.06\textwidth]{./figures/paris_patch/2_4-eps-converted-to.pdf} 
 %\hspace*{0.001cm}
%\includegraphics[width=0.06\textwidth]{./figures/paris_patch/2_5-eps-converted-to.pdf} \\

\includegraphics[width=0.06\textwidth]{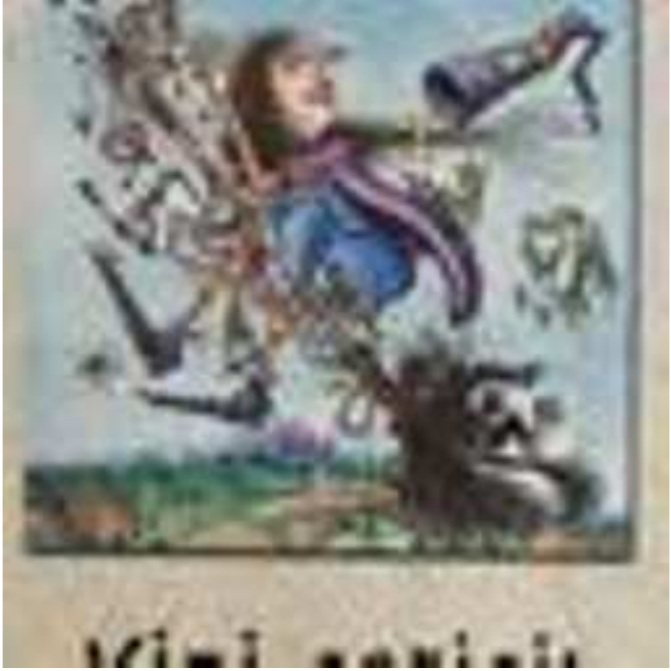} 
 \hspace*{0.001cm}
\includegraphics[width=0.06\textwidth]{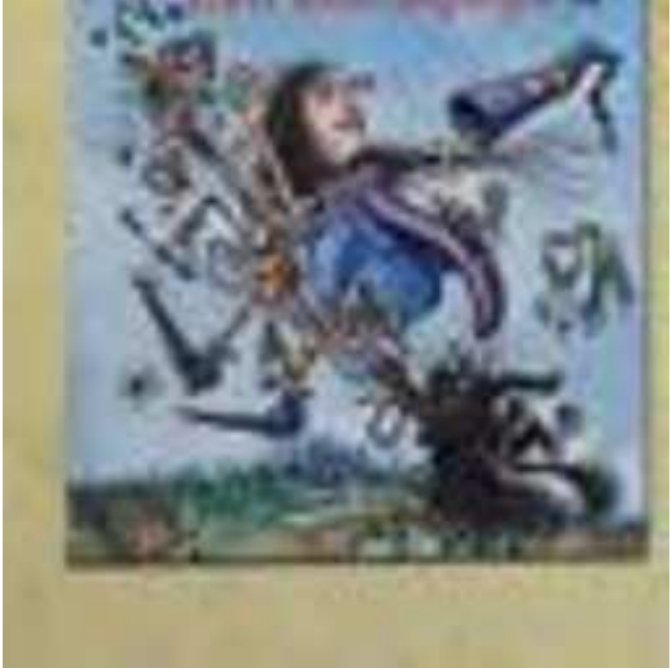}  
 \hspace*{0.001cm}
\includegraphics[width=0.06\textwidth]{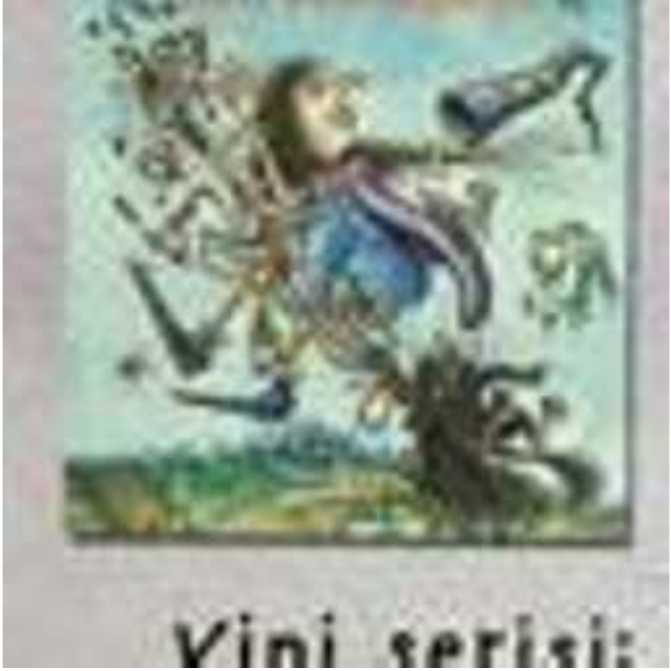} 
 \hspace*{0.001cm}
\includegraphics[width=0.06\textwidth]{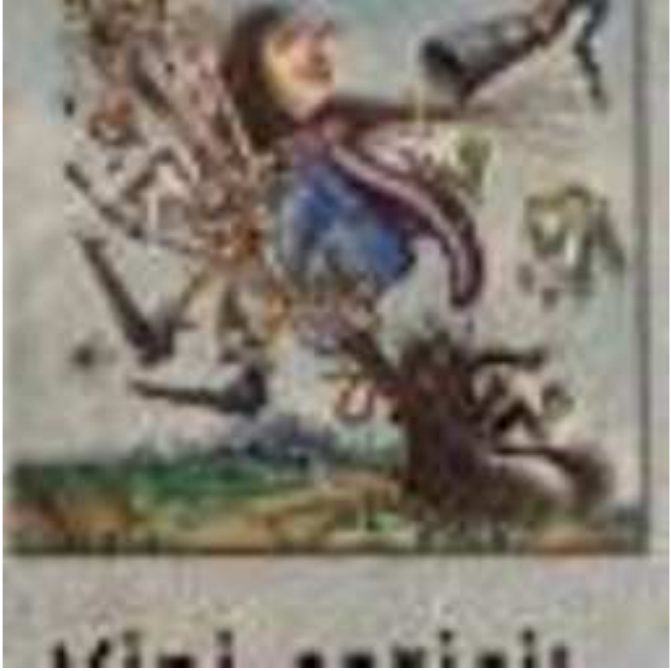} 
 \hspace*{0.001cm}
\includegraphics[width=0.06\textwidth]{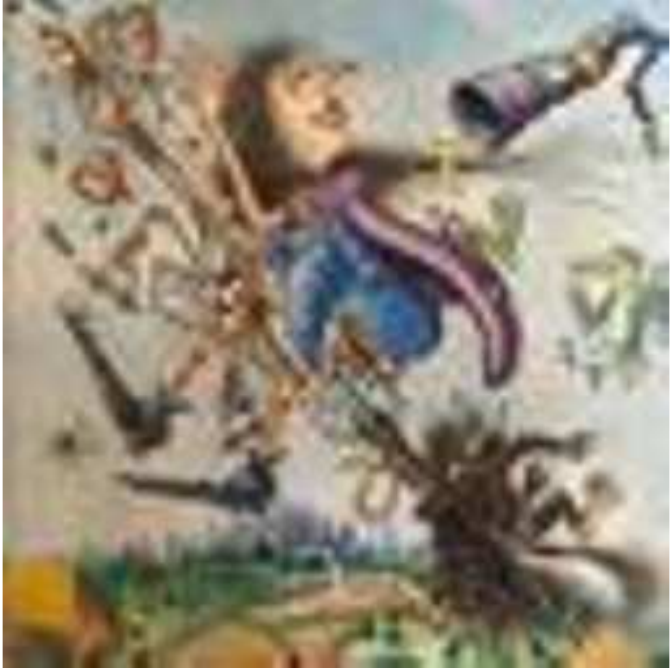} 
 \end{center}
 \caption{Disriminative patches from Korky Paul obtained by~\cite{paris}.  The first four rows correspond to discriminative stylistic parts seen in his illustrations. Note that the writing style is also captured as discriminative. However, as in the last row repeating images at the back pages of all books were also selected as discriminative as a failure case.}
\label{fig:disc_patches_korky}
\end{figure}%

\subsection{Representative and discriminative elements}
\label{sec:disc_patch_exp}
First, we aimed to find representative illustrations of each illustrator. As depicted in Figure~\ref{fig:disc_patches}, we compared the method in \cite{paris}, with the method in \cite{fame} first using HOG features in both methods. Then, we utilised color dense SIFT and VGG19 fined tuned features with \cite{fame} as well. 
Note that, since~\cite{paris} produces patches while~\cite{fame} gives images, only way to compare results of both algorithms was to find images which contain most of the extracted patches. While \cite{paris} is likely to choose the pages with text as considering the font style being discriminative, \cite{fame} is more likely to capture the style forced by the chosen feature. VGG19 was able to capture the dark colors and the strokes better than the others. Since the visual examples are subjective, in order to quantitatively compare the performance of different methods for selection of representatives we used the categorisation performance. For the first 50 images \cite{paris} resulted in 1 incorrect classification and the others reported 100\% accuracy. For a better analysis though we should look at the full list and find better comperative measures. Figure~\ref{fig:disc_patches_others} shows the representatives for some other illustrators using VGG19 features with \cite{fame}. As a final experiment, we explored the patches extracted by \cite{paris} in Figure~\ref{fig:disc_patches_korky} for the Korky Paul images. As seen, we are able to select stylistic elements like the head of the witch, leafless trees, or furniture, and even the typeface of fonts as discriminative elements.

\section{Conclusion}

We attacked the problem of recognizing style of illustrators as a pioneering work in this area. On the new dataset constructed we reported qualitative and quantitative results for three different applications: illustrator recognition, style transfer and representative instance selection. In our future work, we plan to expand the dataset with more illustrators. Moreover, better metrics are required to evaluate the quality of style transfer and selection of representatives.

\bibliographystyle{ACM-Reference-Format}
\bibliography{egbib} 

\end{document}